%% file: main.tex
\documentclass{article} 
\usepackage{arxiv}
\setcitestyle{authoryear,open={(},close={)}}

\usepackage{fullpage}

\input{math_commands.tex}

\usepackage{url}

\usepackage{graphicx}
\usepackage{wrapfig}
\usepackage{subfig}
\usepackage{algorithm}
\usepackage[noend]{algorithmic}
\usepackage{bbm}

\usepackage{amsmath}
\usepackage{amssymb}
\usepackage{mathtools}
\usepackage{amsthm}
\usepackage{enumitem}
\usepackage{natbib}
\usepackage{xcolor}

\theoremstyle{plain}
\newtheorem{theorem}{Theorem}[section]
\newtheorem{proposition}[theorem]{Proposition}

\theoremstyle{definition}

\theoremstyle{remark}

\title{\textbf{Pre-Training and Fine-Tuning\\Generative Flow Networks}}

\author{
{\large Ling Pan$^{1}$\thanks{Correspondence to \texttt{penny.ling.pan@gmail.com}.} ~ Moksh Jain$^{1}$ ~ Kanika Madan$^{1}$ ~ Yoshua Bengio$^{1}$} \\
 \\
{\large $^{1}$ Mila, Universit\'e de Montr\'eal $^{2}$ CIFAR AI Chair}
}

\begin{document}

\maketitle

\begin{abstract}
Generative Flow Networks (GFlowNets) are amortized samplers that learn stochastic policies to sequentially generate compositional objects from a given unnormalized reward distribution.
They can generate diverse sets of high-reward objects, which is an important consideration in scientific discovery tasks. However, as they are typically trained from a given extrinsic reward function, it remains an important open challenge about how to leverage the power of pre-training and train GFlowNets in an unsupervised fashion for efficient adaptation to downstream tasks.
Inspired by recent successes of unsupervised pre-training in various domains, we introduce a novel approach for reward-free pre-training of GFlowNets. By framing the training as a self-supervised problem, we propose an outcome-conditioned GFlowNet (OC-GFN) that learns to explore the candidate space. Specifically, OC-GFN learns to reach any targeted outcomes, akin to goal-conditioned policies in reinforcement learning. 
We show that the pre-trained OC-GFN model can allow for a direct extraction of a policy capable of sampling from any new reward functions in downstream tasks.
Nonetheless, adapting OC-GFN on a downstream task-specific reward involves an intractable marginalization over possible outcomes. We propose a novel way to approximate this marginalization by learning an amortized predictor enabling efficient fine-tuning.
Extensive experimental results validate the efficacy of our approach, demonstrating the effectiveness of pre-training the OC-GFN, and its ability to swiftly adapt to downstream tasks and discover modes more efficiently.
This work may serve as a foundation for further exploration of pre-training strategies in the context of GFlowNets.
\end{abstract}

\section{Introduction}
Unsupervised learning on large stores of data on the internet has resulted in significant advances in a variety of domains~\citep{howard2018universal,devlin2018bert,radford2019language,henaff2020data}. Pre-training with unsupervised objectives, such as next-token prediction in auto-regressive language models~\citep{radford2019language}, on large-scale unlabelled data enables the development of models that can be effectively fine-tuned for novel tasks using few samples~\citep{brown2020language}. Unsupervised learning at scale allows models to learn good representations, which enables data-efficient adaptation to novel tasks, and is central to the recent development towards larger models. 

On the other hand, in the context of amortized inference, Generative Flow Networks~\citep[GFlowNets;][]{bengio2021flow} enable learning generative models for sampling from high-dimensional distributions over discrete compositional objects. Inspired by reinforcement learning (RL), GFlowNets learn a stochastic policy to sequentially generate compositional objects with a probability proportional to a given reward, instead of reward maximization. Therefore, GFlowNets have found success in applications to scientific discovery problems to generate high-quality and diverse candidates~\citep{bengio2021flow,jain2023gflownets,jain2023multi} as well as alternatives to Monte-Carlo Markov chains and variational inference for modeling Bayesian posteriors~\citep{deleu2022bayesian,van2022nesi}. 

As a motivating example, consider the drug discovery pipeline. A GFlowNet can be trained to generate candidate RNA sequences that bind to a target using as reward the binding affinity of the RNA sequence with the target~\citep{lorenz2011viennarna,sinai2020adalead} (which can be uncertain and imperfect based on the current understanding of the biological system and available experimental data). However, there is no way to efficiently adapt the GFlowNet to sample RNA sequences binding to a different target of interest that reflects new properties. 
Unlike human intelligence, GFlowNets currently lack the ability to leverage previously learned knowledge to efficiently adapt to new tasks with unseen reward functions, and need to be trained from scratch to learn a policy for matching the given extrinsic reward functions for different tasks.
Inspired by the success of the unsupervised pre-training and fine-tuning paradigm in vision, language, and RL~\citep{jaderberg2016reinforcement,sekar2020planning} domains, it is natural to ask how can this paradigm benefit GFlowNets. As a step in this direction, we propose a fundamental approach to realize this paradigm for GFlowNets. 

In this paper, we propose a novel method for reward-free unsupervised pre-training of GFlowNets. We formulate the problem of pre-training GFlowNets as a 
self-supervised problem of learning an outcome-conditioned GFlowNet (OC-GFN) which learns to reach any outcome (goal) as a functional understanding of the environment (akin to goal-conditioned RL~\citep{chebotar2021actionable}).
The reward for training this OC-GFN is defined as the success of reaching the outcome. 
Due to the inherent sparse nature of this task-agnostic reward, it introduces critical challenges for efficient training OC-GFN in complex domains, particularly with higher-dimensional outcomes and long-horizon problems, since it is difficult to reach the outcome to get a meaningful reward signal. 
To tackle these challenges, we introduce a novel contrastive learning procedure to train OC-GFN to effectively handle such sparse rewards, which induces an implicit curriculum for efficient learning that resembles goal relabeling~\citep{andrychowicz2017hindsight}.
To enable efficient learning in long-horizon tasks, we further propose a goal teleportation scheme to effectively propagate the learning signal to each step.

A remarkable result is that one can directly convert this pre-trained OC-GFN to sample proportionally to a new reward function for downstream tasks~\citep{bengio2023foundations}.
It is worth noting that in principle, this can be achieved even without re-training the policy, which is usually required for fine-tuning in RL, as it only learns a reward-maximizing policy that may discard many useful information.
Adapting the pre-trained OC-GFN model to a new reward function, however, involves an intractable marginalization over possible outcomes.
We propose a novel alternative by learning a predictor that amortizes this marginalization, allowing efficient fine-tuning of the OC-GFN to downstream tasks. Our key contributions can be summarized as follows:
\begin{itemize}[leftmargin=*] 
    \item We propose reward-free pre-training for GFlowNets as training outcome-conditioned GFlowNet (OC-GFN) that learns to sample trajectories to reach any outcome. 
    \item We investigate how to leverage the pre-trained OC-GFN model to adapt to downstream tasks with new rewards, 
    and we also introduce an efficient method to learn an amortized predictor to approximate an intractable marginal required for fine-tuning the pre-trained OC-GFN model.
    \item Through extensive experiments on the GridWorld domain, we empirically validate the efficacy of the proposed pre-training and fine-tuning paradigm. We also demonstrate its scalability to larger-scale and challenging biological sequence design tasks, which achieves consistent and substantial improvements over strong baselines, especially in terms of diversity of the generated samples.
\end{itemize}

\section{Preliminaries} \label{sec:prelim}
Given a space of compositional objects $\mathcal{X}$, and a non-negative reward function $R: \mathcal{X}\mapsto\mathbb{R}^+$, the GFlowNet policy $\pi$ is trained towards sampling objects $x\in \mathcal{X}$ from the distribution defined by $R(x)$, i.e., $\pi(x)\propto R(x)$. The compositional objects are each sampled sequentially, with each step involving the addition of a building block $a\in \mathcal{A}$ (action space) to the current partially constructed object $s\in\mathcal{S}$ (state space).
We can define a directed acyclic graph (DAG) $\mathcal{G}=\{\mathcal{S}, \mathcal{A}\}$ with the partially constructed objects forming nodes of $\mathcal{G}$, including a special empty state $s_0$. The edges are $s\rightarrow s'$, where $s'$ is obtained by applying an action $a\in\mathcal{A}$ to $s$. The complete objects $\mathcal{X}$ are the terminal (childless) nodes in the DAG.
The generation of an object $x\in \mathcal{X}$ corresponds to complete trajectories in the DAG starting from $s_0$ and terminating in a terminal state $s_n=x$, i.e., $\tau=(s_0\rightarrow \dots \rightarrow x)$.
We assign a non-negative weight, called \emph{state flow} $F(s)$ to each state $s\in \mathcal{S}$. 
The forward policy $P_F(s'|s)$ is a collection of distributions over the children of each state and the backward policy $P_B(s|s')$ is a collection of distributions over the parents of each state. The forward policy induces a distribution over trajectories $P_F(\tau)$, and the marginal likelihood of sampling a terminal state is given by marginalizing over trajectories terminating in $x$, $P_F^\top(x)=\sum_{\tau=(s_0\rightarrow\dots\rightarrow x)}P_F(\tau)$. GFlowNets solve the problems of learning a parameterized policy $P_F(\cdot\mid\cdot;\theta)$ such that $P_F^\top(x)\propto R(x)$. \\

\noindent {\textbf{Training GFlowNets.}} 
We use the \textit{detailed balance} learning objective~\citep[DB;][]{bengio2023foundations} to learn the parameterized policies and flows based on Eq.~(\ref{eq:db}), which considers the flow consistency constraint in the edge level (i.e., the incoming flow for edge $s\to s'$ matches the outgoing flow). When it is trained to completion,
the objective yields the desired policy.
\begin{equation}
\forall s\to s' \in \mathcal{A}, \quad\quad F(s)P_F(s'|s) = F(s')P_B(s|s').
\label{eq:db}
\end{equation}
As in reinforcement learning, exploration is a key challenge in GFlowNets.
Generative Augmented Flow Networks~\citep[GAFlowNets;][]{pan2022gafn} incorporate intrinsic intermediate rewards represented as augmented flows in the flow network to drive exploration, where $r_i(s \to s')$ is specified by intrinsic motivation~\citep{burda2018exploration}, yielding the following variant of detailed balance.
\begin{equation}
\forall s\rightarrow s' \in \mathcal{A},\quad\quad F(s)P_F(s'|s) = F(s') P_B(s|s') + r(s \to s').
\label{eq:uncond_constraint_db}
\end{equation}

\section{Related Work}
{\textbf{Generative Flow Networks.}} 
Initially inspired  by reinforcement learning, GFlowNets~\citep{bengio2021flow,bengio2023foundations} also have close connections to hierarchical variational inference~\citep{malkin2022gfnhvi,zimmermann2022variational}. While several learning objectives have been proposed for improving credit assignment and sample efficiency in GFlowNets such as detailed balance~\citep{bengio2023foundations}, sub-trajectory balance~\citep{madan2022learning} and forward-looking objectives~\citep{pan2023better}, they need to be trained from scratch with a given reward function, which may limit its applicability to more practical problems.
Owing to their flexibility, GFlowNets have been applied to wide range of problems where diverse high-quality candidates are need, such as molecule generation~\citep{bengio2021flow}, biological sequence design~\citep{jain2022biological}, combinatorial optimization~\citep{zhang2023robust,zhang2023let}, multi-objective~\citep{jain2023multi}.
As ammortized samplers, GFlowNets have been used for Bayesian structure learning~\citep{deleu2022bayesian}, neurosymbolic programming~\citep{van2022nesi}, and for variational EM~\citep{hu2023gflownet}. Extensions for continuous~\citep{lahlou2023theory} and stochastic spaces~\citep{pan2023stochastic,zhang2023distributional} have also been studied.\\

\noindent {\textbf{Unsupervised Pre-Training in Reinforcement Learning.}} 
Following the progress in language modeling and computer vision, there has been growing interest in pre-training reinforcement learning (RL) agents in an unsupervised stage without access to task-specific rewards for learning representations. 
Agents typically learn a set of different skills~\citep{eysenbach2018diversity,hansen2019fast,zhao2021mutual,liu2021aps}, and then fine-tune the learned policy to downstream tasks. Contrary to reward-maximization, GFlowNets learn a stochastic policy to match the reward distribution. As we show in the next section, such a learned policy can be adapted to a new reward function even without re-training~\citep{bengio2023foundations}.\\

\noindent {\textbf{Goal-Conditioned Reinforcement Learning.}}
Different from standard reinforcement learning (RL) methods that learn policies or value functions based on observations, goal-conditioned RL~\citep{kaelbling1993learning} also take goals into consideration by augmenting the observations with an additional input of the goal and have been studied in a number of prior works~\citep{schaul2015universal,nair2018visual,veeriah2018many,eysenbach2020c}.
Goal-conditioned RL is trained to greedily achieve different goals specified explicitly as input, making it possible for agents to generalize their learned abilities across different environments.

\section{Pre-training and fine-tuning generative flow networks}
In the original formulation, a GFlowNet 
need to be trained from scratch whenever it is faced with a previously unseen reward function.
In this section, we aim to leverage the power of pre-training in GFlowNets for efficient adaptation to downstream tasks.

To tackle this important challenge, we propose a novel approach to frame the problem of pre-training GFlowNets as a self-supervised problem, by training an outcome-conditioned GFlowNet that learns to reach any input terminal state (outcome) without task-specific rewards. 
Then, we propose to leverage the power of the pre-trained GFlowNet model for efficiently fine-tuning it to downstream tasks with a new reward function.

\begin{wrapfigure}{r}{0.45\textwidth} \vspace{-.2in}
\centering
\includegraphics[width=1.\linewidth]{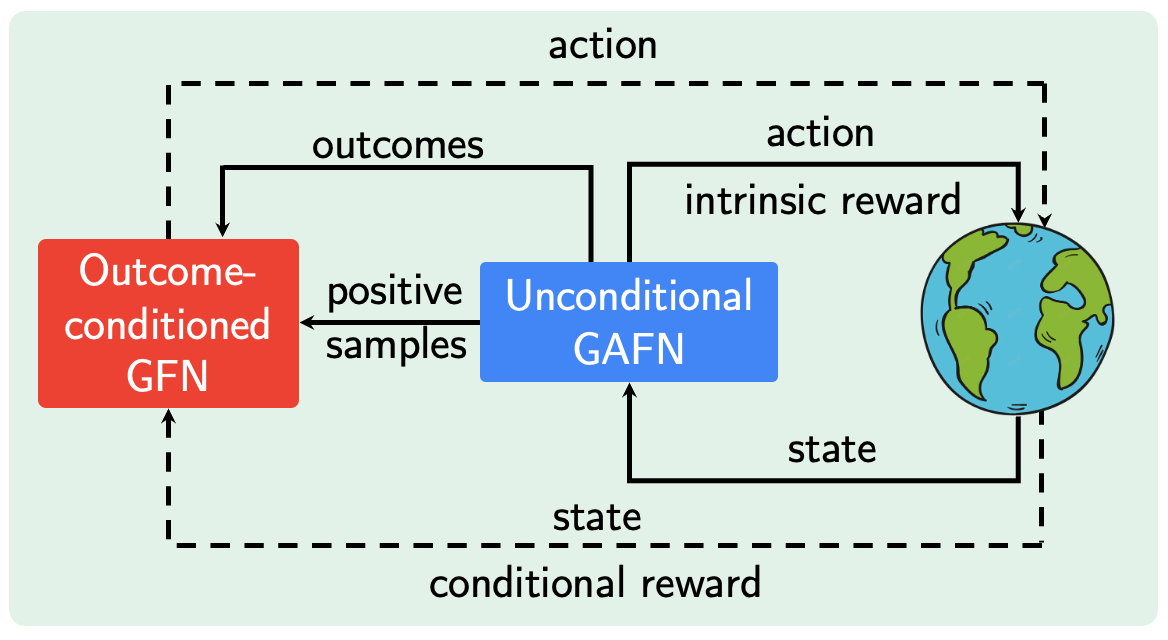}
\caption{The unsupervised pre-training phase of outcome-conditioned GFlowNet.}
\vspace{-.1in}
\label{fig:stage1}
\end{wrapfigure}
\subsection{Unsupervised Pre-Training Stage}
As GFlowNets are typically learned with a given reward function, it remains an open challenge for how to pre-train them in a reward-free fashion.
We propose a novel method for unsupervised pre-training of GFlowNets without task-specific rewards.
We formulate the problem of pre-training GFlowNets as a self-supervised problem of learning an outcome-conditioned GFlowNet (OC-GFN) which learns to reach any input target outcomes, inspired by the success of goal-conditioned reinforcement learning in generalizing to a variety of tasks when it is tasked with different goals~\citep{fang2022planning}.
We denote the outcome as $y=f(s)$, where $f$ is an identity function and $s$ is a terminal state, so the space of outcomes is the same as the state space $\mathcal{X}$.
What makes OC-GFN special is that, when fully trained, given a reward $R$ a posterior as a function $r$ of the outcome, i.e., $R(s) = r(y)$, one can adapt the OC-GFN to sample from this reward, which can generalize to tasks with different rewards.

\begin{algorithm}[t]
\caption{Reward-free Pre-training of Unsupervised GFlowNets.}
\begin{algorithmic}[1]
\REQUIRE GAFlowNet $F^U(s), P_F^U(s'|s), P_B^U(s|s')$; fixed target network $\bar{\phi}$; predictor network $\phi$; Outcome-conditioned GFlowNet (OC-GFN) $F^C(s|y), P_F^C(s'|s, y), P_B^C(s|s',y)$
\FOR {each training step $t=1$ to $T$}
\STATE Collect a trajectory $\tau^+=\{s_0^+ \to \dots \to s_n^+\}$ with $P_F^U$ \\
\STATE  {\color{gray}// \emph{Update the outcome-conditioned GFlowNet model}} \\
\STATE $y^+\gets f(s^+)$,
$R(s_n^+|y^+) \gets  \mathbbm{1} \{ {f(s^+_n) = y^+} \}\equiv 1$ \\ 
\STATE Update OC-GFN towards minimizing Eq.~(\ref{eq:gt_oc_gfn}) with $\tau^+$ and $R(s_n^+|y^+)$ \\
\STATE 
Collect a trajectory 
$\tau^-=\{s_0^- \to \dots \to s_n^-\}$ with $P_F^C$ conditioned on $y^+$\\
\STATE 
$y^-\gets f(s^-)$,
$R(s_n^-|y^+) \gets  \mathbbm{1} \{ {f(s^-_n) = y^+} \}$ \\
\STATE Update OC-GFN towards minimizing Eq.~(\ref{eq:gt_oc_gfn}) with $\tau^-$ and $R(s_n^-|y^+)$ \\
\STATE {\color{gray}// \emph{Update the unconditional GAFlowNet model}}\\
\STATE Update GAFlowNet towards optimizing Eq.~(\ref{eq:uncond_constraint_db}) based on $\tau^+$ and $r_i \gets ||\bar{\phi}(s) - \phi(s)||_2$ \\
\STATE Update the predictor network $\phi$ towards minimizing $||\bar{\phi}(s) - \phi(s)||_2$ \\
\ENDFOR
\end{algorithmic}
\label{alg:stage1}
\end{algorithm}

\subsubsection{Outcome-Conditioned GFlowNets (OC-GFN)}
We extend the idea of flow functions and policies in GFlowNets to OC-GFN that can generalize to different outcomes $y$ in the outcome space, which is trained to achieve specified $y$.
OC-GFN can be realized by providing an additional input $y$ to the flows and policies, resulting in outcome-conditioned forward and backward policies $P_F(s'|s,y)$ and $P_B(s|s',y)$, and flows $F(s|y)$. The resulting learning objective for OC-GFN 
for intermediate states is shown in Eq.~(\ref{eq:cond_constraint_db}).
\begin{equation}
F(s|y)P_F(s'|s,y) = F(s'|y) P_B(s|s',y).
\label{eq:cond_constraint_db}
\end{equation}

\paragraph{Outcome generation}
We can train OC-GFN by conditioning outcomes-conditioned flows and policies on a specified outcome $y$, and we study how to generate them autotelicly. 
It is worth noting that we need to train it with full-support over $y$. 
We propose to leverage GAFN~\citep{pan2022gafn} with the introduction of augmented flows that enable efficient reward-free exploration purely by intrinsic motivation~\citep{burda2018exploration}.
In practice, we generate diverse outcomes $y$ with GAFN, and provide them to OC-GFN to sample an outcome-conditioned trajectory $\tau=(s_0, \cdots, s_n)$.
The resulting conditional reward is given as 
${R}(s_n|y) = \mathbbm{1} \{ {f(s_n) = y} \}$.
Therefore, OC-GFN receives a zero reward if it fails to reach the target outcome, and a positive reward otherwise, which results in learning an outcome achieving policy. 

\paragraph{Contrastive training}
However, it can be challenging to efficiently train OC-GFN in problems with large outcome spaces.
This is because it can be hard to actually achieve the outcome and obtain a meaningful learning signal owing to the sparse nature of the rewards --- most of the conditional rewards $R(s_n|y)$ will be zero if $f(s_n) \neq y$ when OC-GFN fails to reach the target.

To alleviate this we propose a contrastive learning objective for training the OC-GFN.
After sampling a trajectory $\tau^+=\{s_0^+ \to \cdots \to s_n^+\}$ from unconditional GAFN, we first train an OC-GFN based on this off-policy trajectory by assuming it has the ability to achieve $y^+=s_n^+$ when conditioned on $y^+$. Note that the resulting conditional reward $R(s_n^+|y^+) = \mathbbm{1} \{ {f(s^+_n) = y^+} \} \equiv 1$ in this case, as all correspond to successful experiences that provide meaningful learning signals to OC-GFN.
We then sample another on-policy trajectory $\tau^-=\{s_0^- \to \cdots \to s_n^-\}$ from OC-GFN by conditioning it on $y^+$, and evaluate the conditional reward by $R(s_n^-|y^+) = \mathbbm{1} \{ {f(s^-_n) = y^+} \}$.
Although 
most of $R(s_n^-|y^+)$ can be zero in large outcome spaces during early learning, we provide sufficient successful experiences in the initial phase, which is also related to goal relabeling~\citep{andrychowicz2017hindsight}.
This can be viewed as an implicit curriculum for improving the training of OC-GFN.

\paragraph{Outcome teleportation} The contrastive training paradigm can significantly improve learning efficiency by providing a bunch of successful trajectories with meaningful learning signals, which tackles the particular challenge of sparse rewards during learning. 
However, the agent may still suffer from poor learning efficiency in long-horizon tasks, as it cannot effectively propagate the success/failure signal back to each step.
We propose a novel technique, \emph{outcome teleportation}, for further improving the learning efficiency of OC-GFN as in Eq.~(\ref{eq:gt_oc_gfn_obj}), which considers the terminal reward $R(x|y)$ for every transition. 
\begin{equation}
F(s|y)P_F(s'|s,y) = F(s'|y) P_B(s|s',y)R(x|y).
\label{eq:gt_oc_gfn_obj}
\end{equation}
This formulation can efficiently propagate the guidance signal to each transition in the trajectory, which can significantly improve learning efficiency, particularly in high-dimensional outcome spaces, as investigated in Section~\ref{exp:bit}. 
It can be interpreted as a form of reward decomposition in outcome-conditioned tasks with binary rewards. 
In practice, we train OC-GFN by minimizing the following loss function ${\mathcal{L}}_{{\text OC-GFN}}(\tau, y)$ in log-scale obtained from Eq.~(\ref{eq:gt_oc_gfn}), i.e.,
\begin{equation}
\sum_{s \to s' \in \tau} \left( \log F(s|y) + \log P_F(s'|s, y) - \log F(s'|y) - \log P_B(s|s', y) - \log R(x|y) \right)^2.
\label{eq:gt_oc_gfn}
\end{equation}

\noindent {\textbf{Theoretical justification.}} We now justify that when OC-GFN is trained to completion,
it can successfully reach any specified outcome. The proof can be found in Appendix~\ref{app:proof_oc_gfn}.

\begin{proposition}
If $\mathcal{L}_{{\text OC-GFN}}(\tau,y)=0$ for all trajectories $\tau$ and outcomes $y$, then the outcome-conditioned forward policy $P_F(s'|s,y)$ can successfully reach the target outcome $y$.
\label{thm:oc_gfn}
\end{proposition}
The resulting procedure for the reward-free unsupervised pre-training of OC-GFN is summarized in Algorithm~\ref{alg:stage1} and illustrated Figure~\ref{fig:stage1}.

\subsection{Supervised Fine-Tuning Stage} \label{sec:stage2}
In this section, we study how to leverage the pre-trained OC-GFN model and adapt it for downstream tasks with new reward functions. 

A remarkable aspect of GFlowNets is the ability to demonstrate the reasoning potential in generating a task-specific policy.
By utilizing the pre-trained OC-GFN model with outcome-conditioned flows $F(s|y)$ and policies $P_F(s'|s,y)$,
we can directly obtain a policy that samples according to a new task-specific reward function $R(s)=r(y)$ according to Eq.~(\ref{eq:policy}), which is based on~\citep{bengio2023foundations}. A detailed analysis for this can be found in Appendix~\ref{app:thm_convert_policy}.
\begin{equation}
P_F^{r}(s'|s) = \frac{\sum_y r(y) F(s|y) P_F(s'|s,y)}{\sum_y r(y) F(s|y)}
\label{eq:policy}
\end{equation}
Eq.~(\ref{eq:policy}) serves as the foundation for converting a pre-trained OC-GFN model to handle downstream tasks with new and even out-of-distribution rewards.
Intriguingly, \textit{this conversion can be achieved without any re-training} for OC-GFN on downstream tasks. 
This sets it apart from the typical fine-tuning process in reinforcement learning, which typically requires re-training to adapt a policy, since they generally learn reward-maximizing policies that may discard valuable information.

Directly estimating the above summation often necessitates the use of Monte-Carlo averaging for making each decision.
Yet, this can be computationally expensive in high-dimensional outcome spaces if we need to calculate this marginalization at each decision-making step, which leads to slow thinking.
To improve its efficiency in complex scenarios, it is essential to develop strategies that are both efficient while maintaining accurate estimations.

\begin{figure}[!h]
\centering
\hspace{-0.2cm}
\subfloat{\includegraphics[width=0.373\linewidth]{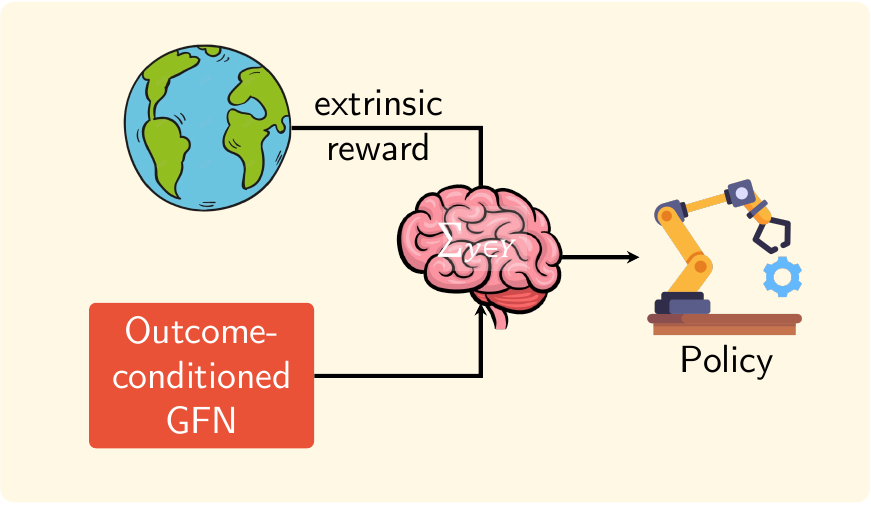}}
\hspace{0.5cm}
\subfloat{\includegraphics[width=0.3\linewidth]{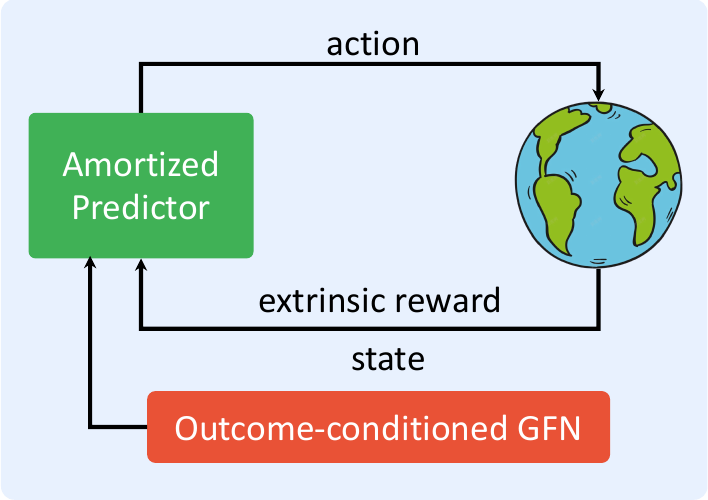}}
\caption{\textit{Left}: Convert the outcome-conditioned GFlowNet to downstream tasks without re-training the networks. \textit{Right}: An efficient amortized predictor in the supervised fine-tuning phase.}
\label{fig:stage2}
\end{figure}

\begin{algorithm}[t]
\caption{Supervised Fine-Tuning of Outcome-Conditioned GFlowNets.}
\begin{algorithmic}[1]
\STATE Initialize the numerator network $N(s'|s)$ and the GFlowNet-like predictor network $Q(y|s',s)$ \\
\STATE Obtain the pre-trained outcome-conditioned state $F(s|y)$ and forward policy $P_F(s'|s,y)$ \\
\FOR {each training step $t=1$ to $T$}
\STATE Collect a trajectory $\tau=\{s_0 \to \cdots s_n\}$ with $N(s'|s)$ \\
\STATE Sample outcomes $y$ from a tempered/$\epsilon$-greedy version of $Q(\cdot|s', s)$ \\
\STATE Update $N$ and $Q$ towards minimizing Eq.~(\ref{eq:amortized}) \\
\ENDFOR
\STATE Compute the policy $P_F^r(s'|s) = {N(s'|s)}/{\sum_{s'} N(s'|s)}$
\end{algorithmic}
\label{alg}
\end{algorithm}

\paragraph{Learning an Amortized Predictor}
In this section, we propose a novel approach to approximate this marginal by learning an amortized predictor.

Concretely, we propose to estimate the intractable sums in the numerator in Eq.~(\ref{eq:policy}) by a numerator network 
$N(s'|s) \approx \sum_y r(y) F(s|y) P_F(s'|s,y)$.
This would allow us to efficiently estimate the intractable sum with the help of $N(s'|s)$ that directly estimates $N(\cdot|s)$ for any state $s$, which could benefit from the generalizable structure of outcomes with neural networks.
We can also obtain the corresponding policy by $P_F^{r}(s'|s) = \frac{N(s'|s)}{\sum_{s'} N(s'|s)}$.
For learning the numerator network $N(\cdot|s)$, we need to have a sampling policy for sampling outcomes $y$ given states $s$ and next states $s'$, which can be achieved by
\begin{equation}
Q(y|s',s) = \frac{r(y)F(s|y)P_F(s'|s,y)}{N(s'|s)}.
\label{eq:Q_def}
\end{equation}
Therefore, we have the following constraint according to Eq.~(\ref{eq:Q_def})
\begin{equation}
N(s'|s) Q(y|s',s) = r(y)F(s|y)P_F(s'|s,y),
\end{equation}
based on which we derive the corresponding loss function by minimizing their difference by training in the log domain, i.e.,
\begin{equation}
\mathcal{L}_{\text amortized} = \left( \log N(s'|s) + \log Q(y|s',s) - \log r(y) - \log F(s|y) - \log P_F(s'|s,y) \right)^2.
\label{eq:amortized}
\end{equation}

\noindent \textbf{Theoretical justification.}
In Proposition~\ref{thm}, we show that $N(s'|s)$ can correctly estimate the summation when it is trained to completion and the distribution of outcomes $y$ has full support. 
\begin{proposition}
Suppose that $\forall (s,s',y)$, $\mathcal{L}_{\text amortized}(s,s',y)=0$, then the amortized predictor $N(s'|s)$ estimates $\sum_y r(y) F(s|y) P_F(s'|s,y)$.
\label{thm}
\end{proposition}
The proof can be found in Appendix~\ref{app:proof_amortize}. Proposition~\ref{thm} justifies the use of the numerator network as an efficient alternative for estimating the computationally intractable summation in Eq.~(\ref{eq:policy}).\\

\noindent \textbf{Empirical validation.} We now investigate the converted/learned sampling policy in the standard GridWorld domain~\citep{bengio2021flow}, which has a multi-modal reward function.
More details about the setup is in Appendix~\ref{app:exp_setup} due to space limitation.
We visualize the last $2\times 10^5$ samples from different baselines including training GFN from scratch, OC-GFN with the Monte Carlo-based estimation with Eq.~(\ref{eq:policy}) and the amortized marginalization approach with Eq.~(\ref{eq:amortized}).

\begin{figure}[!h]
\centering
\subfloat[]{\includegraphics[width=0.115\linewidth]{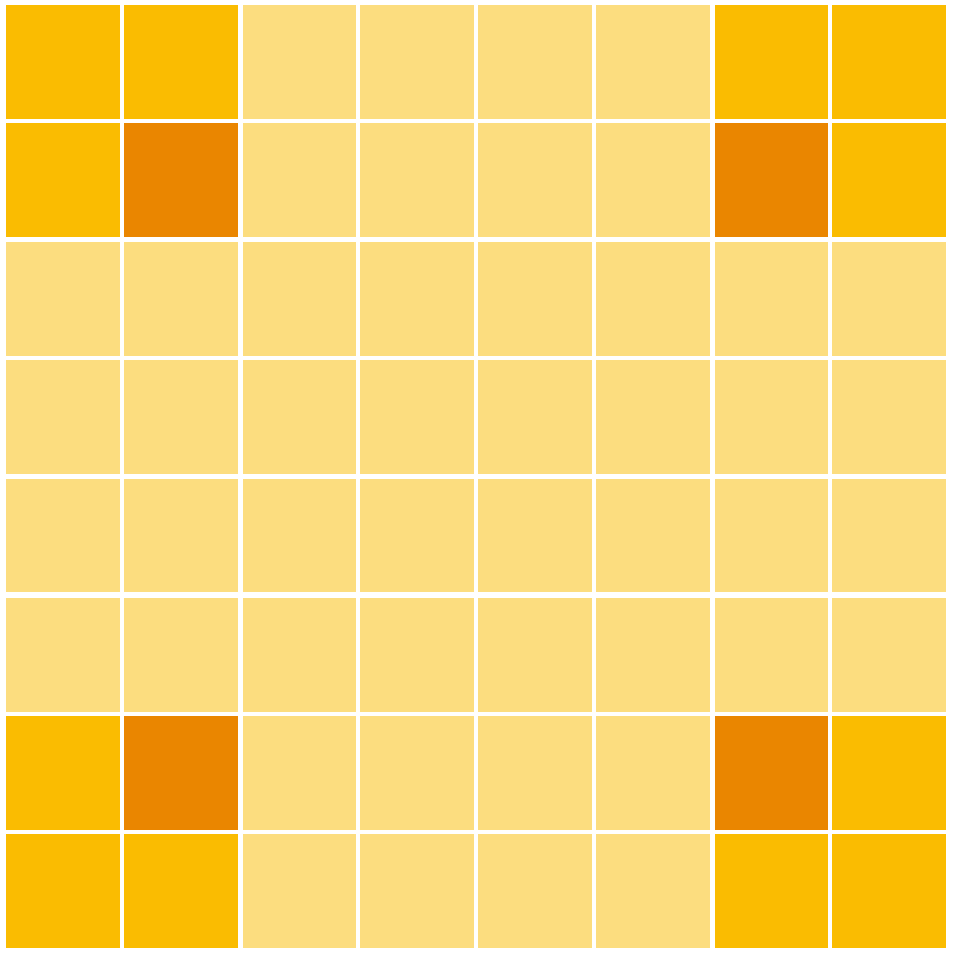}}
\hspace{.3cm}
\subfloat[]{\includegraphics[width=0.162\linewidth]{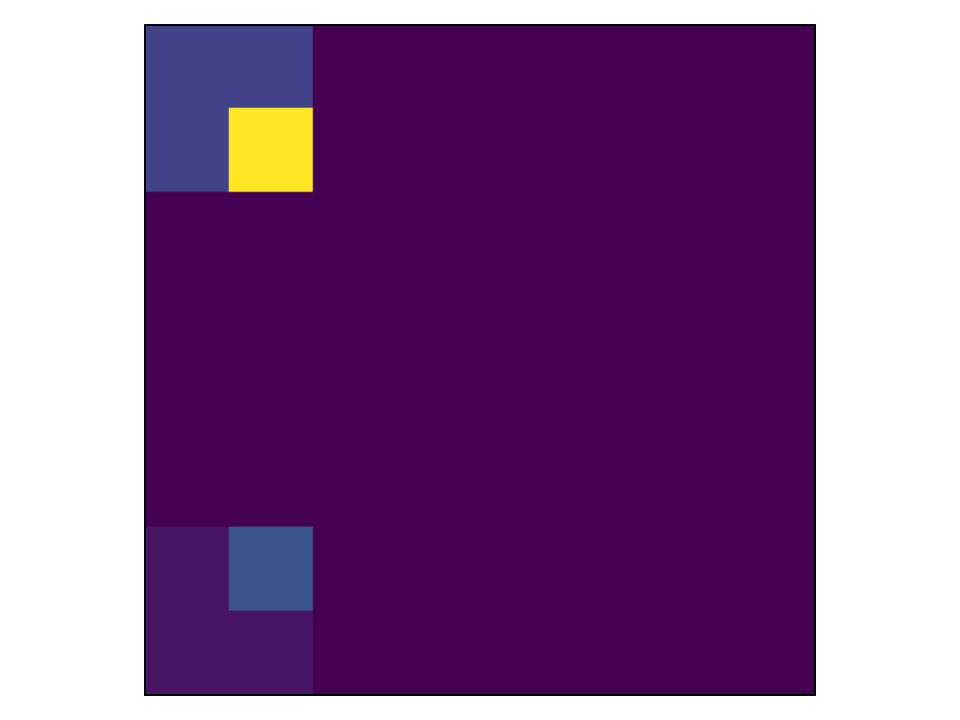}}
\subfloat[]{\includegraphics[width=0.162\linewidth]{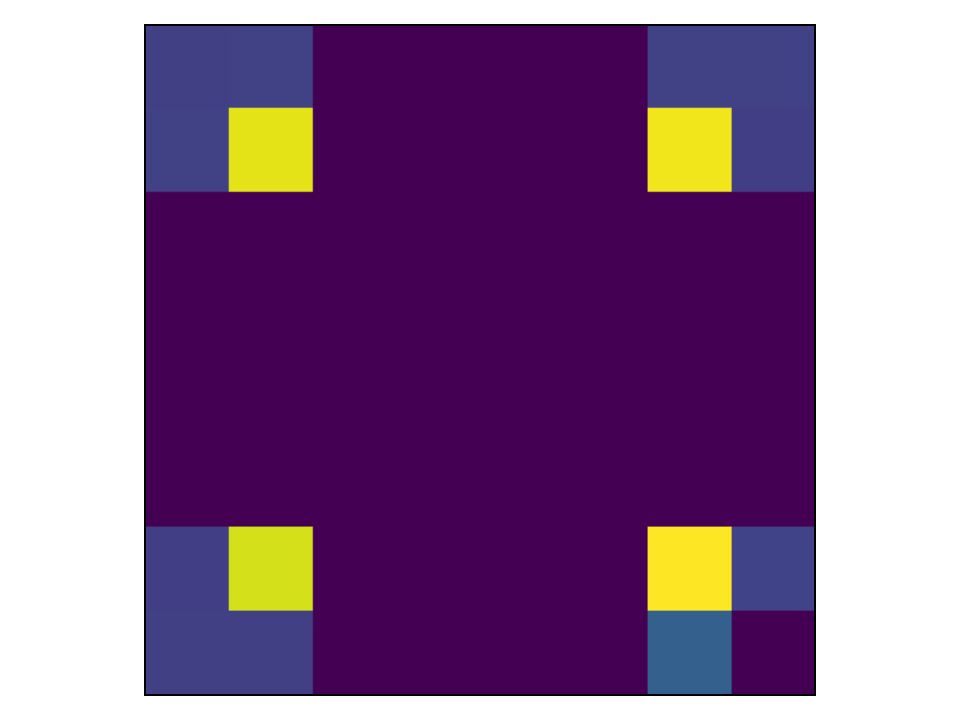}}
\subfloat[]{\includegraphics[width=0.162\linewidth]{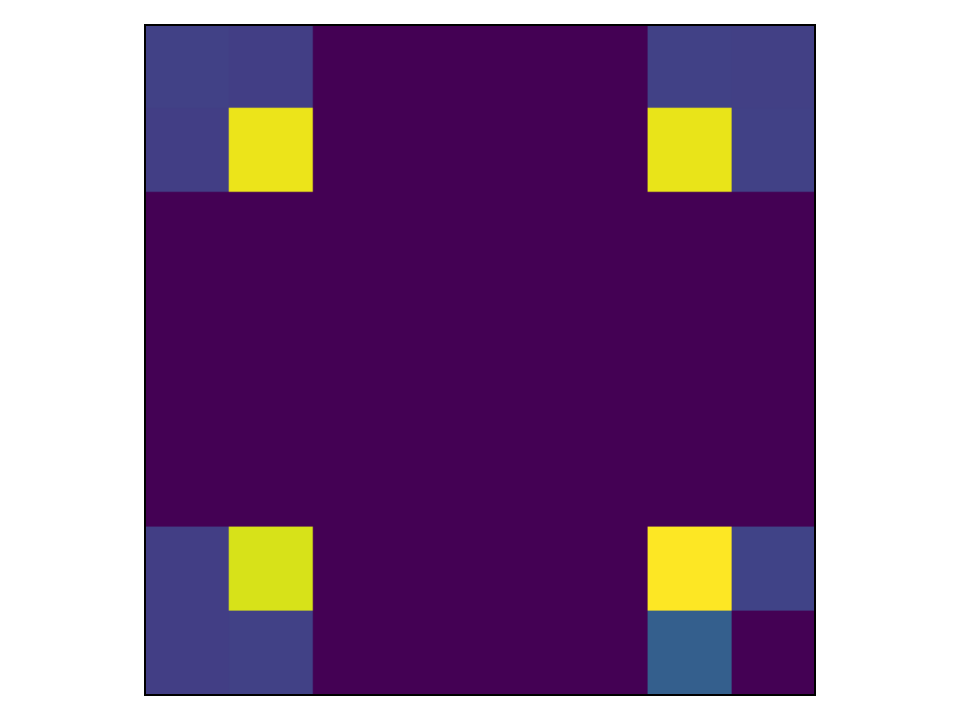}}
\caption{Distribution of $2 \times 10^5$ samples from different baselines. (a) Target distribution. (b) GFN (from scratch). (c) OC-GFN (Monte Carlo-based). (d) OC-GFN (Amortizer-based).
}
\label{fig:grid_dist}
\end{figure}

As shown in Figure~\ref{fig:grid_dist}(b), directly training GFN with trajectory balance~\citep{malkin2022trajectory} can suffer from the mode collapse problem and fail to discover all modes of the target distribution in Figure~\ref{fig:grid_dist}(a). On the contrary, we can directly obtain a policy from the pre-trained OC-GFN model that samples proportionally to the target rewards as shown in Figure~\ref{fig:grid_dist}(c), while fine-tuning OC-GFN with the amortized inference could also match the target distribution as in Figure~\ref{fig:grid_dist}(d), which validates its effectiveness in estimating the marginal.

\noindent \textbf{Practical implementation.}
In practice, we can train the amortized predictor $N(\cdot|s)$ and the sampling policy $Q(\cdot|s',s)$ in a GFlowNet-like procedure. We sample $(s,s')$ with $N$
by interacting with the environment. This can also be realized by sampling from the large and diverse dataset $\mathcal{D}$ obtained in the pre-training stage.
We then sample outcomes $y$ from the sampling policy $Q(\cdot|s',s)$, which can incorporate its tempered version or with $\epsilon$-greedy exploration~\citep{bengio2021flow} for obtaining rich distributions of $y$.
We can then learn the amortized predictor and the sampling policy by one of these sampled experiences. Finally, we can derive the policy that will be converted to downstream tasks by $P_F^{r}(s'|s) = {N(s'|s)}/ {\sum_{s'} N(s'|s)}$. 
The resulting training algorithm is summarized in Algorithm~\ref{alg} and the right part in Figure~\ref{fig:stage2}.

\section{Experiments} \label{sec:exp}
In this section, we conduct extensive experiments to better understand the effectiveness of our approach and aim to answer the following key questions: \textbf{(i)} How do outcome-conditioned GFlowNets (OC-GFN) perform in the reward-free unsupervised pre-training stage? \textbf{(ii)} What is the effect of key modules? \textbf{(iii)} Can OC-GFN transfer to downstream tasks efficiently? 
\textbf{(iv)} Can they scale to complex and practical scenarios like biological sequence design?

\subsection{GridWorld} \label{exp:grid}
We first conduct a series of experiments on GridWorld~\citep{bengio2021flow} to understand the effectiveness of the proposed approach. 
In the reward-free unsupervised pre-training phase, we train a (unconditional) GAFN~\citep{pan2022gafn} and an outcome-conditioned GFN (OC-GFN) on a map without task-specific rewards, where GAFN is trained purely from self-supervised intrinsic rewards according to Algorithm~\ref{alg:stage1}. We investigate how well OC-GFN learns in the unsupervised pre-training stage, while the supervised fine-tuning stage has been discussed in Section~\ref{sec:stage2}.
Each algorithm is run for $3$ different seeds and the mean and standard deviation are reported.
A detailed description of the setup and hyperparameters can be found in Appendix~\ref{app:exp_setup}.

\begin{figure}[!h]
\centering
\subfloat[]{\includegraphics[width=0.16\linewidth]{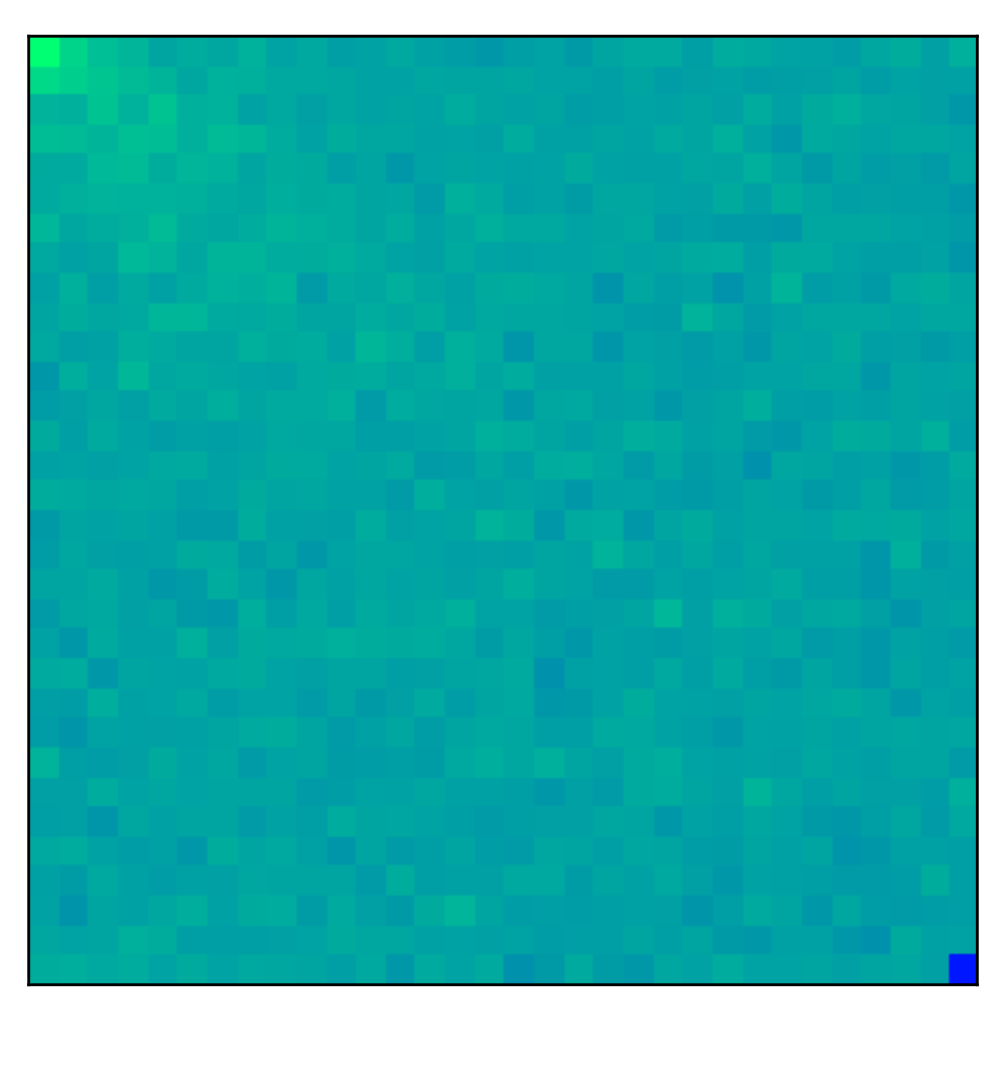}}
\hspace{.1cm}
\subfloat[]{\includegraphics[width=0.23\linewidth]{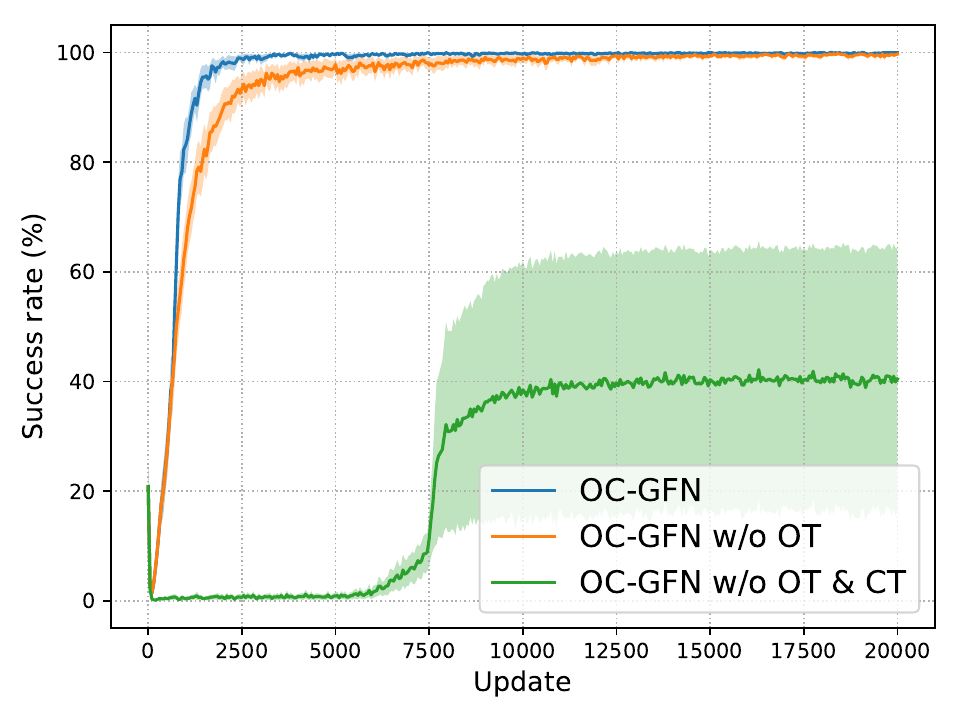}}
\subfloat[]{\includegraphics[width=0.23\linewidth]{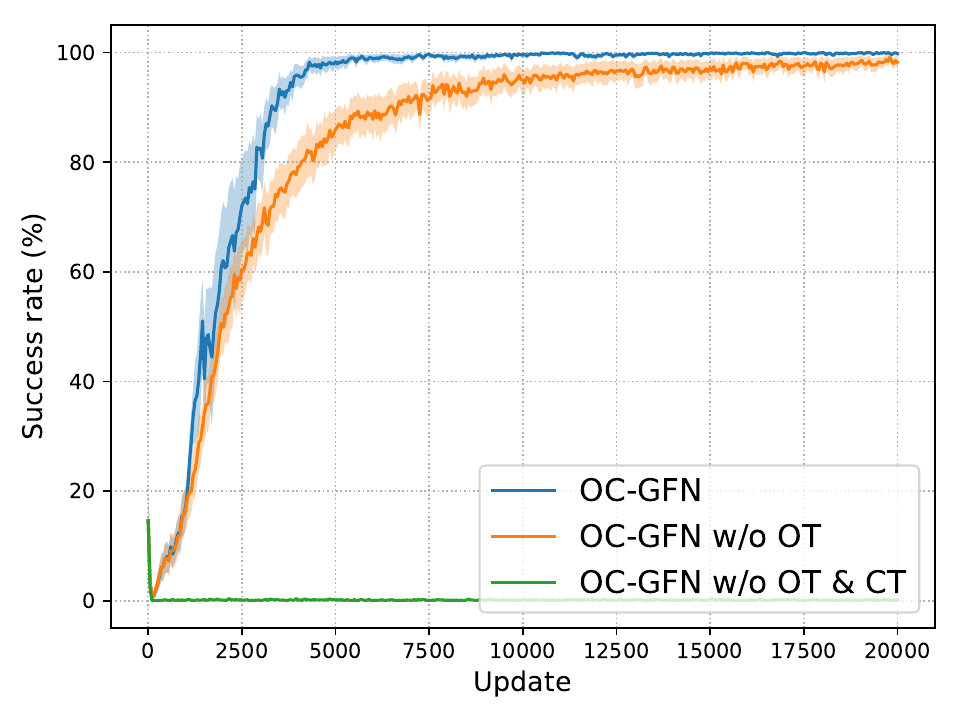}}
\subfloat[]{\includegraphics[width=0.23\linewidth]{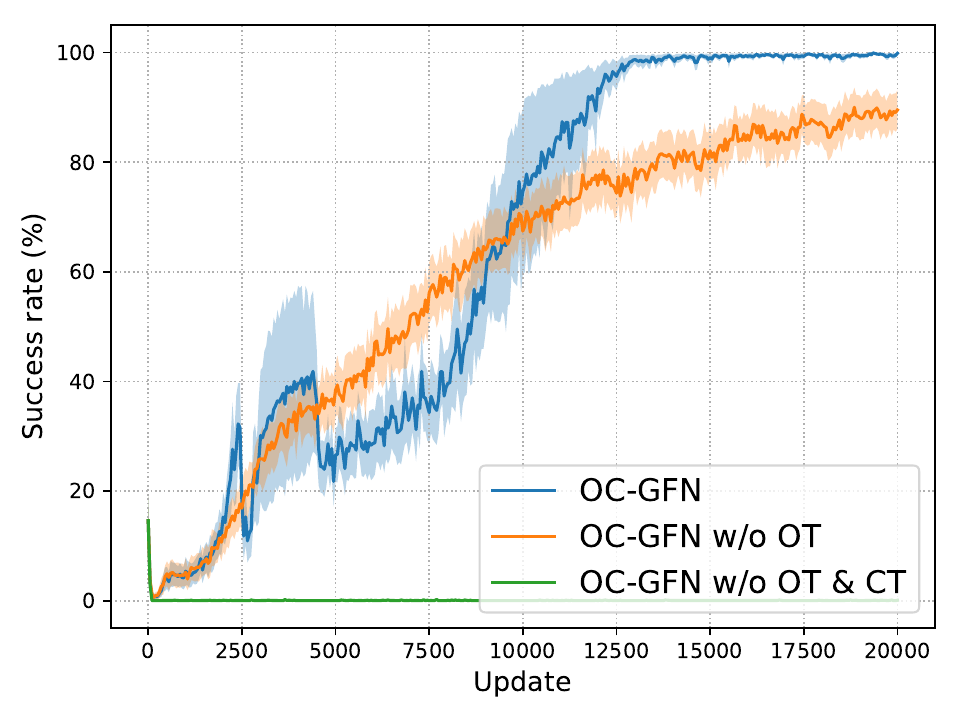}}
\caption{Results in GridWorld in different with different scales of the task. (a) Outcome distribution. (b)-(d) Success rate of OC-GFN and its variants in small, medium, and large maps, respectively.}
\label{fig:grid_succ}
\end{figure}

\textbf{Outcome distribution.} 
We first evaluate the quality of the exploratory data collected by GAFN, as it is essential for training OC-GFN with a rich distribution of outcomes.
We demonstrate the sample distribution from GAFN in Figure~\ref{fig:grid_succ}(a), which has great diversity and coverage, and validates its effectiveness in collecting unlabeled exploratory trajectories for providing diverse target outcomes for OC-GFN to learn from.

\textbf{Outcome reaching performance.}
We then investigate the key designs in pre-training OC-GFN by analyzing its success rate for achieving target outcomes. We also ablate key components to investigate the effect of contrastive training (CT) and outcome teleportation (OT).
The success rates of OC-GFN and its variants are summarized in Figures~\ref{fig:grid_succ}(b)-(d) in different sizes of the map from small to large.
As shown, OC-GFN can successfully learn to reach specified outcomes with a success rate of nearly $100\%$ at the end of training in maps with different sizes.
Disabling the outcome teleportation component (OC-GFN w/o OT) leads to lower sample efficiency, in particular in larger maps. 
Further deactivating the contrastive training process (OC-GFN w/o OT \& CT) fails to learn well as the size of the outcome space grows, since the agent can hardly collect successful trajectories for meaningful updates.
This indicates that both contrastive training and outcome teleportation are important for successfully training OC-GFN in large outcome spaces. It is also worth noting that outcome teleportation can significantly boost learning efficiency in problems with a combinatorial explosion of outcome spaces (e.g., sequence generation in the following sections).

\begin{wrapfigure}{r}{0.3\textwidth} \vspace{-.3in}
\centering
\subfloat{\includegraphics[width=0.5\linewidth]{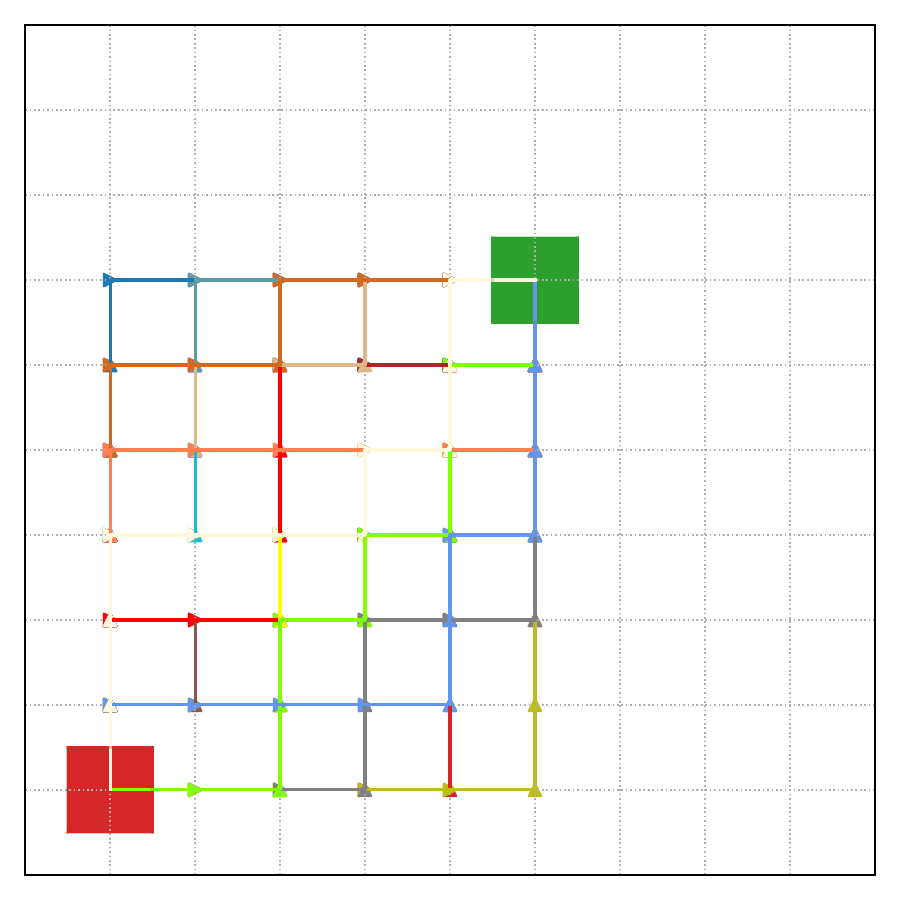}}
\subfloat{\includegraphics[width=0.5\linewidth]{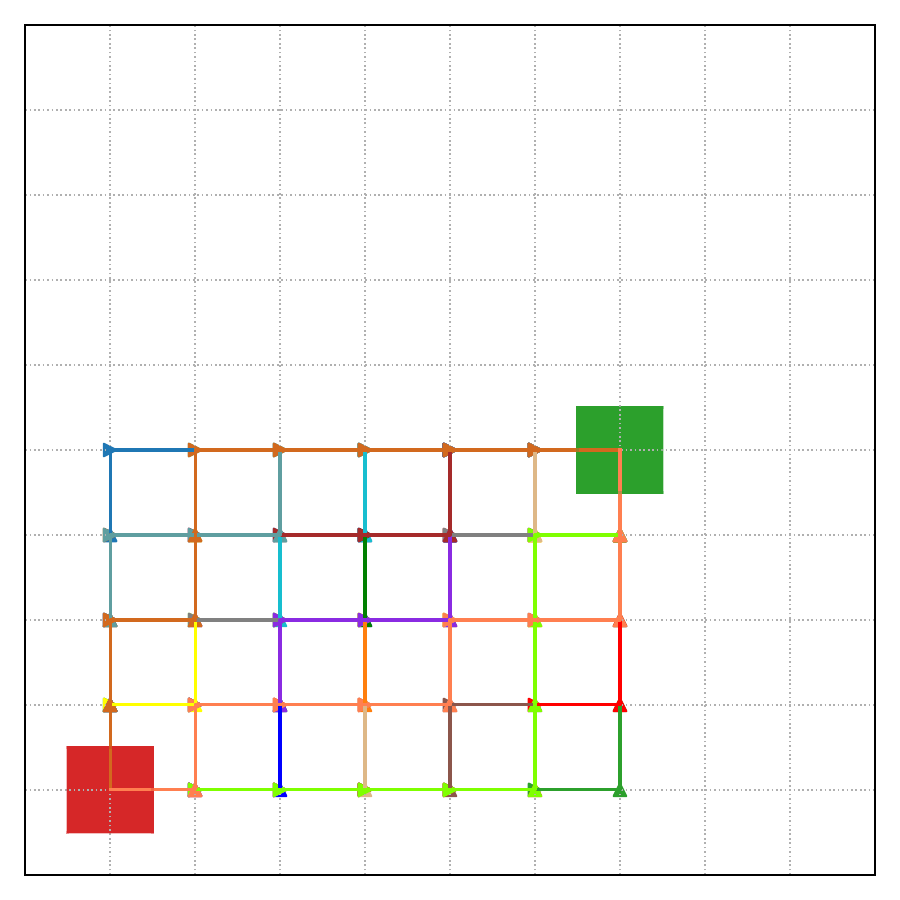}}
\vspace{-.05in}
\caption{Visualization of the behaviors of OC-GFN.}
\vspace{-.15in}
\label{fig:grid_behavior}
\end{wrapfigure}
\textbf{Outcome-conditioned behaviors.}
We now visualize the learned behaviors of OC-GFN in Figure~\ref{fig:grid_behavior}, where the red and green squares denote the starting point and the goal, respectively. We find that OC-GFN can not only reach different specified outcomes with a high success rate, but also discover diverse trajectories
leading to a specified outcome $y$ with the forward policy $P_F$. 
OC-GFN is able to generate diverse trajectories, which is different from typical goal-conditioned reinforcement learning approaches~\citep{schaul2015universal} that usually only learn a single solution to the goal state. This ability can be very helpful for generalizing to downstream tasks with similar structure but subtle changes (e.g., obstacles in the maze)~\citep{kumar2020one}.

\subsection{Bit Sequence Generation} \label{exp:bit}
We study the bit sequence generation task~\citep{malkin2022trajectory}, where the agent generates sequences of length $n$ in a left-to-right manner.
At each time step, the agent appends a $k$-bit ``word" from a vocabulary $V$ to the current state. 
We follow the same procedure for pre-training OC-GFN without task-specific rewards as in Section~\ref{exp:grid} with more details in Appendix~\ref{app:exp_setup}. \\

\begin{figure}[!h]
\centering
\subfloat[Small.]{\includegraphics[width=0.23\linewidth]{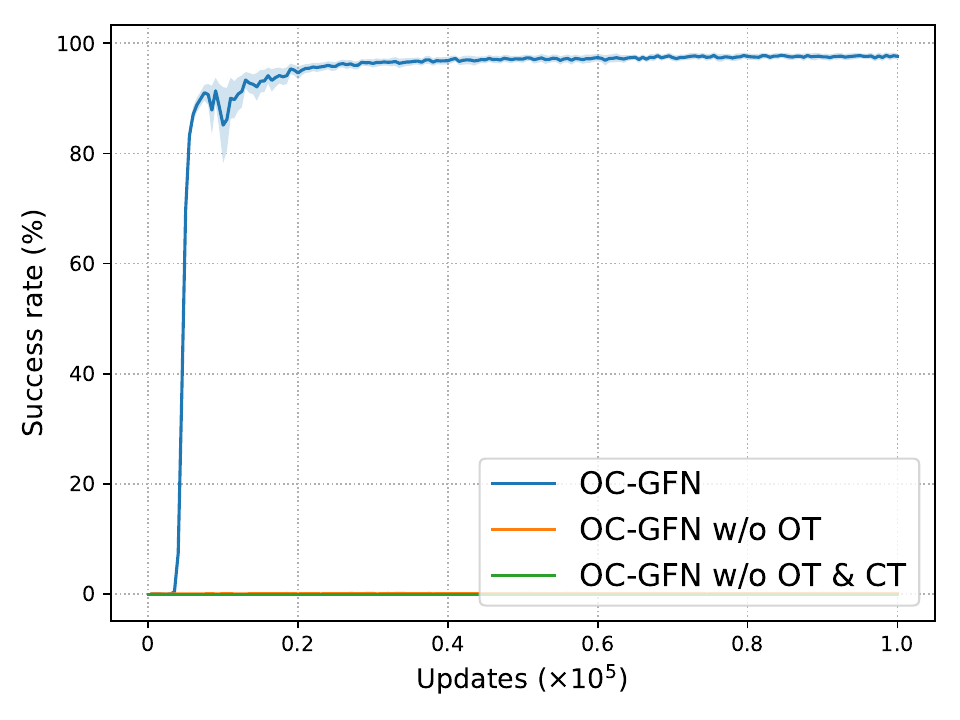}}
\subfloat[Medium.]{\includegraphics[width=0.23\linewidth]{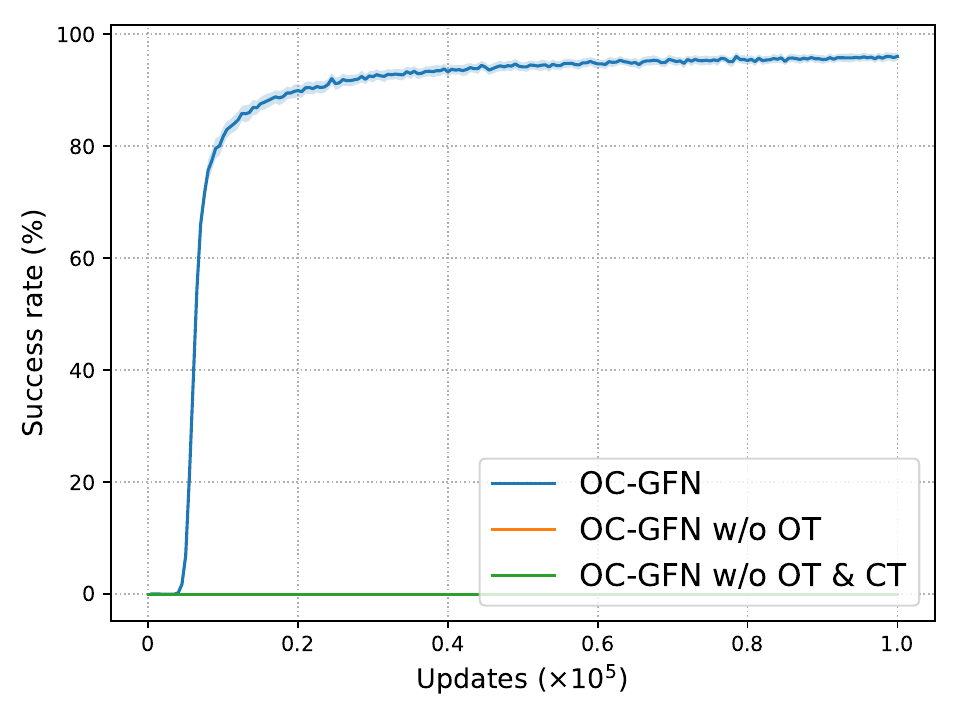}}
\subfloat[Large.]{\includegraphics[width=0.23\linewidth]{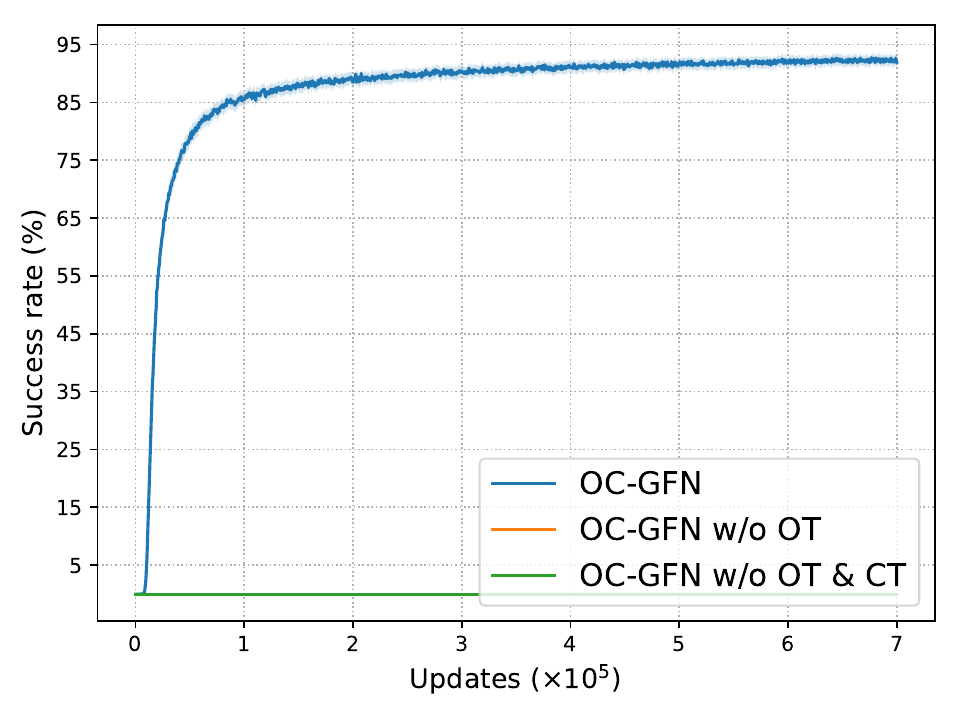}}
\subfloat[Summary.]{
\footnotesize{\def\arraystretch{1.3}\begin{tabular}{cc} & \vspace{-3cm}\\
\hline
\multicolumn{2}{c}{\textbf{Size} \quad \quad \textbf{Success rate}} \\ \hline
Small & $98.37\pm0.33$\%\\ 
Medium & $97.22\pm1.70$\%\\ 
Large & $93.75\pm2.04$\% \\ 
\hline
\end{tabular}}
}
\vspace{-.05in}
\caption{Success rates in the bit sequence generation task with different scales of the task.}
\label{fig:bit_succ}
\end{figure}

\noindent {\textbf{Analysis of the unsupervised pre-training stage.}}
We first analyze how well OC-GFN learns in the unsupervised pre-training stage by investigating its success rate for achieving specified outcomes in bit sequence generation problems with different scales including small, medium, and large.
As shown in Figures~\ref{fig:bit_succ}(a)-(c), OC-GFN can successfully reach outcomes, leading to a high success rate as summarized in Figure~\ref{fig:bit_succ}(d).
It is also worth noting that it fails to learn when the outcome teleportation module is disabled due to the particularly large outcome spaces. In such highly challenging scenarios, the contrastive training paradigm alone does not lead to efficient learning.

\begin{wrapfigure}{r}{0.25\textwidth} \vspace{-.18in}
\centering
{\includegraphics[width=1.\linewidth]{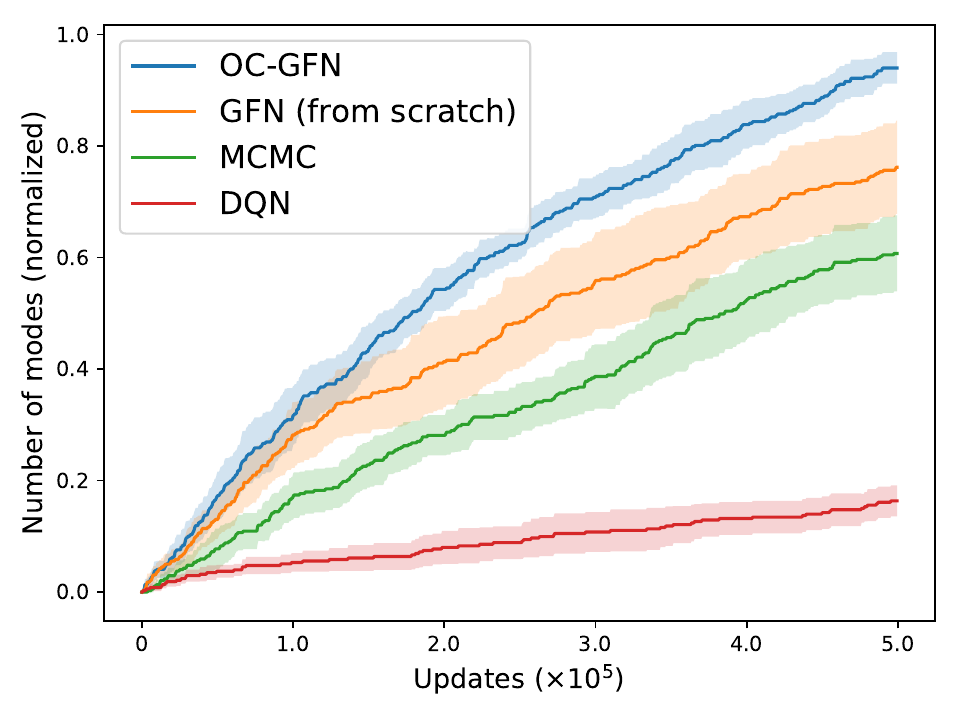}}
\vspace{-.27in}
\caption{Results in bit sequence generation.}
\vspace{-.25in}
\label{fig:bit_res_avg}
\end{wrapfigure}
\noindent {\textbf{Analysis of the Supervised Fine-Tuning Stage.}}
After we justified the effectiveness of training OC-GFN in the unsupervised pre-training stage, 
we study whether fine-tuning the model with the amortized approach can enable faster mode discovery when adapting to downstream tasks (as described in Appendix~\ref{app:exp_setup}). 
We compare the proposed approach against strong baselines including training a GFN from scratch~\citep{bengio2023foundations}, Metropolis-Hastings-MCMC~\citep{dai2020learning}, and Deep Q-Networks (DQN)~\citep{mnih2015human}.
We evaluate each method in terms of the number of modes discovered during the course of training as in previous works~\citep{malkin2022trajectory}.
We summarize the normalized (between $0$ and $1$ to facilitate comparison across tasks) number of modes averaged over each downstream task in Figure~\ref{fig:bit_res_avg}, while results for each individual downstream task can be found in Appendix~\ref{app:bit} due to space limitation.

We find that OC-GFN significantly outperforms baselines in mode discovery in both efficiency and performance, while DQN gets stuck and does not discover diverse modes due to the reward-maximizing nature, and MCMC fails to perform well in problems with larger state spaces. 

\subsection{TF Bind Generation} \label{sec:exp_tfb8}

We study a more practical task of generating DNA sequences with high binding activity with targeted transcription factors~\citep{jain2022biological}.
The agent appends a symbol from the vocabulary to the 
current state (a partially constructed sequence) at each step.
We consider $30$ different downstream 
tasks studied in~\citep{barrera2016survey}, which conducted biological experiments to determine the binding properties between a range of transcription factors and every conceivable DNA sequence.

\begin{wrapfigure}{r}{0.25\textwidth} \vspace{-.35in}
\centering
\subfloat{\includegraphics[width=1.\linewidth]{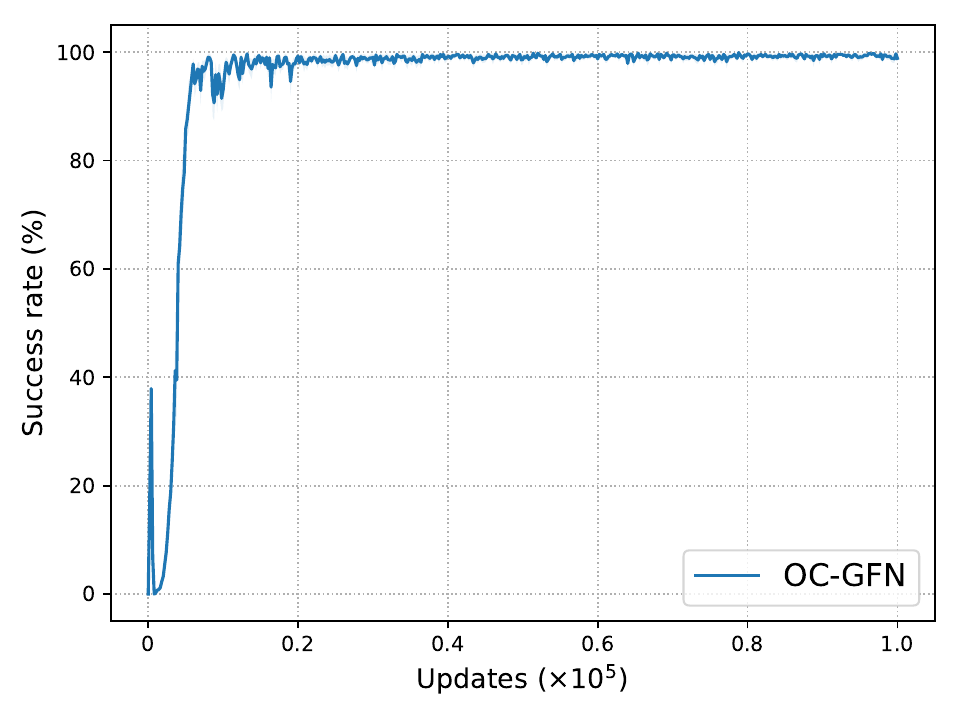}}
\vspace{-.1in}
\caption{
The success rate of OC-GFN in TF Bind generation. 
}
\label{fig:tfb_stage1}
\vspace{-.2in}
\end{wrapfigure}
\noindent {\textbf{Analysis of the unsupervised pre-training stage.}}
We analyze how well OC-GFN learns by evaluating its success rate. 
Figure~\ref{fig:tfb_stage1} shows that OC-GFN can successfully achieve the targeted goals with a success rate of nearly $100\%$ in the practical task of generating DNA sequences.

\noindent {\textbf{Analysis of the supervised fine-tuning stage.}}
We investigate how well OC-GFN can transfer to downstream tasks for generating TF Bind 8 sequences.
We tune the hyperparameters on the task \texttt{PAX3\_R270C\_R1}, and evaluate the baselines on the other $29$ downstream tasks.
We investigate the learning efficiency and performance in terms of the number of modes and the top-$K$ scores during the course of training.
The results in $2$ downstream tasks are shown in Figure~\ref{fig:tfb_res}, with 
the full results in Appendix~\ref{app:tfb} due to space limitation.
Figures~\ref{fig:tfb_res}(a)-(b) illustrate the number of discovered modes while Figures~\ref{fig:tfb_res}(c)-(d) shows the top-$K$ scores.
We find that the proposed approach
discovers more diverse modes faster and achieves higher top-$K$ ($K=100$) scores compared to baselines.
We summarize the rank of each baseline in all $30$ downstream tasks~\citep{zhang2021unifying} in Appendix~\ref{app:tfb}, where OC-GFN ranks highest compared with other methods.
We further visualize in Figure~\ref{fig:tfb_res}(e) the t-SNE plots 
of the TF Bind 8 sequences discovered by transferring OC-GFN with the amortized approach and the more expensive but competitive training of a GFN from scratch.
As shown, training a GFN from scratch only focuses on limited regions while the OC-GFN has a greater coverage.

\begin{figure}[!h]
\centering
\subfloat[]{\includegraphics[width=0.2\linewidth]{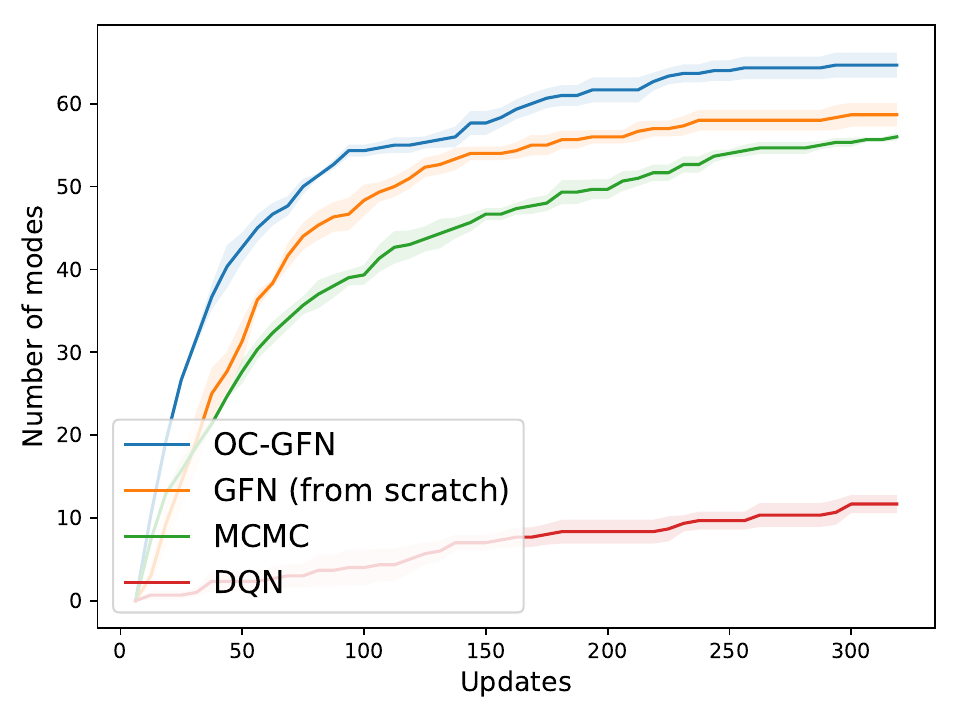}}
\subfloat[]{\includegraphics[width=0.2\linewidth]{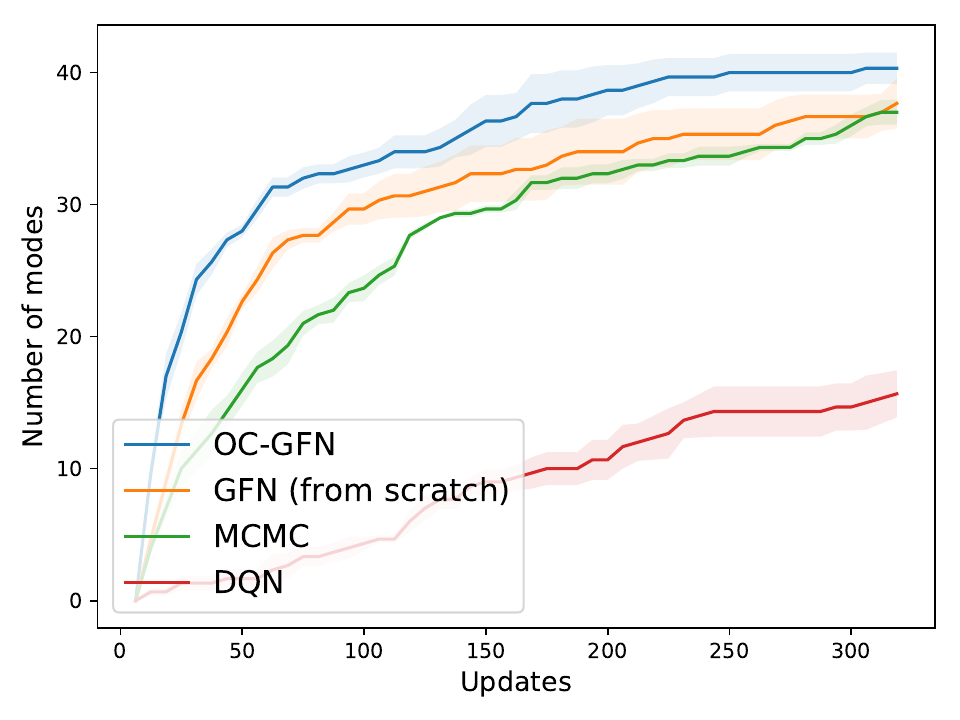}}
\subfloat[]{\includegraphics[width=0.2\linewidth]{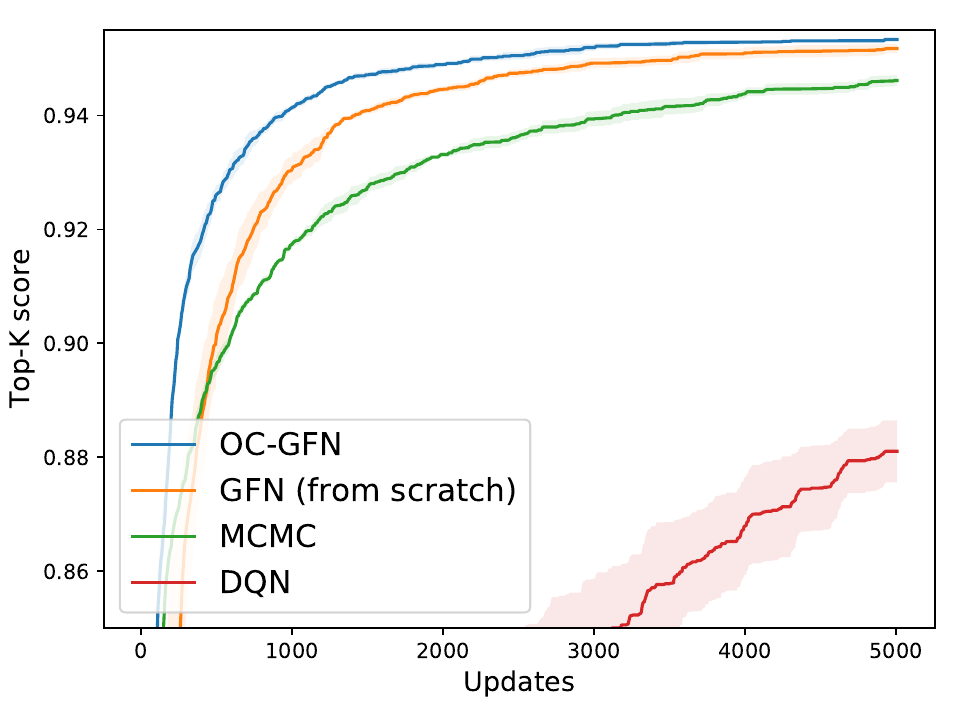}}
\subfloat[]{\includegraphics[width=0.2\linewidth]{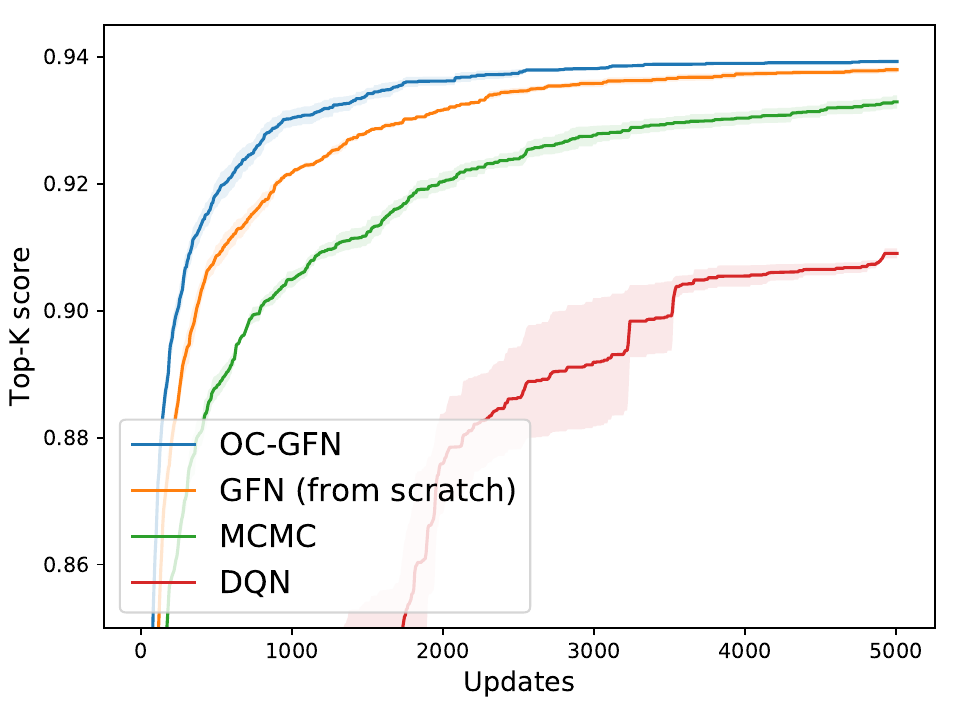}}
\subfloat[]{\includegraphics[width=0.22\linewidth]{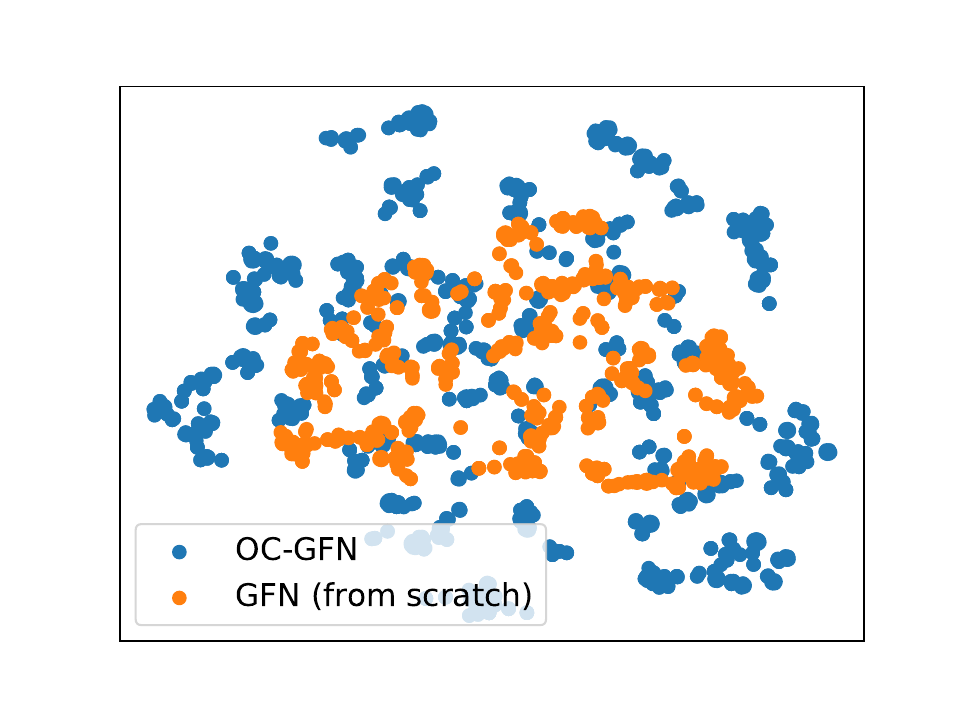}}
\vspace{-.1in}
\caption{Results in the TF Bind sequence generation task in different downstream tasks.}
\label{fig:tfb_res}
\end{figure}

\vspace{-.08in}
\begin{wrapfigure}{r}{0.42\textwidth} \vspace{-.3in}
\centering
\subfloat{\includegraphics[width=0.5\linewidth]{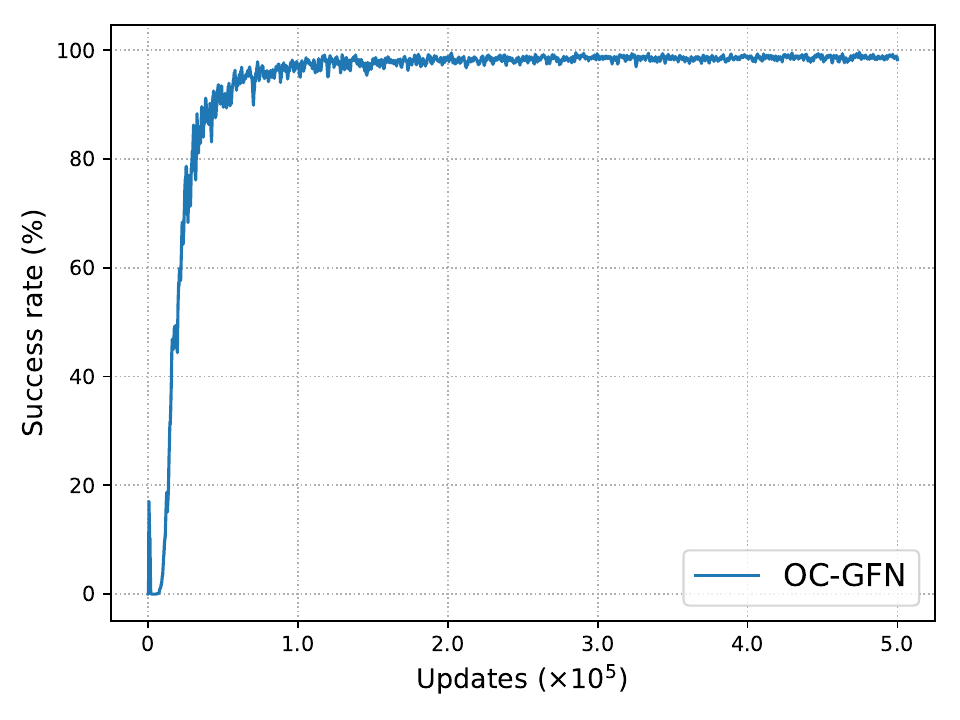}}
\subfloat{\includegraphics[width=0.5\linewidth]{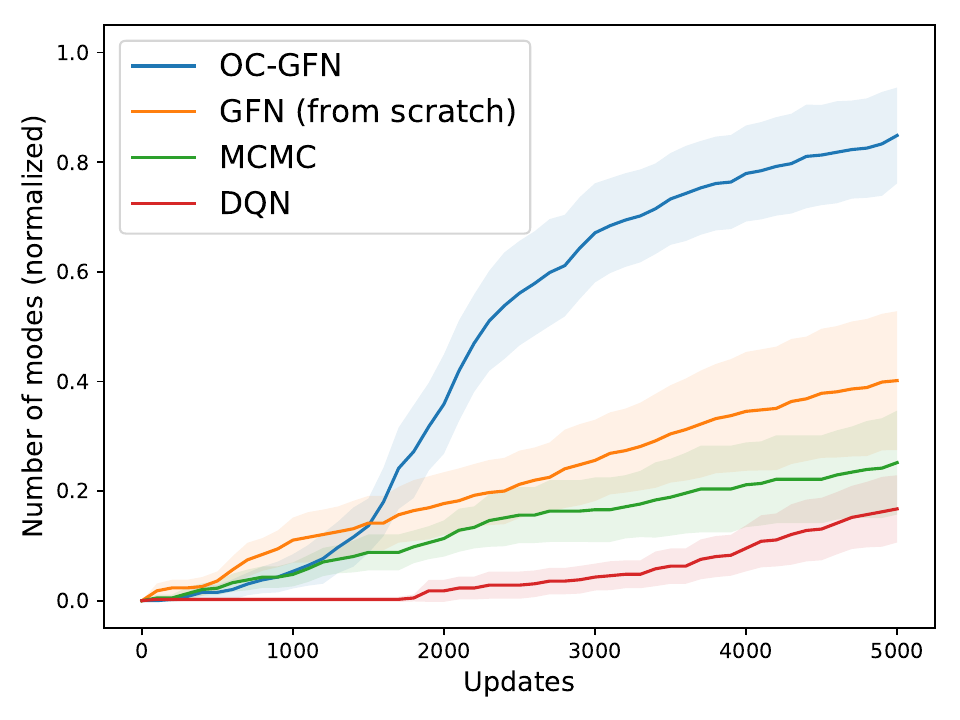}}
\vspace{-.1in}
\caption{Results in RNA generation. \textit{Left}: success rate. \textit{Right}: number of modes (normalized).}
\label{fig:rna_succ}
\vspace{-.25in}
\end{wrapfigure}
\subsection{RNA Generation}
We study the performance of the proposed method in a larger task of generating RNA sequences that bind to a given target introduced in~\citep{lorenz2011viennarna}. 
We follow the same procedure as in Section~\ref{sec:exp_tfb8}, with additional details provided in Appendix~\ref{app:exp_setup}.
We consider four different downstream tasks from ViennaRNA~\citep{lorenz2011viennarna}, each considering the binding energy with a different target as a reward. 

The left part in Figure~\ref{fig:rna_succ} shows that OC-GFN achieves a success rate of almost $100\%$ at the end of the pre-training stage.
The right part in Figure~\ref{fig:rna_succ} summarizes the averaged normalized (between $0$ and $1$ to facilitate comparison across tasks) 
performance in terms of the number of modes averaged over four downstream tasks, where the performance for each individual task is shown in Appendix~\ref{app:rna}.
We observe that OC-GFN is able to achieve much higher diversity than baselines, indicating that the pre-training phase enables the OC-GFN to explore the state space much more efficiently. 

\subsection{Antimicrobial Peptide Generation}
\begin{wrapfigure}{r}{0.42\textwidth} \vspace{-.3in}
\centering
\subfloat[Success rate.]{\includegraphics[width=0.5\linewidth]{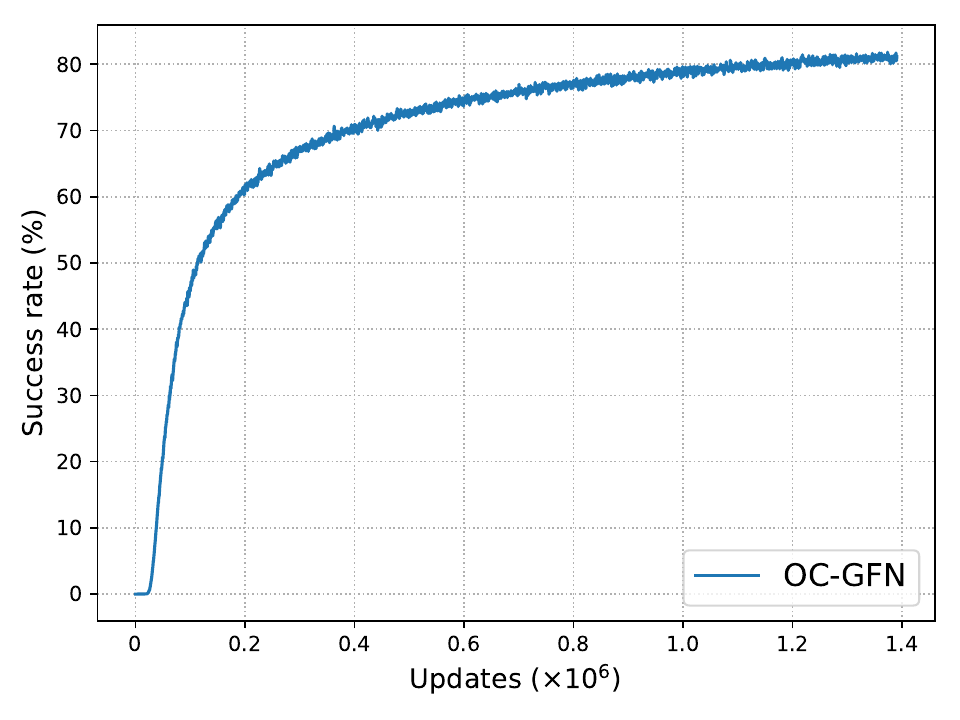}} 
\subfloat[Number of modes.]{\includegraphics[width=0.5\linewidth]{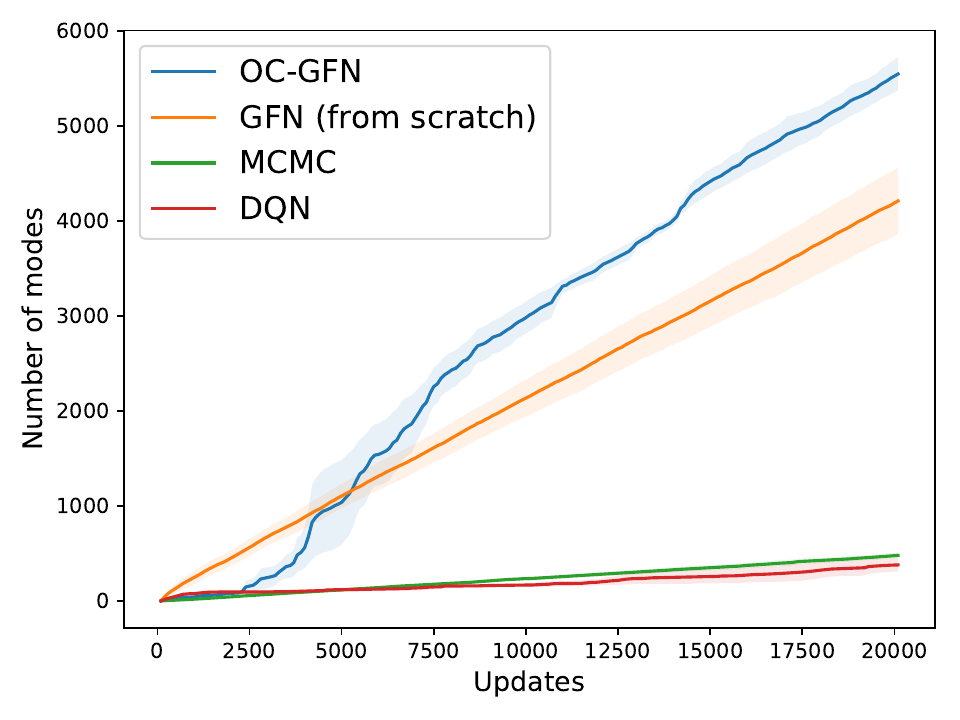}} 
\vspace{-.12in}
\caption{Results in AMP generation.}
\vspace{-.2in}
\label{fig:amp_res}
\end{wrapfigure}
To demonstrate the scalability of our approach to even more challenging and complex scenarios, we also evaluate it on the biological task of generating antimicrobial peptides (AMP) with lengths $50$~\citep{jain2022biological} (resulting in an outcome space of $20^{50}$ candidates).

Figure~\ref{fig:amp_res}(a) demonstrates the success rate of OC-GFN for achieving target outcomes, which can still reach a high success rate in particularly large outcome spaces with our novel training paradigm according to contrastive learning and fast goal propagation.
Figure~\ref{fig:amp_res}(b) shows the number of modes discovered, where OC-GFN provides consistent improvements, which validates the efficacy of OC-GFN to successfully scale to much larger-scale tasks. 

\section{Conclusion}
In this paper, we propose a novel method for unsupervised pre-training of GFlowNets through an outcome-conditioned GFlowNet, coupled with a new approach to efficiently fine-tune the pre-trained model for downstream tasks.
Our work opens the door for GFlowNets to be pre-trained for fine-tuning for downstream tasks by leveraging learned shared structure.
Empirical results on the standard GridWorld domain validate the effectiveness of the proposed approach in successfully achieving targeted outcomes and the efficiency of the amortized predictor. 
We also conduct extensive experiments in the more complex and challenging biological sequence design tasks to demonstrate its practical scalability. The proposed method greatly improves learning performance compared with strong baselines including training a GFlowNet from scratch, particularly in tasks which are challenging for GFlowNets to learn. 

\bibliography{main}
\bibliographystyle{plainnat}

\clearpage
\appendix
\section{Proofs} \label{app:a}
\subsection{Proof of Proposition 4.1} \label{app:proof_oc_gfn}
\textbf{Proposition 4.1}
\emph{
If $\mathcal{L}_{\text{OC-GFN}}(\tau,y)=0$ for all trajectories $\tau$ and outcomes $y$, then the outcome-conditioned forward policy $P_F(s'|s,y)$ can successfully reach any target outcome $y$.
}

\begin{proof}

As $\mathcal{L}_{\text{OC-GFN}}(\tau,y)=0$ is satisfied for all trajectories $\tau=\{s_0, \cdots, s_n=x\}$ and outcomes $y$, we have that
\begin{equation}
F(s_0|y)\prod_{t=0}^{n-1}P_F(s_{t+1}|s_t,y)=\prod_{t=0}^{n-1}R(x|y) P_B(s_{t+1}|s_t,y).
\end{equation}

Since $R(x|y)$ is either $1$ or $0$ in outcome-condition tasks, we get that
\begin{equation}
F(s_0|y)\prod_{t=0}^{n-1}P_F(s_{t+1}|s_t,y)=R(x|y) \prod_{t=0}^{n-1} P_B(s_{t+1}|s_t,y).
\label{eq:oc_traj}
\end{equation}

Then, the probability of reaching the target outcome $y$ is
\begin{equation}
P(x= y|y)=\sum_{\tau,s_n=y}P(\tau|y).
\end{equation}

By definition, we have that
\begin{equation}
P(\tau|y)=\prod_{t=0}^{n-1}P_F(s_{t+1}|s_t,y).
\label{eq:p_def}
\end{equation}

Therefore, we get that
\begin{equation}
P(x=y|y)=\sum_{\tau,s_n=y}\prod_{t=0}^{n-1}P_F(s_{t+1}|s_t,y).
\end{equation}

Combing Eq.~(\ref{eq:oc_traj}) with Eq.~(\ref{eq:p_def}), and due to the law of total probability, we obtain that
\begin{equation}
F(s_0|y)P(x=y|y)=\sum_{\tau,s_n=y}\prod_{t=0}^{n-1} P_B(s_{t+1}|s_t,y)=1.
\label{eq:x_eq_y}
\end{equation}

With the same analysis for the case where the agent fails to reach the target outcome $y$, i.e., $x \neq y$, we have that
\begin{equation}
\forall x\neq y, \quad F(s_0|y)P(x|y)=0.
\label{eq:x_neq_y}
\end{equation}

Combing Eq.~(\ref{eq:x_eq_y}) with Eq.~(\ref{eq:x_neq_y}), we have that
$P(x=y|y)=1$, i.e., the outcome-conditioned forward policy $P_F(s'|s, y)$ can successfully reach any target outcome $y$.

\end{proof}

\subsection{Analysis of the Conversion Policy} \label{app:thm_convert_policy}
We now elaborate on more details about the effect of Eq.~(\ref{eq:policy}) in the text based on~\citep{bengio2023foundations}. 

When the outcome-conditioned GFlowNet (OC-GFN) is trained to completion, the following flow consistency constraint in the edge level is satisfied for intermediate states. 
\begin{equation}
F(s|y)P_F(s'|s,y)=F(s'|y)P_B(s|s',y)
\end{equation}

We define the state flow function as $F^r(s)=\sum_{y}r(y)F(s|y)$, and the backward policy as
\begin{equation}
P_B^r(s|s')=\frac{\sum_{y}r(y)F(s'|y)P_B(s|s',y)}{\sum_{y}r(y)F(s'|y)},
\end{equation}
while the forward policy is defined in Eq.~(\ref{eq:policy}), i.e.,
\begin{equation}
P_F^{r}(s'|s) = \frac{\sum_y r(y) F(s|y) P_F(s'|s,y)}{\sum_y r(y) F(s|y)}
\end{equation}

Then, we have 
\begin{equation}
F^r(s) P_F^r(s'|s)=\sum_{y} r(y) F(s|y) P_F(s'|s,y),
\end{equation}
and
\begin{equation}
F^r(s') P_B^r(s|s')=\sum_{y} r(y) F(s'|y) P_B(s|s',y).
\end{equation}

Combining the above equations, we have that
\begin{equation}
F^r (s)P_F^r(s'|s)=F(s')^r P_B^r(s|s'),
\end{equation}
which corresponds to a new flow consistency constraint in the edge level.
A more detailed proof can be found in~\citep{bengio2023foundations}.

\paragraph{Discussion about applicability} It is also worth noting that OC-GFN should be built upon the detailed balance objective~\citep{bengio2023foundations} discussed in Section~\ref{sec:prelim} or other variants which learn flows (e.g., sub-trajectory balance~\citep{madan2022learning}, flow matching~\citep{bengio2021flow}). Instead, the trajectory balance objective~\citep{malkin2022trajectory}, does not learn a state flow function necessary for converting the pre-trained OC-GFN model to the new policy $\pi_{r}$. 

\subsection{Proof of Proposition 4.2} \label{app:proof_amortize}
\textbf{Proposition 4.2}
\emph{
Suppose that $\forall (s,s',y)$, $\mathcal{L}_{\text amortized}(s,s',y)=0$, then the amortized predictor $N(s'|s)$ estimates $\sum_y r(y) F(s|y) P_F(s'|s,y)$.
}

\begin{proof}
As $\mathcal{L}_{\text{amortized}}(s,s',y)=0$ for all states $s$, next states $s'$, and outcomes $y$, we have that
\begin{equation}
\forall s,s',y, \quad N(s'|s)Q(y|s',s)=r(y)F(s|y)P_F(s'|s,y).
\end{equation}

Therefore, by summing over all possible outcomes $y$ on both sides, we obtain that 
\begin{equation}
\forall s, s', \quad \sum_{y}N(s'|s)Q(y|s',s)=\sum_{y}r(y)F(s|y)P_F(s'|s,y).
\end{equation}

As $\sum_{y}Q(y|s',s)=1$, we get that
\begin{equation}
\forall s, s', \quad N(s'|s)=\sum_{y}r(y)F(s|y) P_F(s'|s,y).
\end{equation}

Based on the above analysis, we get that the amortized predictor $N(s'|s)$ estimates the marginal $\sum_y r(y) F(s|y) P_F(s'|s,y)$.

\end{proof}

\section{Experimental Details} \label{app:b}
All baseline methods are implemented based on the open-source implementation,\footnote{\url{https://github.com/GFNOrg/gflownet}} where we follow the default hyperparameters and setup as in~\citep{bengio2021flow}. The code will be released upon publication of the paper.

\subsection{Experimental Setup} \label{app:exp_setup}

\paragraph{GridWorld}
We employ the same standard reward function for GridWorld from~\citep{bengio2021flow} as in Eq.~(\ref{eq:grid_r}), where the target reward distribution is shown in Figure~\ref{fig:grid_dist}(a) with $4$ modes located near the corners of the maze. 
\begin{equation}
R(x)=\frac{1}{2} \prod_i \mathbb{I}\left(0.25<\left|x_i / H-0.5\right|\right)+ 2 \prod_i \mathbb{I}\left(0.3<\left|x_i / H-0.5\right|<0.4\right) + 10^{-6}.
\label{eq:grid_r}
\end{equation}

The GFlowNet model is a feedforward network consisting of two hidden layers with $256$ hidden units per layer using LeakyReLU activation. We use a same network structure for the outcome-conditioned GFlowNet model, except that we concatenate the state and outcome as input to the network. For the unconditional GAFlowNet model, we follow~\citep{pan2022gafn} and leverage random network distillation as the intrinsic reward mechanism.
The coefficient for intrinsic rewards is set to be $1.0$ (which purely learns from intrinsic motivation without task-specific rewards).
We also use a same network structure for the amortized predictor network.
We train all models with the Adam~\citep{kingma2014adam} optimizer (learning rate is $0.001$) based on samples from a parallel of $16$ rollouts in the environment. 

\paragraph{Bit Sequence Generation}
We consider generating bit sequences with fixed lengths based on~\citep{malkin2022trajectory}.
The GFlowNet model is a feedforward network that consists of $2$ hidden layers with $2048$ hidden units with ReLU activation.
The exploration strategy is $\epsilon$-greedy with $\epsilon=0.0005$, while we set the sampling temperature to $1.0$, and use a reward exponent of $3$.
The learning rate for training the GFlowNet model is $5 \times 10^{-3}$ with the Adam optimizer, with a batch size of $16$.
We use a same network structure for the outcome-conditioned GFlowNet model and the amortized predictor network. 
We train all models for $50000$ iterations, using a parallel of $16$ rollouts in the environment.

\paragraph{TF Bind and RNA Generation}
For the TFBind-8 generation task, we follow the same setup as in~\citep{jain2022biological}.
The vocabulary consists of $4$ nucleobases, and the trajectory length is $8$ and $14$.
The GFlowNet model is a feedforward network that consists of $2$ hidden layers with $2048$ hidden units and ReLU activation.
The exploration strategy is $\epsilon$-greedy with $\epsilon=0.001$, while the reward exponent is $3$.
The learning rate for training the GFlowNet model is $1e-4$, with a batch size of $32$.
We train all models for $5000$ iterations.

\paragraph{AMP Generation}
We basically follow the same setup for the antimicrobial peptide generation task as in \citet{malkin2022trajectory,jain2022biological}, where we consider generating AMP sequence with length $50$.
The GFlowNet model is an MLP that consists of $2048$ hidden layers with $2$ hidden units with ReLU activation.
The exploration strategy is $\epsilon$-greedy with $\epsilon=0.01$, while the sampling temperature is set to $1$, and uses a reward exponent of $3$.
The lr for training the GFlowNet model is $0.001$, and the batch size is $16$.

\subsection{Additional Results on the Bit Sequence Generation Task} \label{app:bit}
\begin{figure}[!h]
\centering
\subfloat[]{\includegraphics[width=0.2\linewidth]{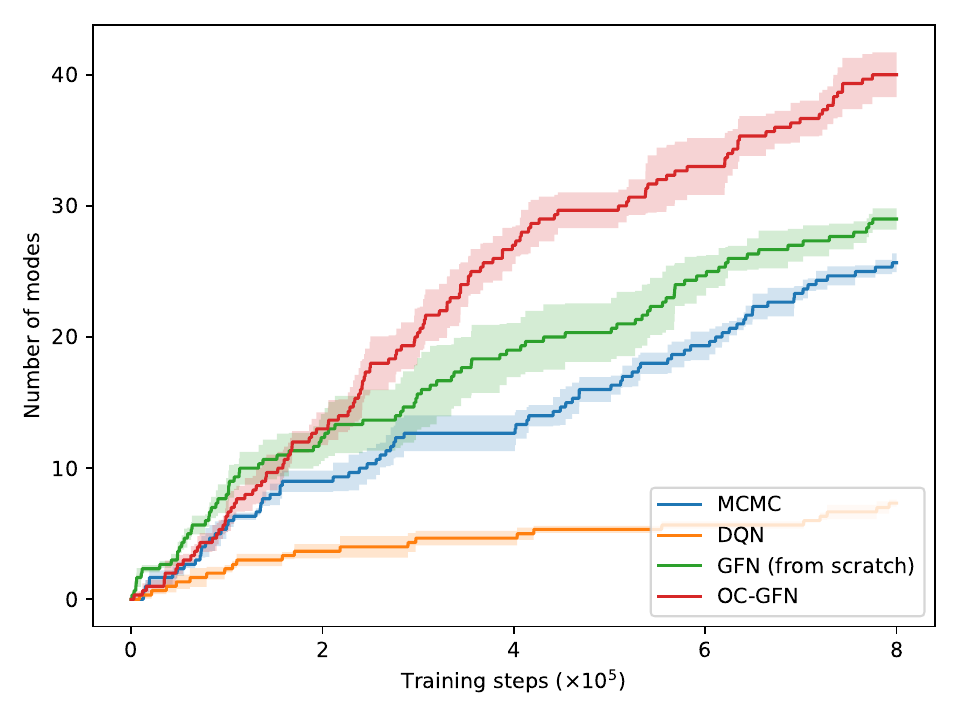}}
\subfloat[]{\includegraphics[width=0.2\linewidth]{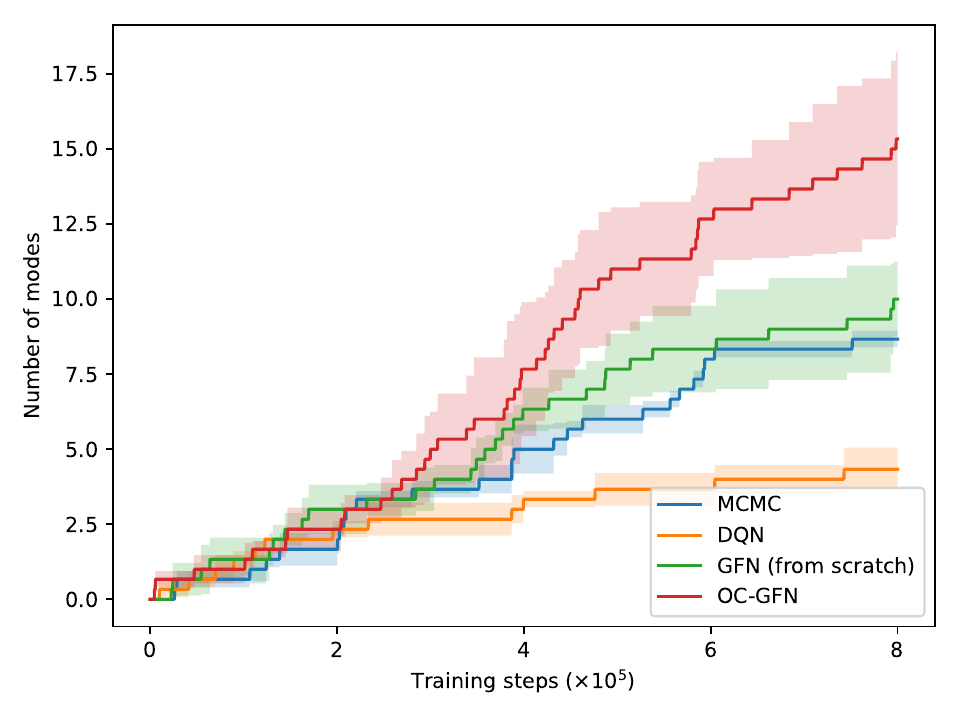}}
\subfloat[]{\includegraphics[width=0.2\linewidth]{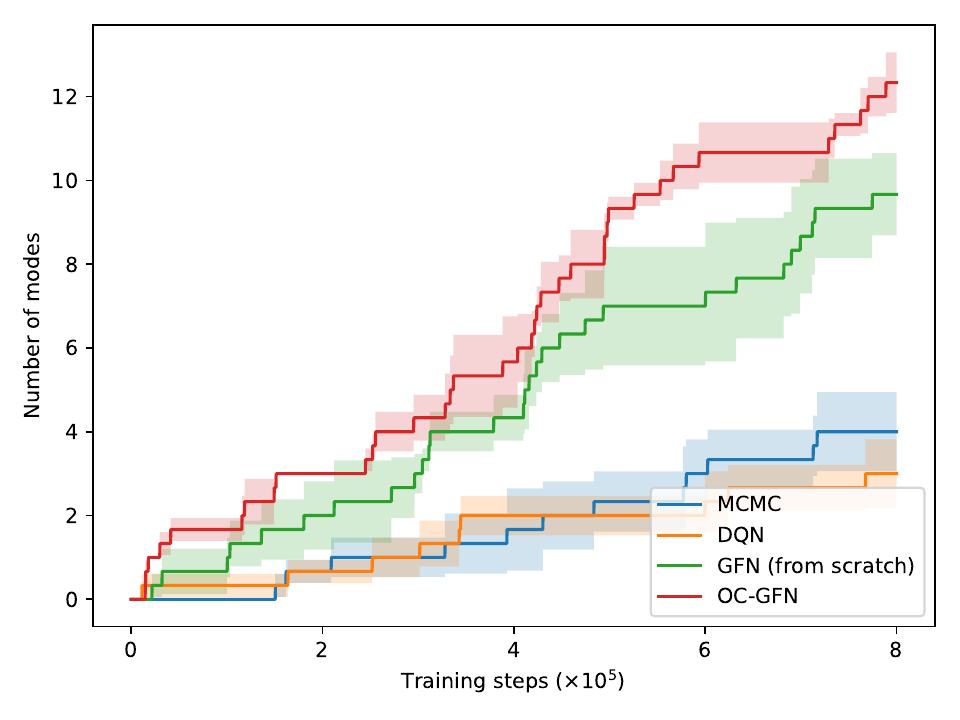}}
\subfloat[]{\includegraphics[width=0.2\linewidth]{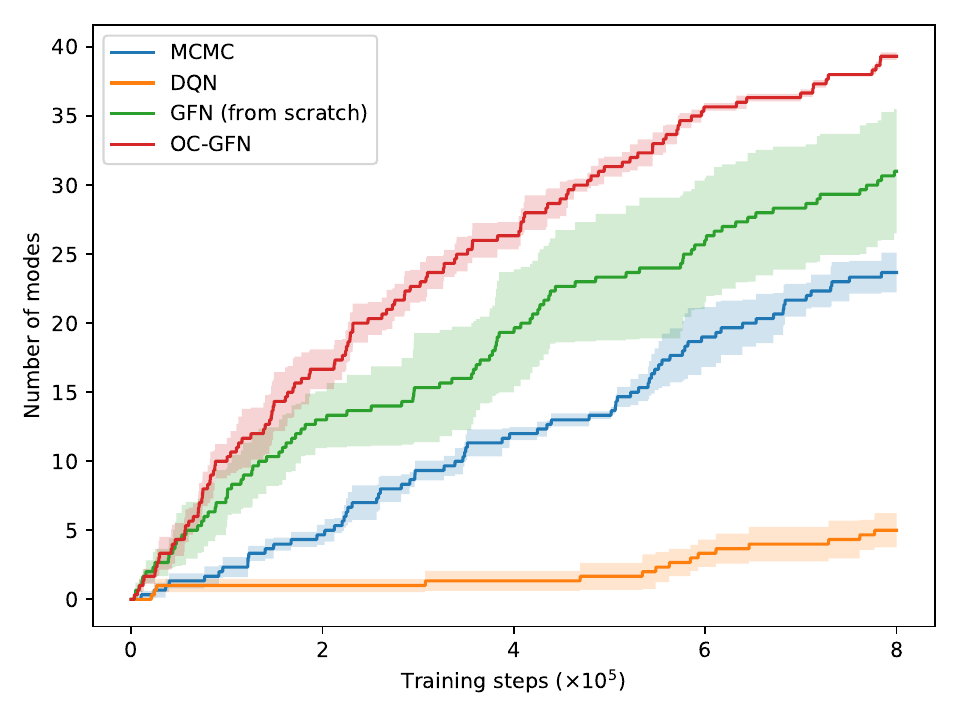}}
\subfloat[]{\includegraphics[width=0.2\linewidth]{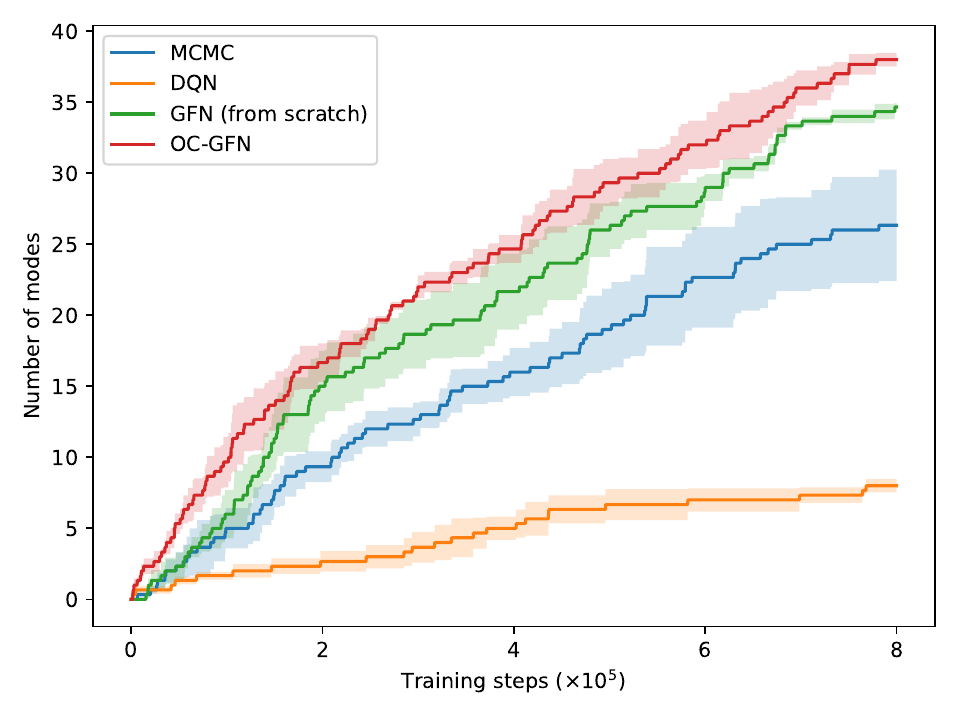}}
\vspace{-.1in}
\caption{Results in the bit sequence generation task in different with different scales of the task.}
\vspace{-.1in}
\label{fig:bit_res}
\end{figure}

As shown in Figures~\ref{fig:bit_res}(a)-(c), OC-GFN is more efficient at discovering modes and discovers more modes, compared with baselines in different scales of the task, from small to large.  DQN gets stuck and does not discover diverse modes due to the reward-maximizing nature of regular reinforcement learning algorithms, while MCMC fails to perform well in problems with larger state spaces. 
Besides different scales of the problem, we also consider different downstream tasks as in Figures~\ref{fig:bit_res}(d)-(e), which further demonstrate the effectiveness of OC-GFN in the supervised fine-tuning stage.

\subsection{Additional Results on the TF Bind Generation Task} \label{app:tfb}
\begin{figure}[!h]
\centering
\subfloat{\includegraphics[width=0.2\linewidth]{./figs/tfb_0_modes.pdf}} 
\subfloat{\includegraphics[width=0.2\linewidth]{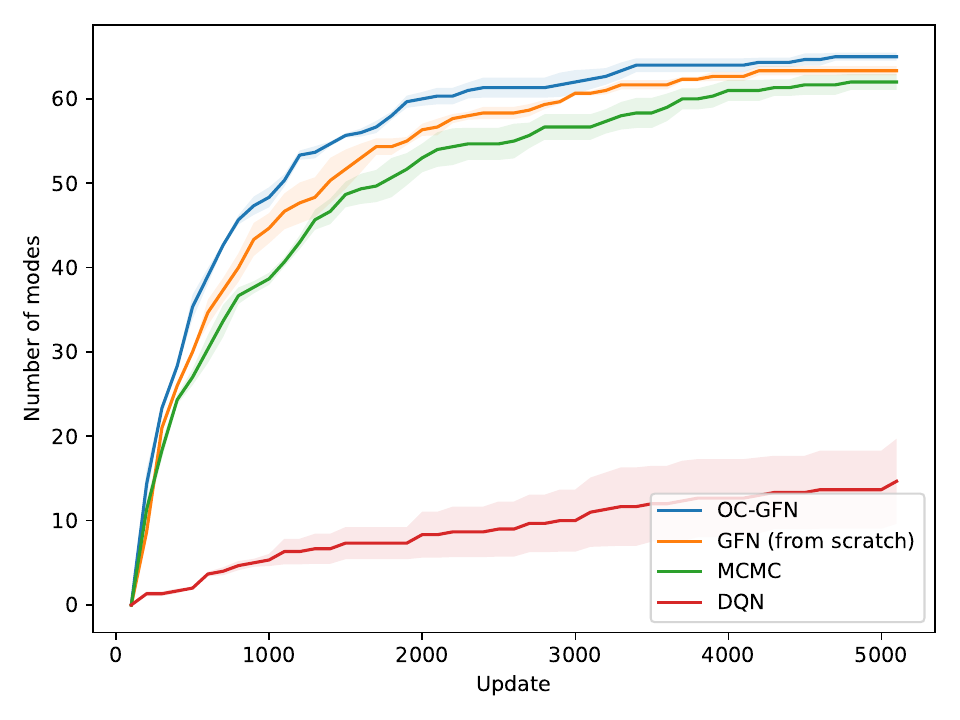}} 
\subfloat{\includegraphics[width=0.2\linewidth]{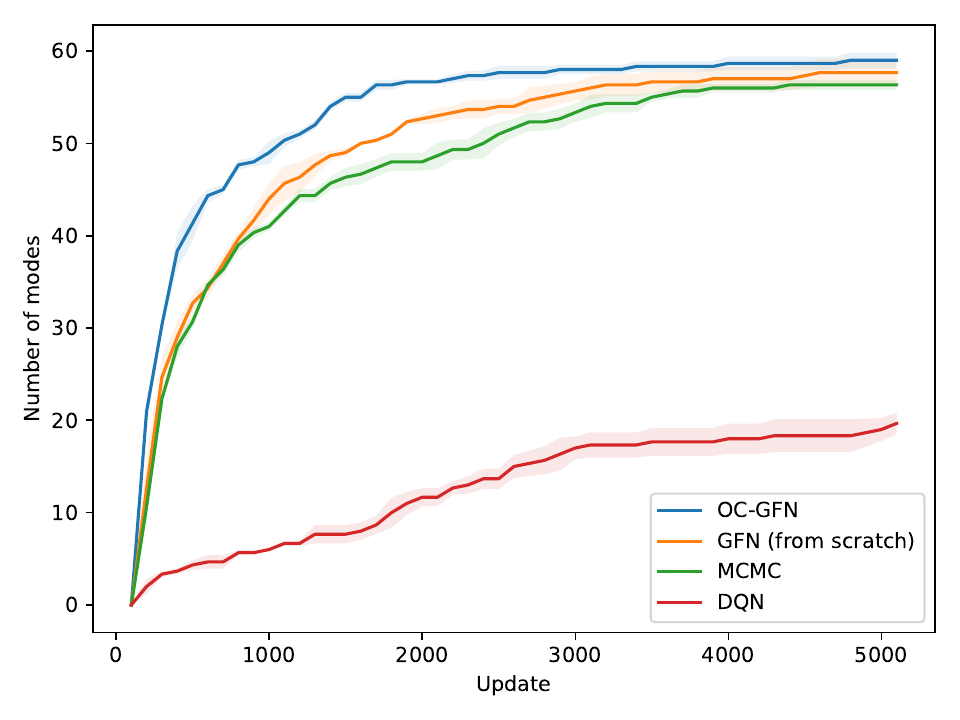}} 
\subfloat{\includegraphics[width=0.2\linewidth]{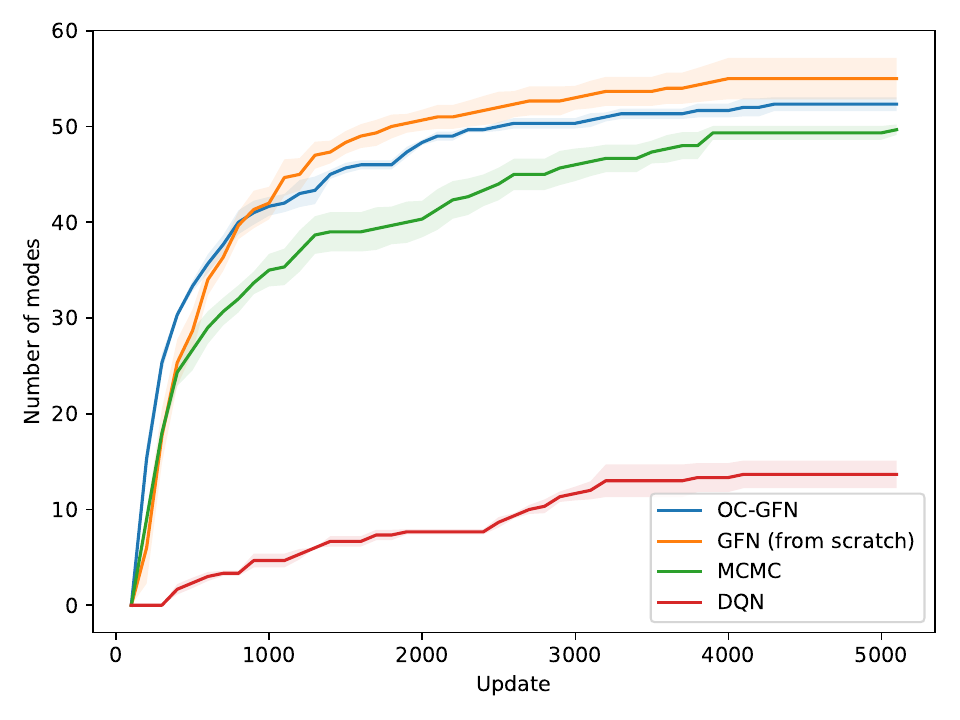}} 
\subfloat{\includegraphics[width=0.2\linewidth]{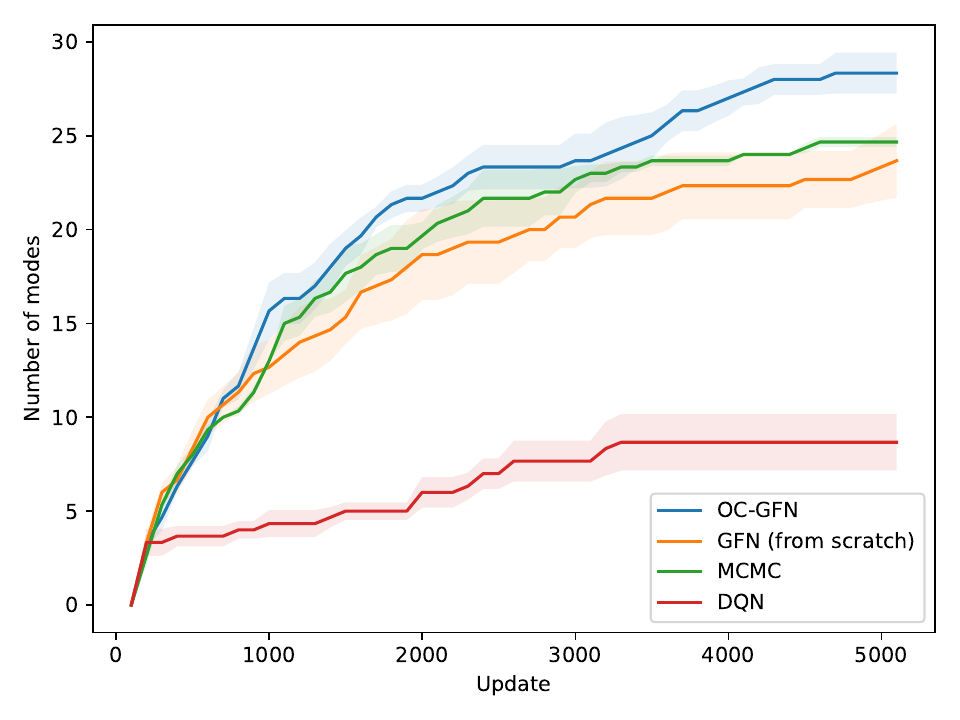}} \\
\subfloat{\includegraphics[width=0.2\linewidth]{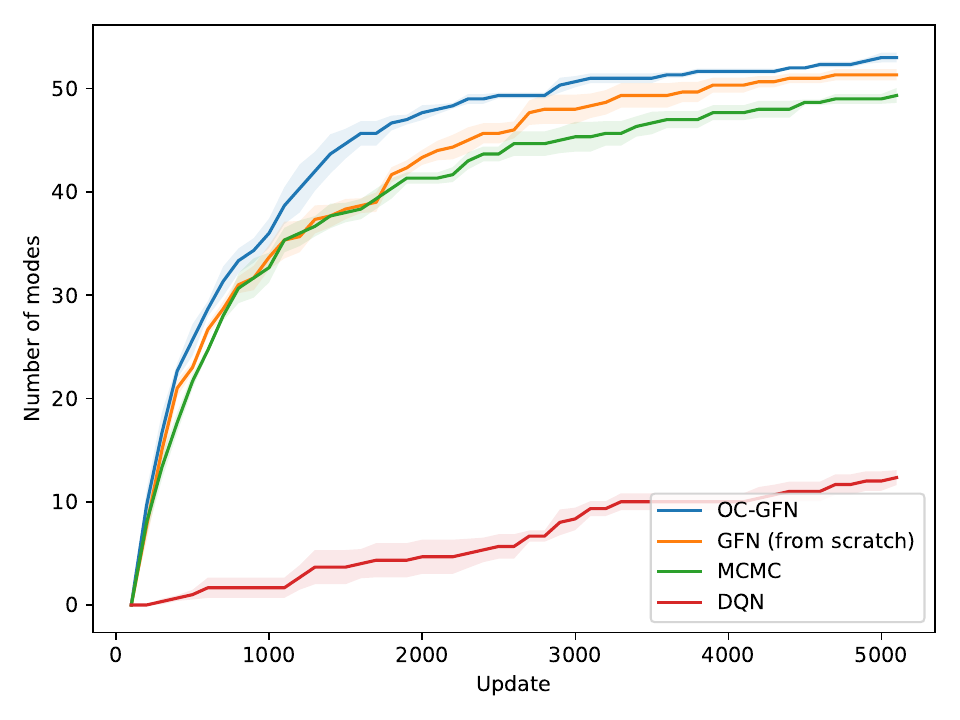}} 
\subfloat{\includegraphics[width=0.2\linewidth]{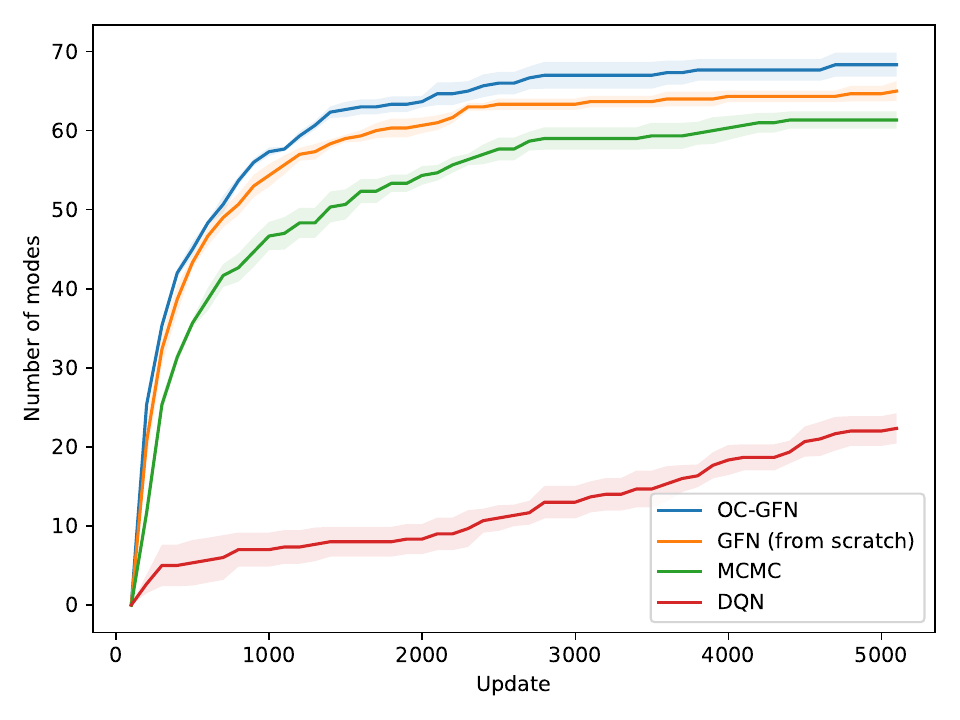}} 
\subfloat{\includegraphics[width=0.2\linewidth]{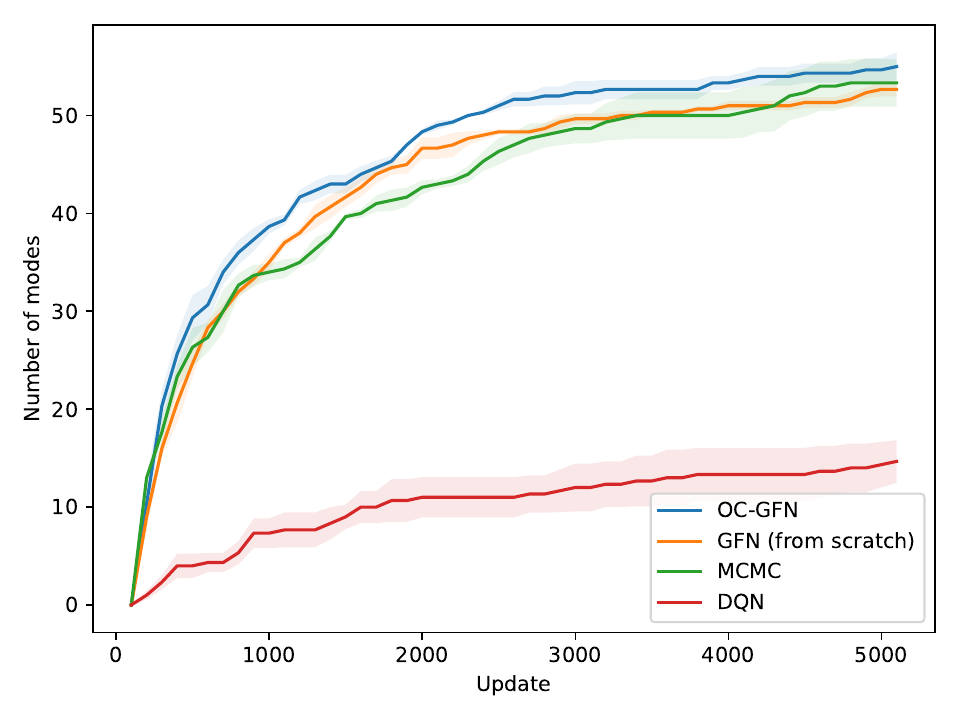}} 
\subfloat{\includegraphics[width=0.2\linewidth]{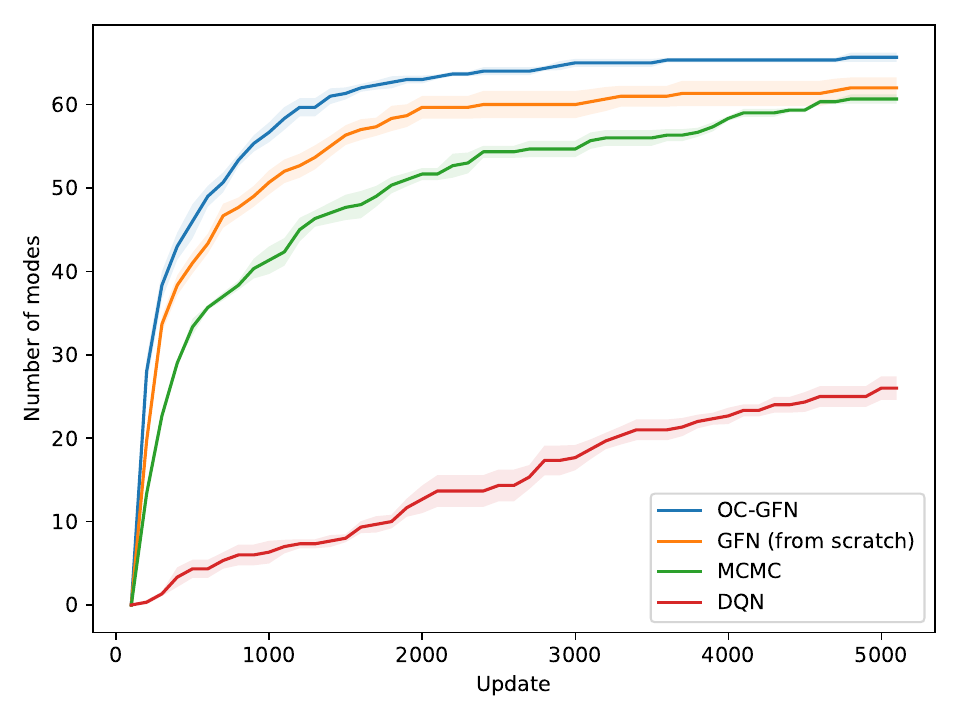}} 
\subfloat{\includegraphics[width=0.2\linewidth]{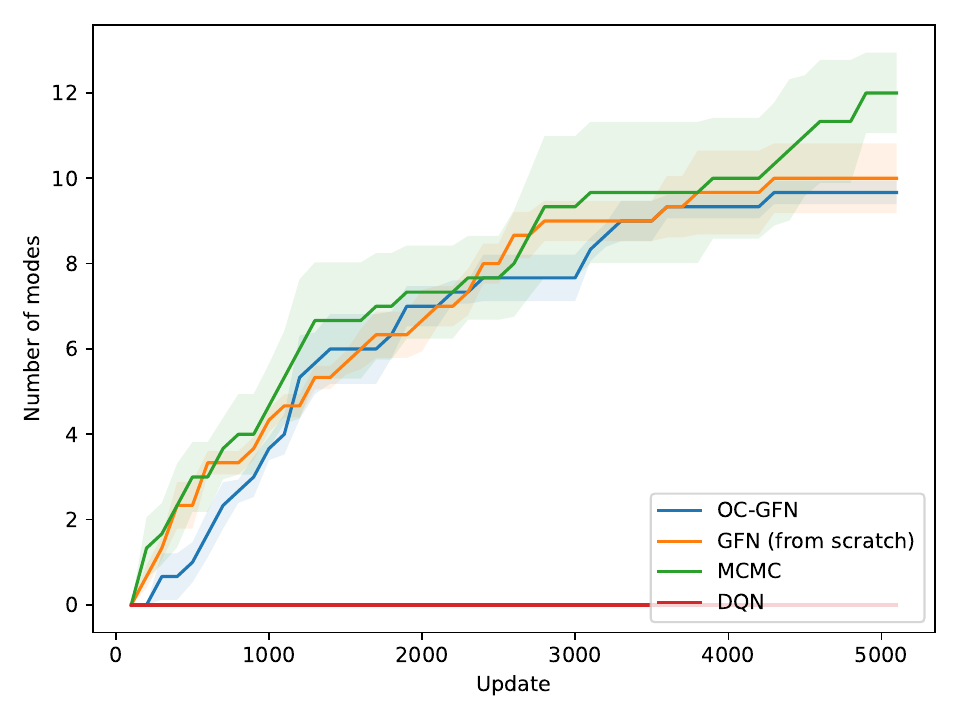}} \\
\subfloat{\includegraphics[width=0.2\linewidth]{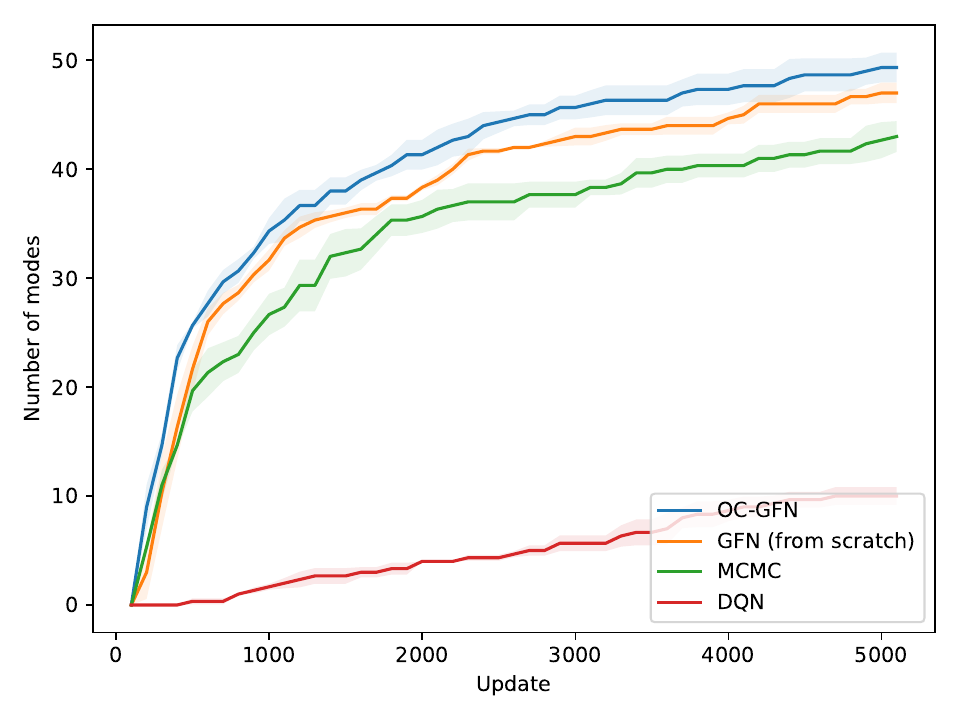}} 
\subfloat{\includegraphics[width=0.2\linewidth]{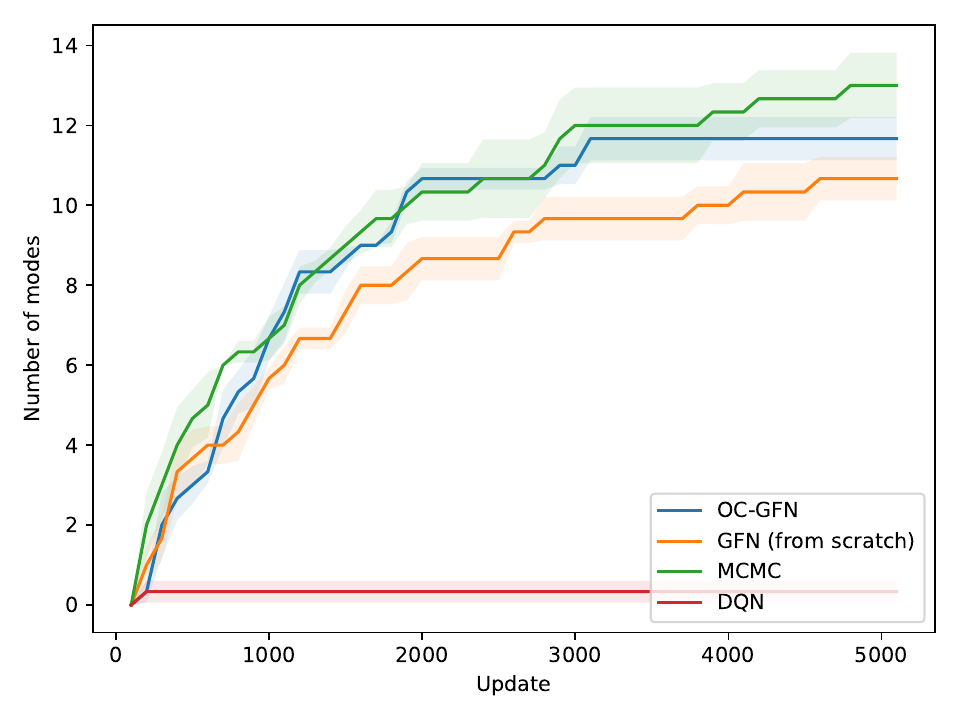}} 
\subfloat{\includegraphics[width=0.2\linewidth]{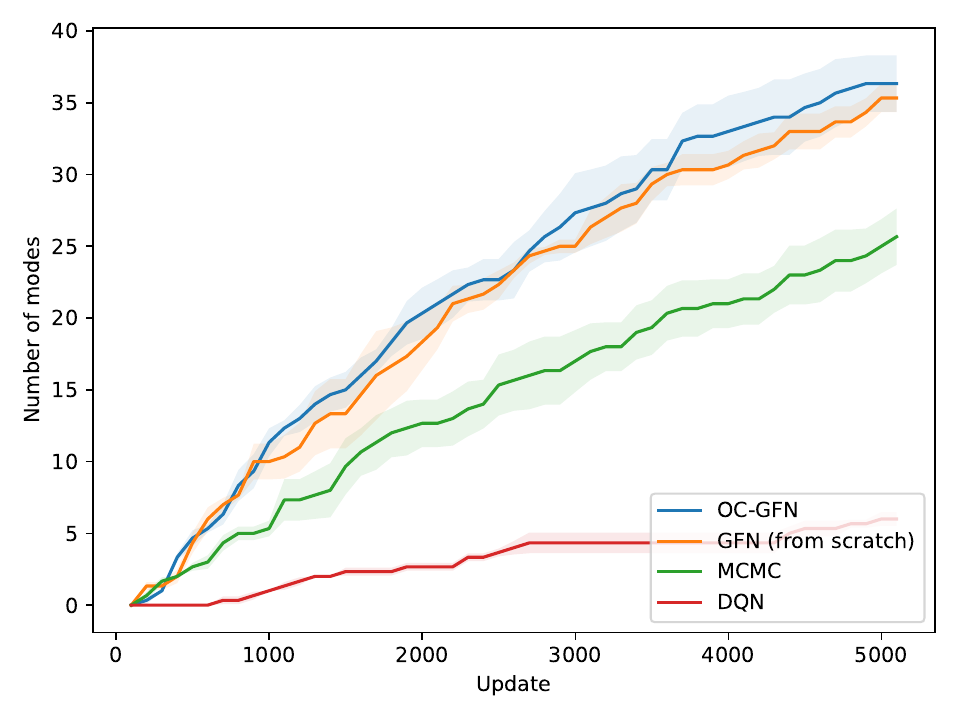}} 
\subfloat{\includegraphics[width=0.2\linewidth]{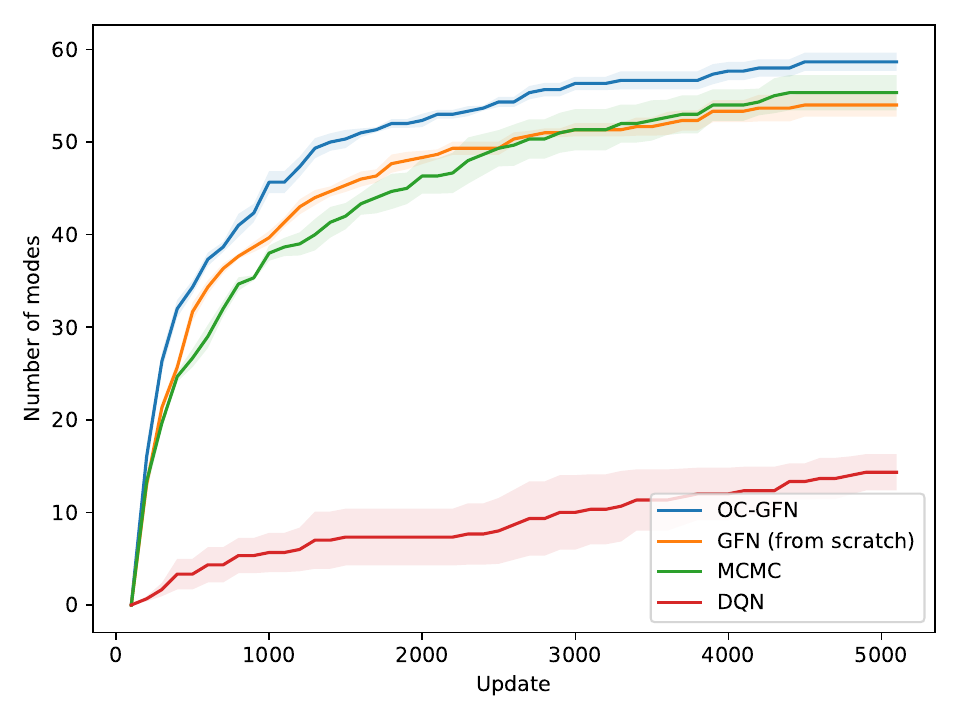}} 
\subfloat{\includegraphics[width=0.2\linewidth]{./figs/tfb_14_modes.pdf}} \\
\subfloat{\includegraphics[width=0.2\linewidth]{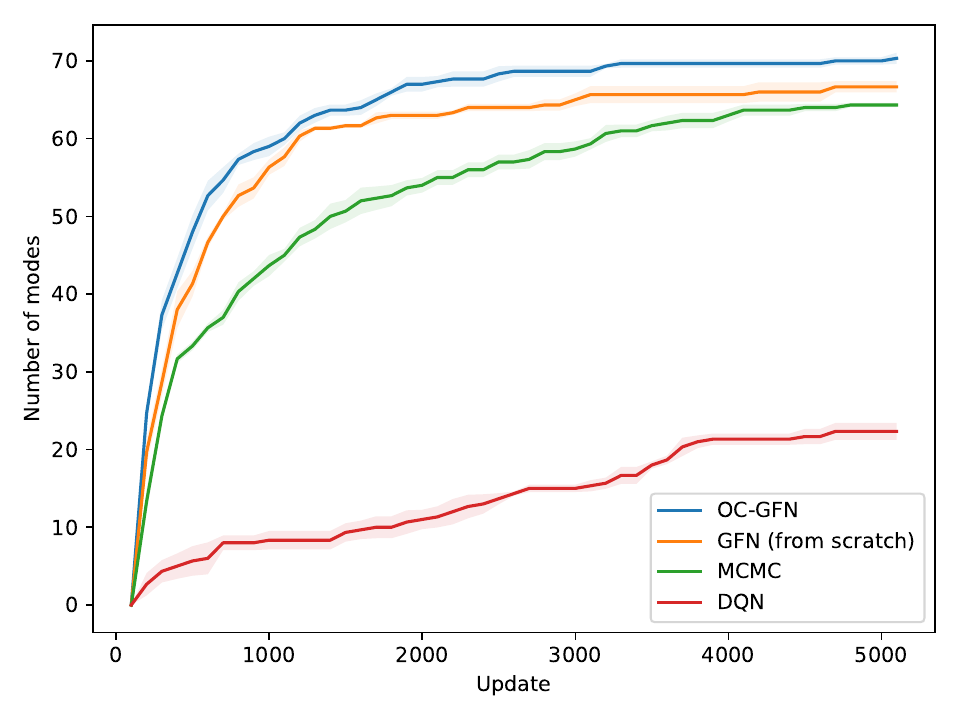}} 
\subfloat{\includegraphics[width=0.2\linewidth]{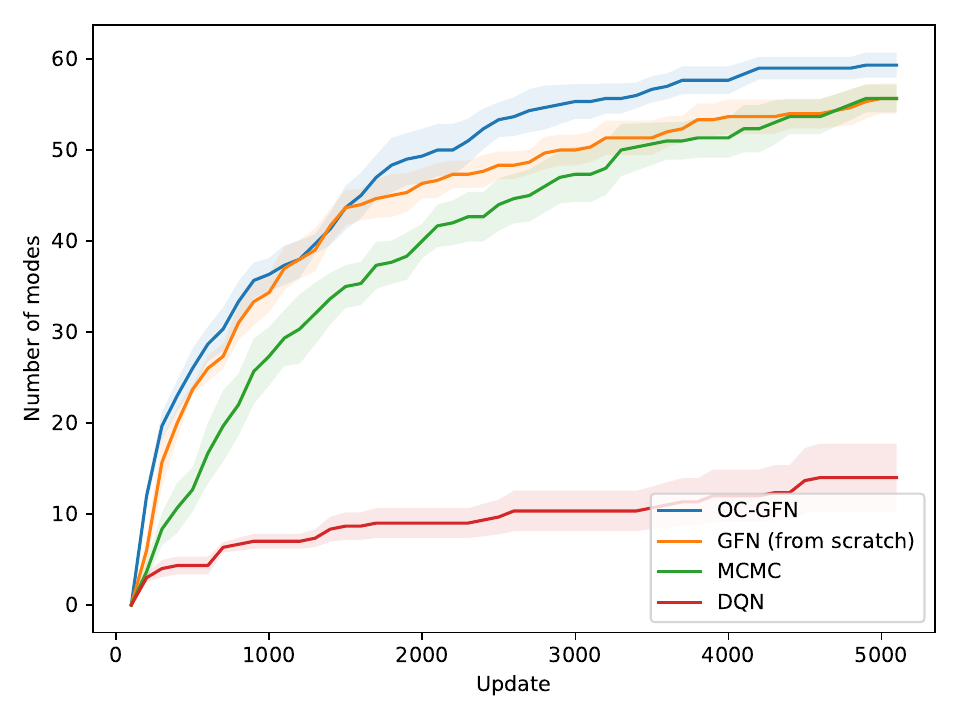}} 
\subfloat{\includegraphics[width=0.2\linewidth]{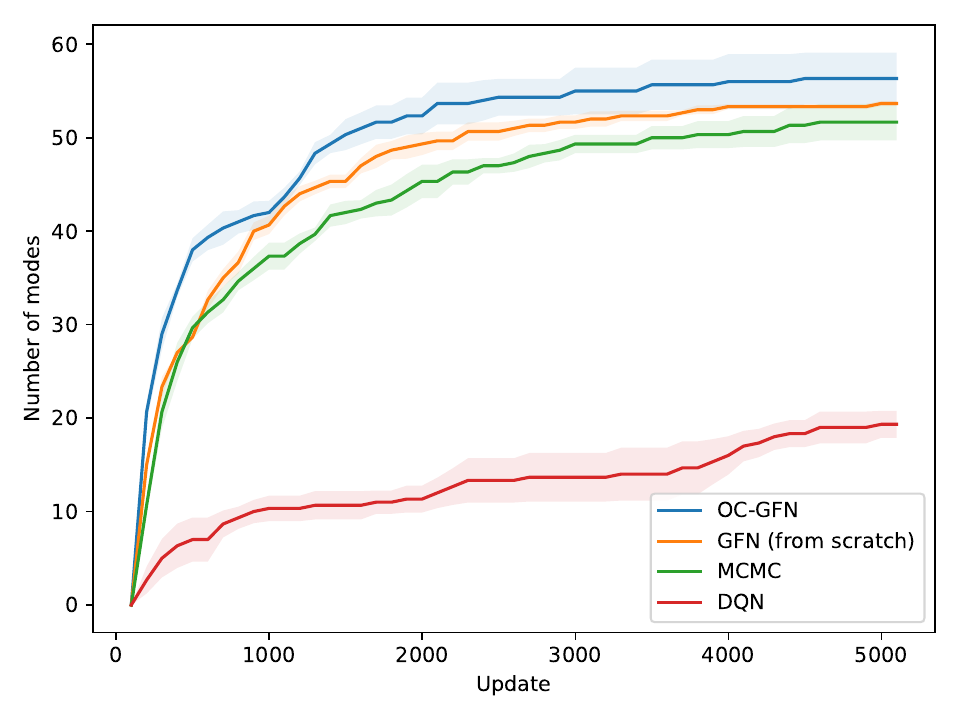}} 
\subfloat{\includegraphics[width=0.2\linewidth]{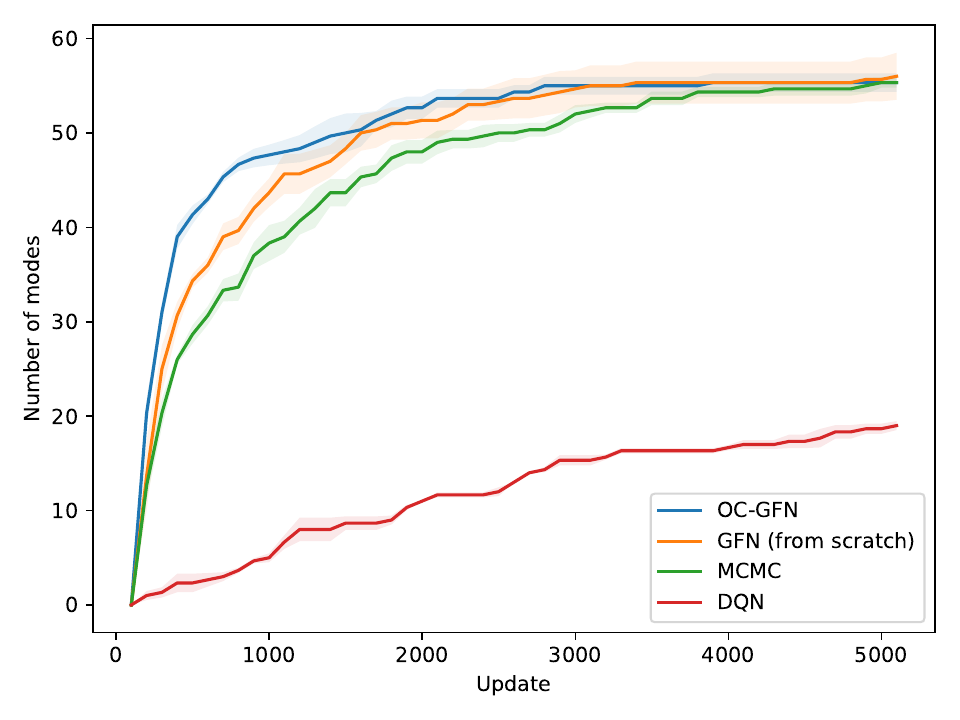}} 
\subfloat{\includegraphics[width=0.2\linewidth]{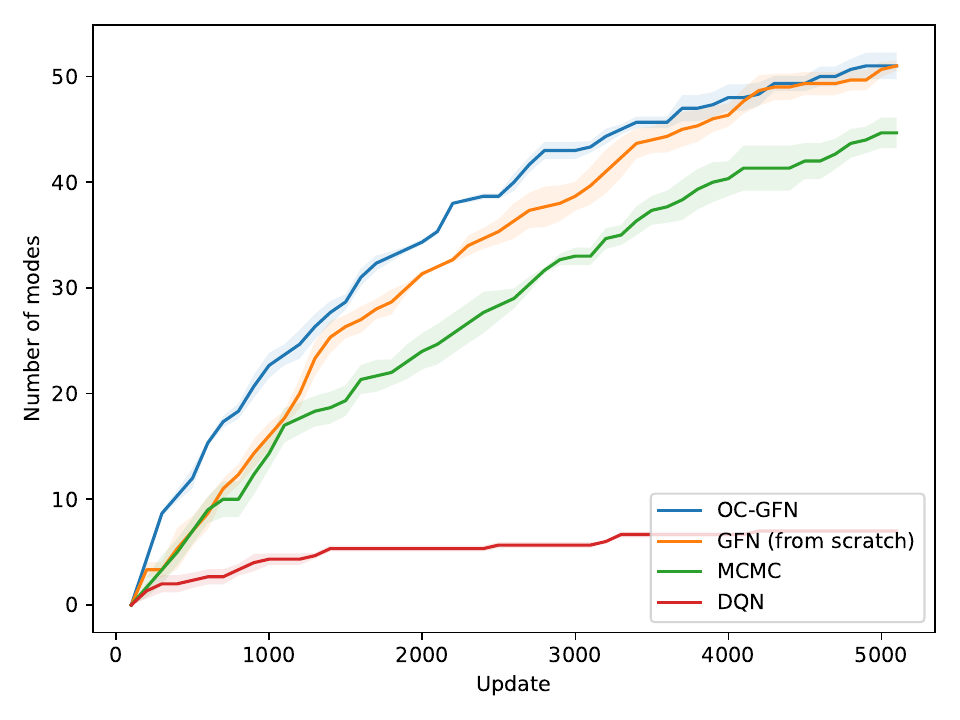}} \\
\subfloat{\includegraphics[width=0.2\linewidth]{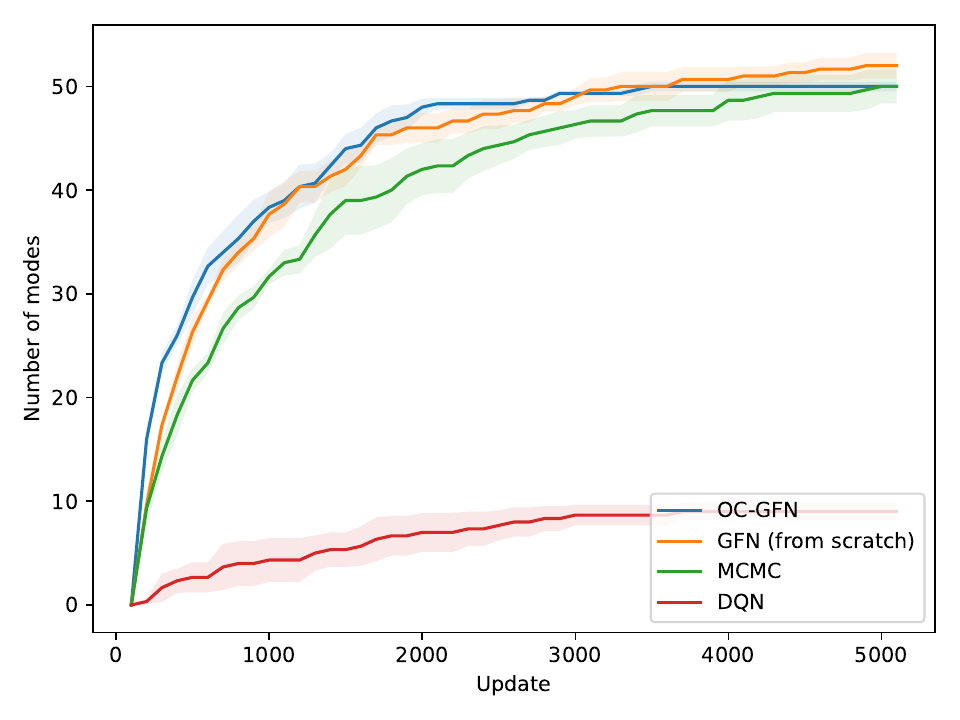}} 
\subfloat{\includegraphics[width=0.2\linewidth]{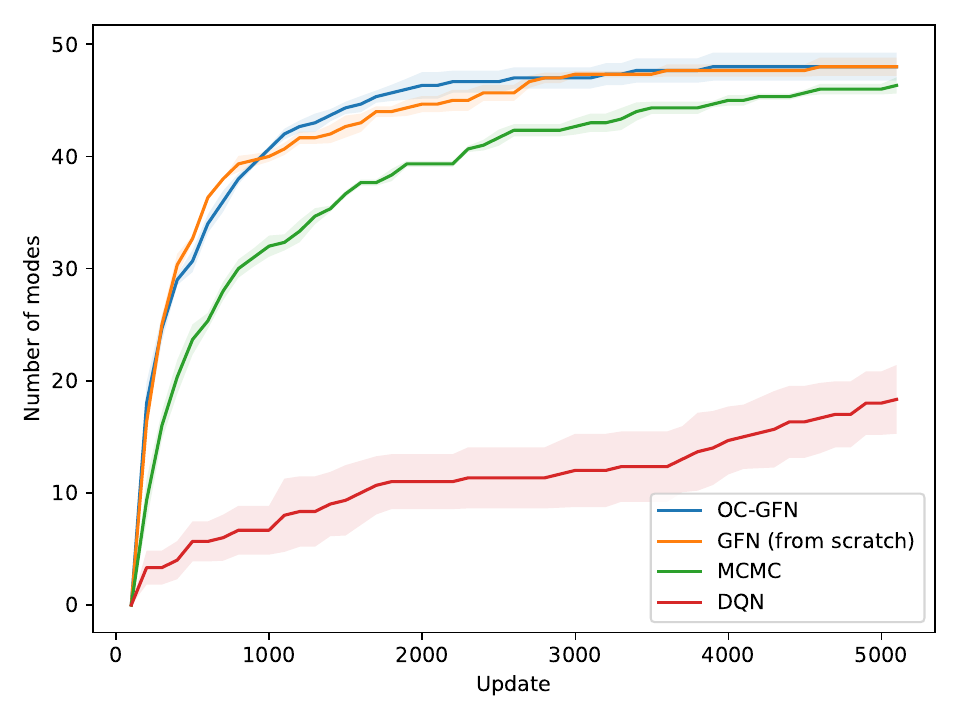}} 
\subfloat{\includegraphics[width=0.2\linewidth]{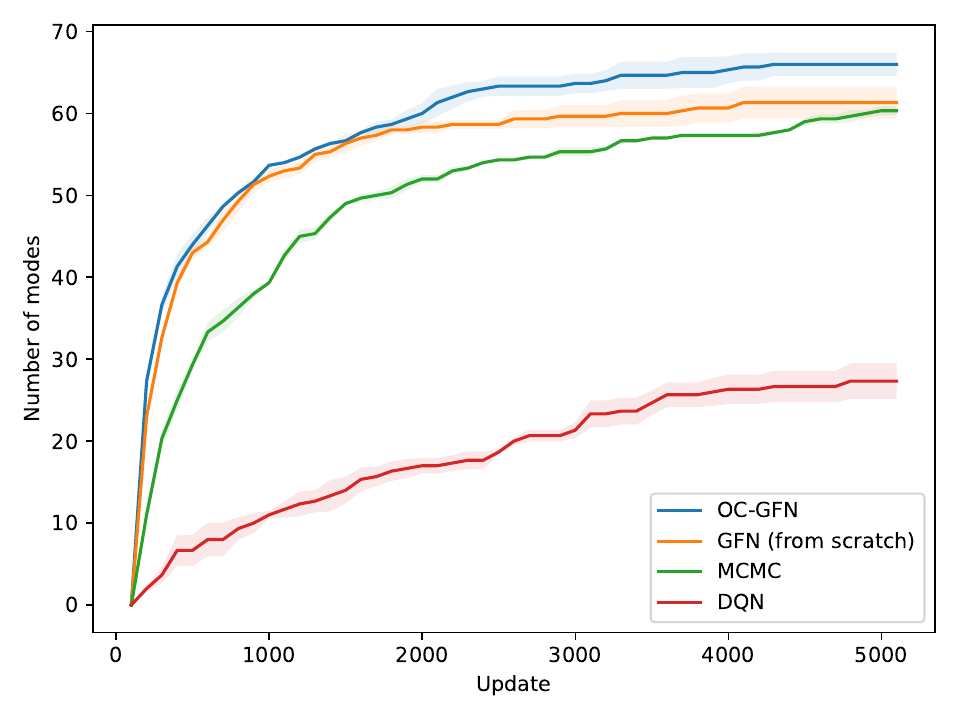}} 
\subfloat{\includegraphics[width=0.2\linewidth]{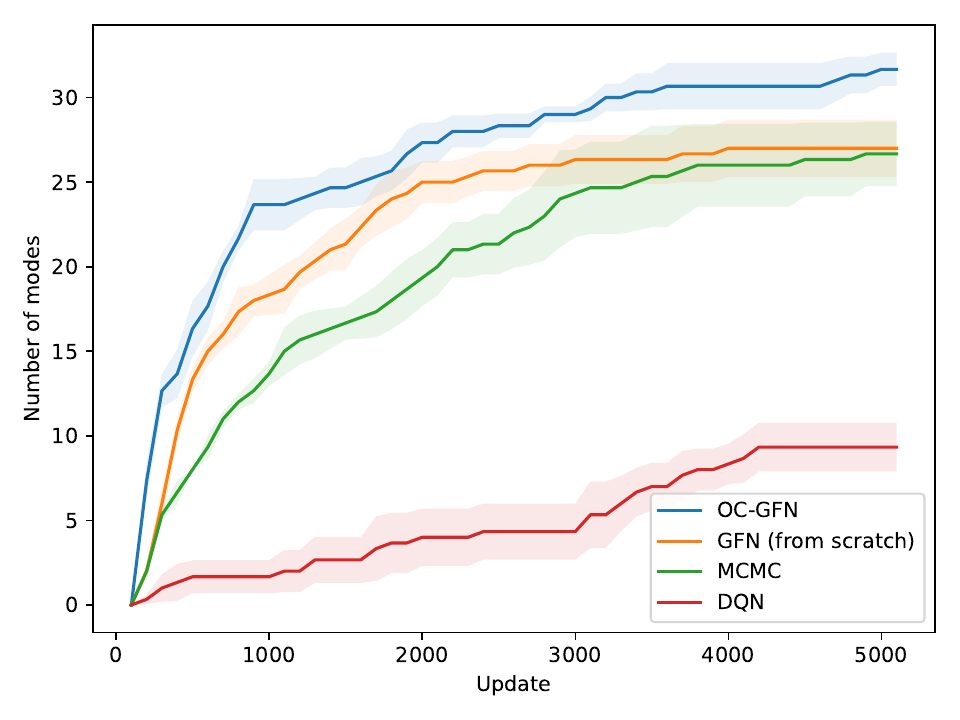}} 
\subfloat{\includegraphics[width=0.2\linewidth]{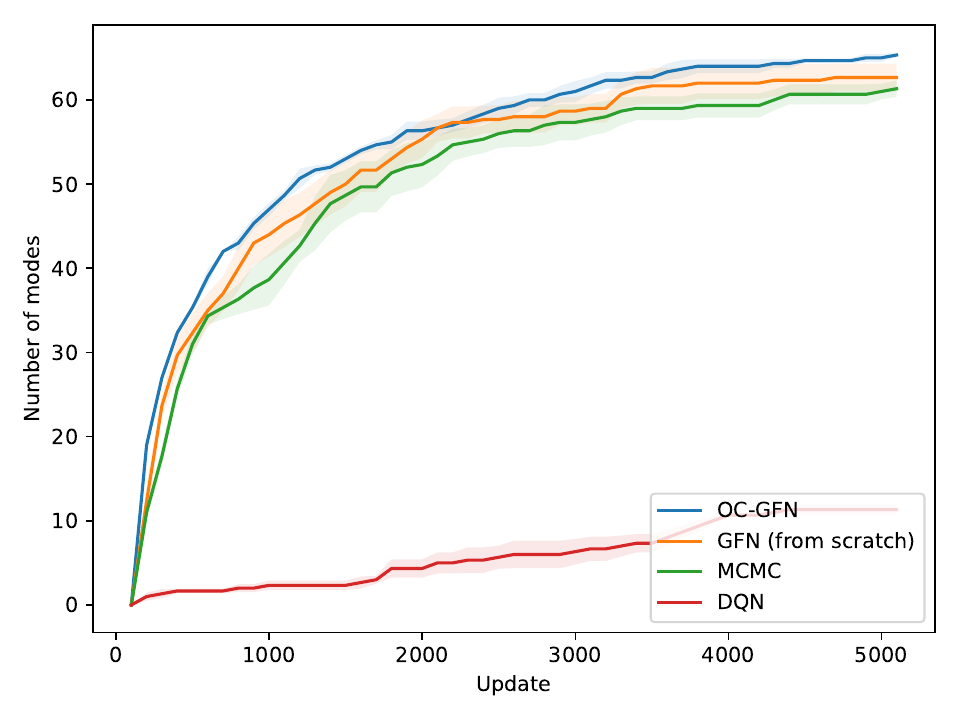}} \\
\subfloat{\includegraphics[width=0.2\linewidth]{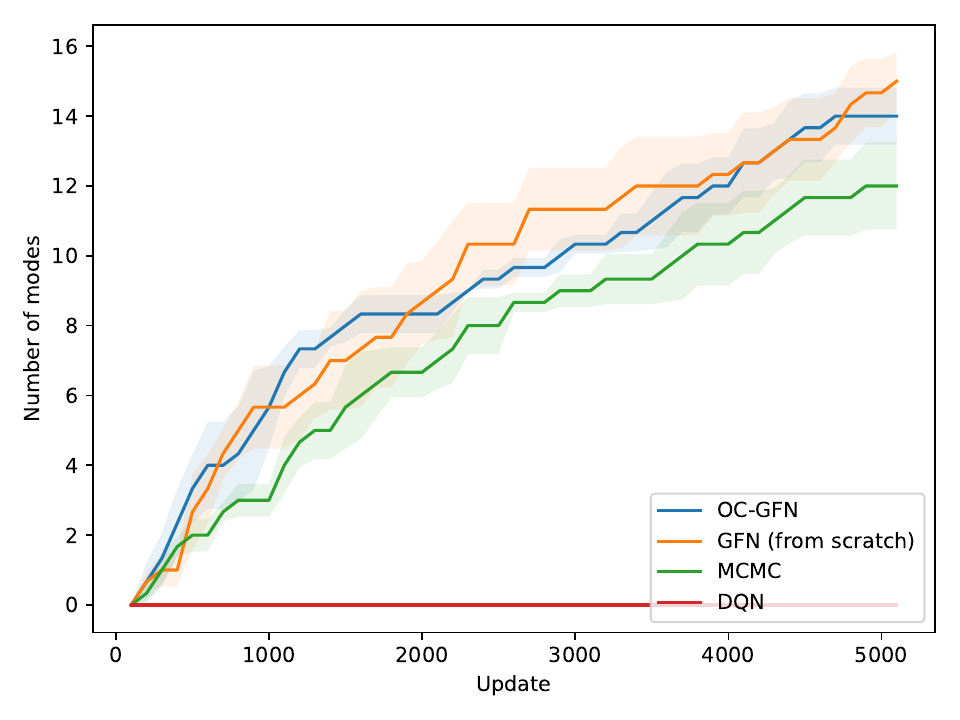}} 
\subfloat{\includegraphics[width=0.2\linewidth]{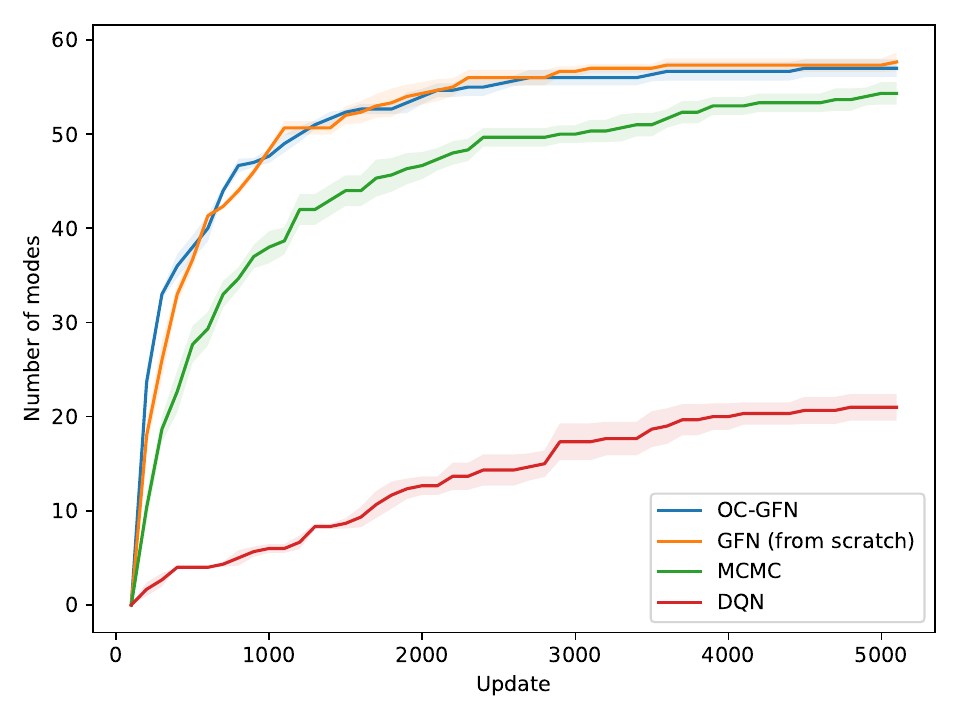}} 
\subfloat{\includegraphics[width=0.2\linewidth]{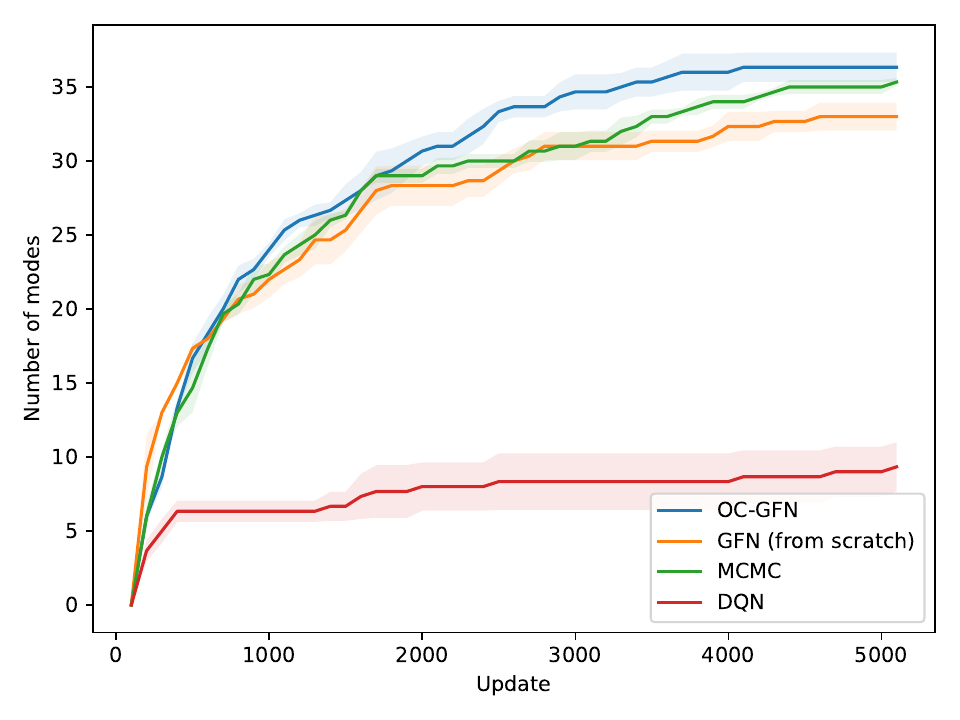}} 
\subfloat{\includegraphics[width=0.2\linewidth]{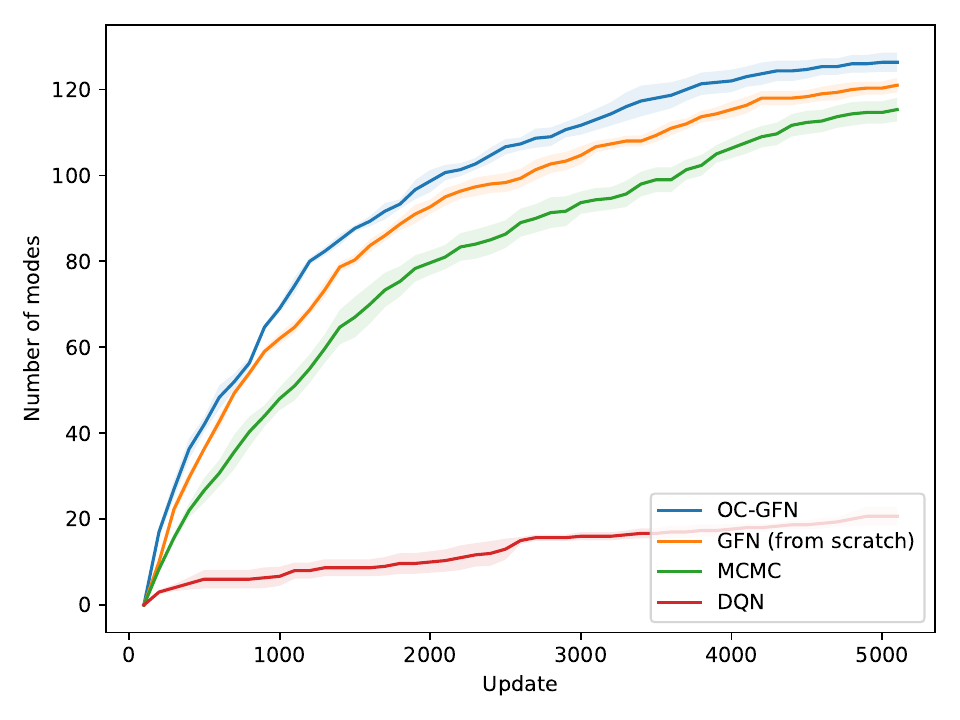}} 
\subfloat{\includegraphics[width=0.2\linewidth]{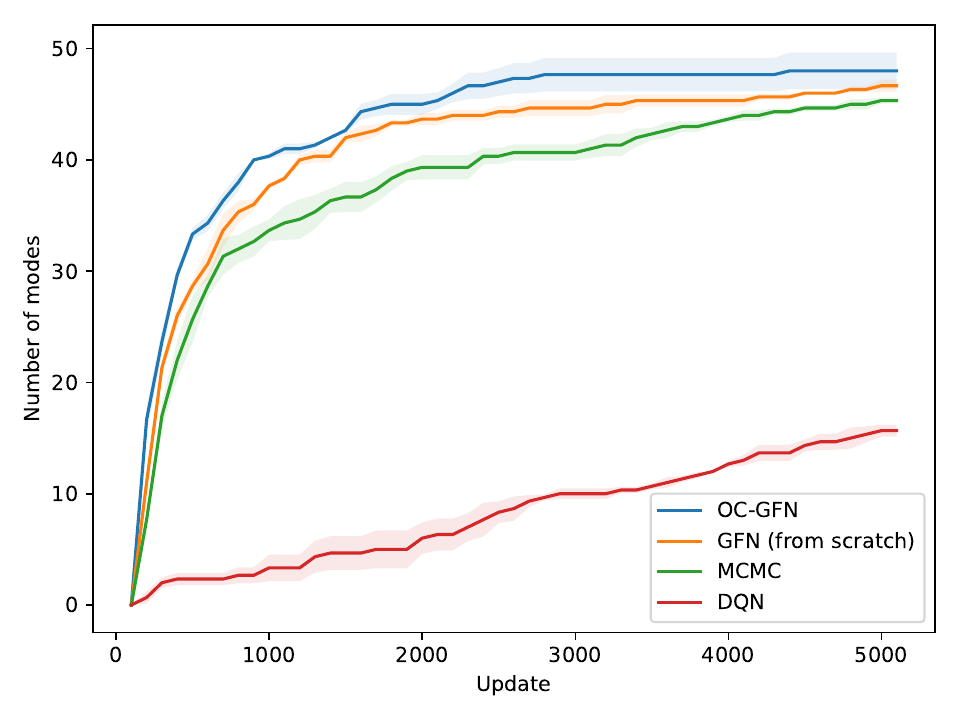}} 
\caption{Full Results in the TF Bind generation task.}
\label{fig:all_tfb}
\end{figure}

The full results in the TF Bind generation task are demonstrated in Figure~\ref{fig:all_tfb} for the $30$ downstream from~\citep{lorenz2011viennarna}. 
The mean rank averaged over all tasks is summarized in Table~\ref{tab:tfb_rank}, where OC-GFN significantly outperforms baselines and is able to discover more modes in a more efficient way.

\begin{table}[!h]
\caption{Mean rank of each baseline in all downstream tasks (the lower the better).}
\label{tab:tfb_rank}
\centering
\begin{tabular}{ccccc}
\hline
 ~ & DQN & MCMC & GFN (from scratch) & OC-GFN \\
\hline
Number of modes & $4.00$ & $2.77$ & $1.97$ & $\textbf{1.27}$ \\
Top-$K$ reward & $4.00$ & $2.80$ & $2.00$ & $\textbf{1.20}$ \\
\hline
\end{tabular}
\end{table}

\subsection{Additional Results on the RNA Generation Task} \label{app:rna}
\begin{figure}[!h]
\centering
\subfloat[]{\includegraphics[width=0.2\linewidth]{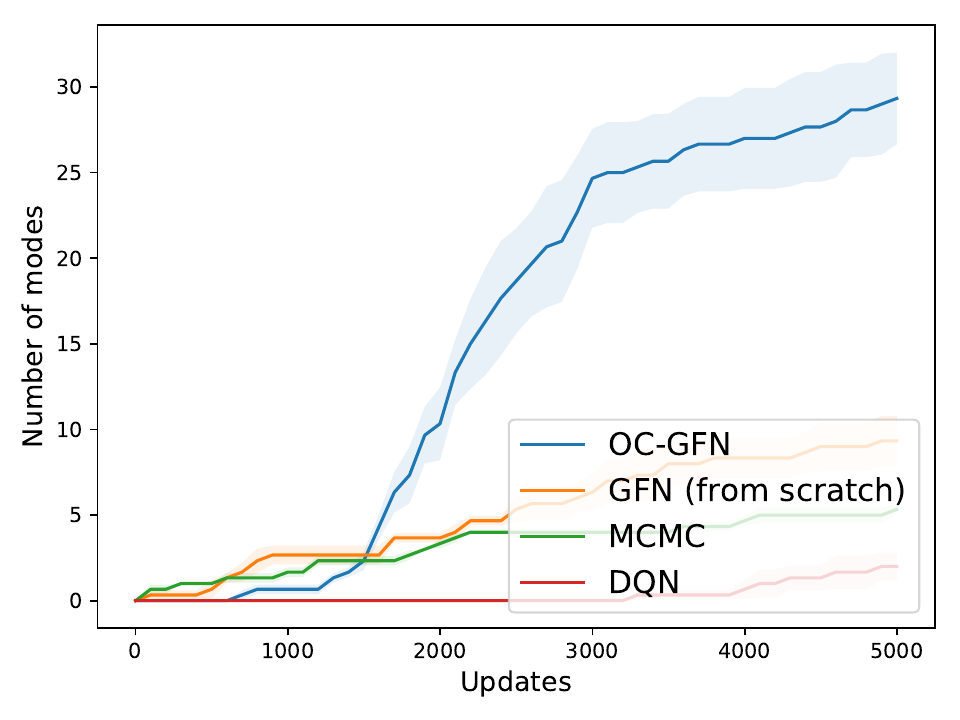}}
\subfloat[]{\includegraphics[width=0.2\linewidth]{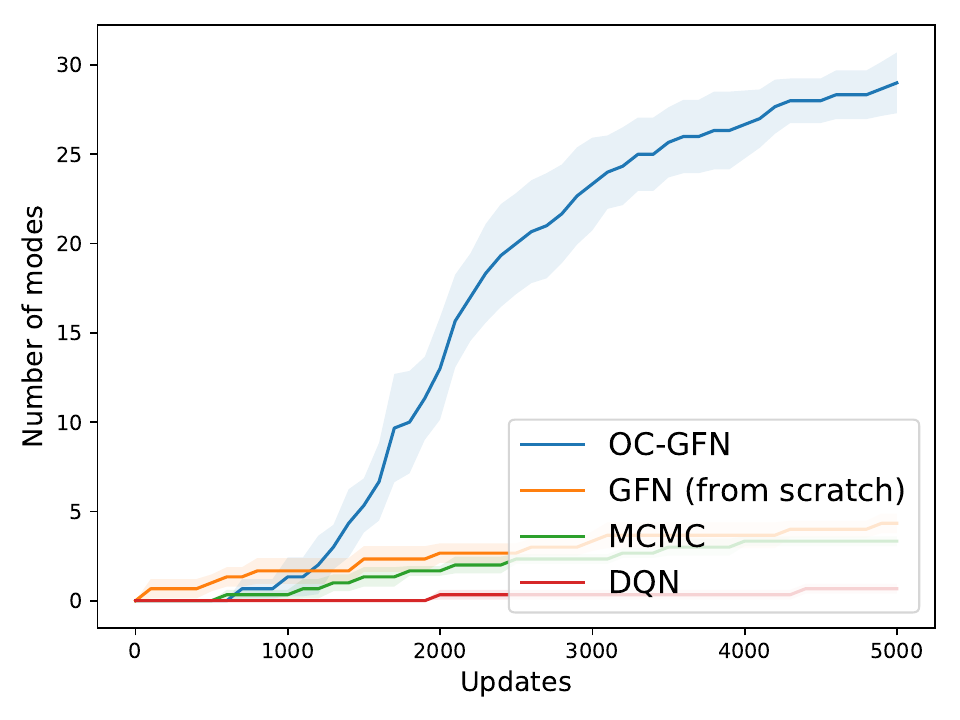}}
\subfloat[]{\includegraphics[width=0.2\linewidth]{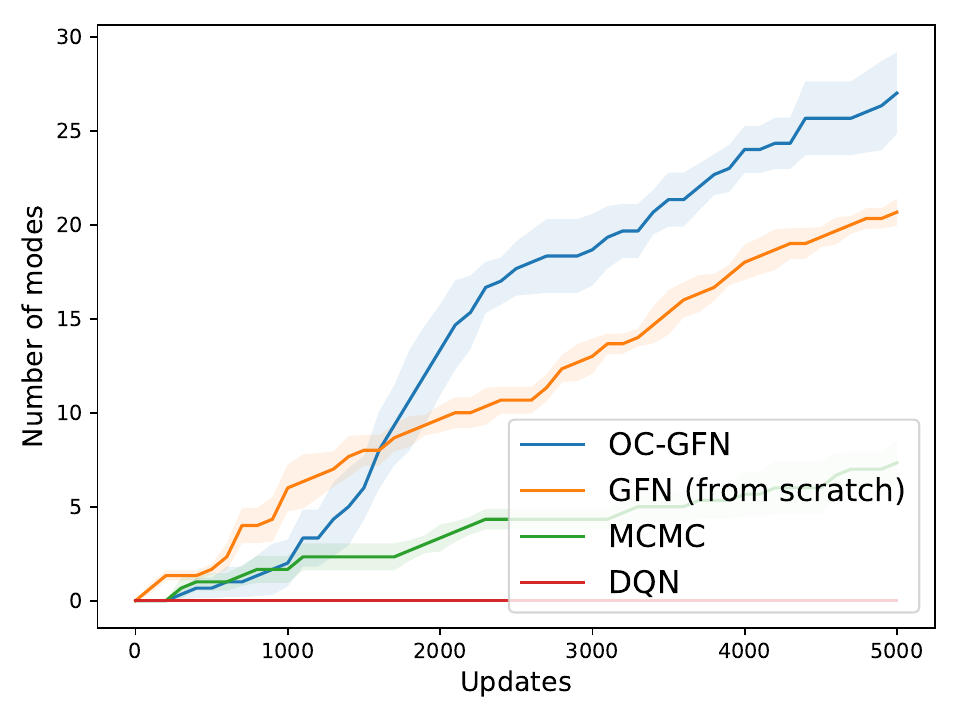}}
\subfloat[]{\includegraphics[width=0.2\linewidth]{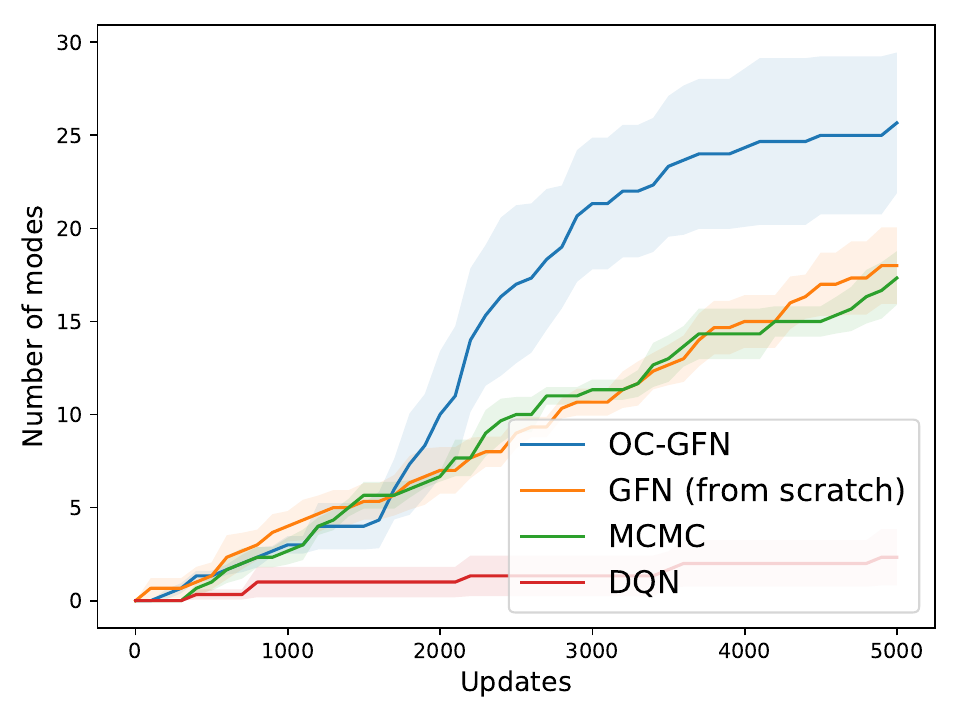}}
\vspace{-.1in}
\caption{Full Results in the RNA generation task.}
\vspace{-.1in}
\label{fig:all_rna}
\end{figure}

The full results in the RNA sequence generation task are demonstrated in Figure~\ref{fig:all_rna} for the four downstream from~\citep{lorenz2011viennarna}, where OC-GFN significantly outperforms baseline methods achieving better diversity in a more efficient manner.

\section{Limitations and future directions} \label{app:c}
We address the potential challenge of sparse rewards in training outcome-conditioned GFlowNet in the unsupervised pre-training stage with a contrastive learning procedure based on the concept of goal relabeing. It is an interesting future direction for considering smoother rewards or other techniques for addressing this problem. In addition, it is also promising to consider continuous outcomes in future works (which has the potential to tackle even larger-scale problems), while our work mainly focuses on discrete outcomes.

\end{document}

%% file: math_commands.tex
\usepackage{amsmath,amsfonts,bm}

\def\eqref#1{equation~\ref{#1}}
\def\1{\bm{1}}

\DeclareMathAlphabet{\mathsfit}{\encodingdefault}{\sfdefault}{m}{sl}
\SetMathAlphabet{\mathsfit}{bold}{\encodingdefault}{\sfdefault}{bx}{n}

%% file: main.bbl
\begin{thebibliography}{47}
\providecommand{\natexlab}[1]{#1}
\providecommand{\url}[1]{\texttt{#1}}
\expandafter\ifx\csname urlstyle\endcsname\relax
  \providecommand{\doi}[1]{doi: #1}\else
  \providecommand{\doi}{doi: \begingroup \urlstyle{rm}\Url}\fi

\bibitem[Andrychowicz et~al.(2017)Andrychowicz, Wolski, Ray, Schneider, Fong,
  Welinder, McGrew, Tobin, Pieter~Abbeel, and
  Zaremba]{andrychowicz2017hindsight}
Marcin Andrychowicz, Filip Wolski, Alex Ray, Jonas Schneider, Rachel Fong,
  Peter Welinder, Bob McGrew, Josh Tobin, OpenAI Pieter~Abbeel, and Wojciech
  Zaremba.
\newblock Hindsight experience replay.
\newblock \emph{Advances in neural information processing systems}, 30, 2017.

\bibitem[Barrera et~al.(2016)Barrera, Vedenko, Kurland, Rogers, Gisselbrecht,
  Rossin, Woodard, Mariani, Kock, Inukai, et~al.]{barrera2016survey}
Luis~A Barrera, Anastasia Vedenko, Jesse~V Kurland, Julia~M Rogers, Stephen~S
  Gisselbrecht, Elizabeth~J Rossin, Jaie Woodard, Luca Mariani, Kian~Hong Kock,
  Sachi Inukai, et~al.
\newblock Survey of variation in human transcription factors reveals prevalent
  dna binding changes.
\newblock \emph{Science}, 351\penalty0 (6280):\penalty0 1450--1454, 2016.

\bibitem[Bengio et~al.(2021)Bengio, Jain, Korablyov, Precup, and
  Bengio]{bengio2021flow}
Emmanuel Bengio, Moksh Jain, Maksym Korablyov, Doina Precup, and Yoshua Bengio.
\newblock Flow network based generative models for non-iterative diverse
  candidate generation.
\newblock \emph{Neural Information Processing Systems (NeurIPS)}, 2021.

\bibitem[Bengio et~al.(2023)Bengio, Lahlou, Deleu, Hu, Tiwari, and
  Bengio]{bengio2023foundations}
Yoshua Bengio, Salem Lahlou, Tristan Deleu, Edward~J. Hu, Mo~Tiwari, and
  Emmanuel Bengio.
\newblock Gflownet foundations.
\newblock \emph{Journal of Machine Learning Research}, 24\penalty0
  (210):\penalty0 1--55, 2023.
\newblock URL \url{http://jmlr.org/papers/v24/22-0364.html}.

\bibitem[Brown et~al.(2020)Brown, Mann, Ryder, Subbiah, Kaplan, Dhariwal,
  Neelakantan, Shyam, Sastry, Askell, et~al.]{brown2020language}
Tom Brown, Benjamin Mann, Nick Ryder, Melanie Subbiah, Jared~D Kaplan, Prafulla
  Dhariwal, Arvind Neelakantan, Pranav Shyam, Girish Sastry, Amanda Askell,
  et~al.
\newblock Language models are few-shot learners.
\newblock \emph{Advances in neural information processing systems},
  33:\penalty0 1877--1901, 2020.

\bibitem[Burda et~al.(2018)Burda, Edwards, Storkey, and
  Klimov]{burda2018exploration}
Yuri Burda, Harrison Edwards, Amos Storkey, and Oleg Klimov.
\newblock Exploration by random network distillation.
\newblock \emph{arXiv preprint arXiv:1810.12894}, 2018.

\bibitem[Chebotar et~al.(2021)Chebotar, Hausman, Lu, Xiao, Kalashnikov, Varley,
  Irpan, Eysenbach, Julian, Finn, et~al.]{chebotar2021actionable}
Yevgen Chebotar, Karol Hausman, Yao Lu, Ted Xiao, Dmitry Kalashnikov, Jake
  Varley, Alex Irpan, Benjamin Eysenbach, Ryan Julian, Chelsea Finn, et~al.
\newblock Actionable models: Unsupervised offline reinforcement learning of
  robotic skills.
\newblock \emph{arXiv preprint arXiv:2104.07749}, 2021.

\bibitem[Dai et~al.(2020)Dai, Singh, Dai, Sutton, and
  Schuurmans]{dai2020learning}
Hanjun Dai, Rishabh Singh, Bo~Dai, Charles Sutton, and Dale Schuurmans.
\newblock Learning discrete energy-based models via auxiliary-variable local
  exploration.
\newblock \emph{Advances in Neural Information Processing Systems},
  33:\penalty0 10443--10455, 2020.

\bibitem[Deleu et~al.(2022)Deleu, G\'{o}is, Emezue, Rankawat, Lacoste-Julien,
  Bauer, and Bengio]{deleu2022bayesian}
Tristan Deleu, Ant\'{o}nio G\'{o}is, Chris Emezue, Mansi Rankawat, Simon
  Lacoste-Julien, Stefan Bauer, and Yoshua Bengio.
\newblock Bayesian structure learning with generative flow networks.
\newblock \emph{Uncertainty in Artificial Intelligence (UAI)}, 2022.

\bibitem[Devlin et~al.(2018)Devlin, Chang, Lee, and Toutanova]{devlin2018bert}
Jacob Devlin, Ming-Wei Chang, Kenton Lee, and Kristina Toutanova.
\newblock Bert: Pre-training of deep bidirectional transformers for language
  understanding.
\newblock \emph{arXiv preprint arXiv:1810.04805}, 2018.

\bibitem[Eysenbach et~al.(2018)Eysenbach, Gupta, Ibarz, and
  Levine]{eysenbach2018diversity}
Benjamin Eysenbach, Abhishek Gupta, Julian Ibarz, and Sergey Levine.
\newblock Diversity is all you need: Learning skills without a reward function.
\newblock \emph{International Conference on Learning Representations (ICLR)},
  2018.

\bibitem[Eysenbach et~al.(2020)Eysenbach, Salakhutdinov, and
  Levine]{eysenbach2020c}
Benjamin Eysenbach, Ruslan Salakhutdinov, and Sergey Levine.
\newblock C-learning: Learning to achieve goals via recursive classification.
\newblock \emph{arXiv preprint arXiv:2011.08909}, 2020.

\bibitem[Fang et~al.(2022)Fang, Yin, Nair, and Levine]{fang2022planning}
Kuan Fang, Patrick Yin, Ashvin Nair, and Sergey Levine.
\newblock Planning to practice: Efficient online fine-tuning by composing goals
  in latent space.
\newblock In \emph{2022 IEEE/RSJ International Conference on Intelligent Robots
  and Systems (IROS)}, pages 4076--4083. IEEE, 2022.

\bibitem[Hansen et~al.(2019)Hansen, Dabney, Barreto, Van~de Wiele,
  Warde-Farley, and Mnih]{hansen2019fast}
Steven Hansen, Will Dabney, Andre Barreto, Tom Van~de Wiele, David
  Warde-Farley, and Volodymyr Mnih.
\newblock Fast task inference with variational intrinsic successor features.
\newblock \emph{arXiv preprint arXiv:1906.05030}, 2019.

\bibitem[Henaff(2020)]{henaff2020data}
Olivier Henaff.
\newblock Data-efficient image recognition with contrastive predictive coding.
\newblock In \emph{International conference on machine learning}, pages
  4182--4192. PMLR, 2020.

\bibitem[Howard and Ruder(2018)]{howard2018universal}
Jeremy Howard and Sebastian Ruder.
\newblock Universal language model fine-tuning for text classification.
\newblock In \emph{Proceedings of the 56th Annual Meeting of the Association
  for Computational Linguistics (Volume 1: Long Papers)}, pages 328--339, 2018.

\bibitem[Hu et~al.(2023)Hu, Malkin, Jain, Everett, Graikos, and
  Bengio]{hu2023gflownet}
Edward~J Hu, Nikolay Malkin, Moksh Jain, Katie~E Everett, Alexandros Graikos,
  and Yoshua Bengio.
\newblock Gflownet-em for learning compositional latent variable models.
\newblock In \emph{International Conference on Machine Learning}, pages
  13528--13549, 2023.

\bibitem[Jaderberg et~al.(2016)Jaderberg, Mnih, Czarnecki, Schaul, Leibo,
  Silver, and Kavukcuoglu]{jaderberg2016reinforcement}
Max Jaderberg, Volodymyr Mnih, Wojciech~Marian Czarnecki, Tom Schaul, Joel~Z
  Leibo, David Silver, and Koray Kavukcuoglu.
\newblock Reinforcement learning with unsupervised auxiliary tasks.
\newblock In \emph{International Conference on Learning Representations}, 2016.

\bibitem[Jain et~al.(2022)Jain, Bengio, Hernandez-Garcia, Rector-Brooks,
  Dossou, Ekbote, Fu, Zhang, Kilgour, Zhang, Simine, Das, and
  Bengio]{jain2022biological}
Moksh Jain, Emmanuel Bengio, Alex Hernandez-Garcia, Jarrid Rector-Brooks,
  Bonaventure~F.P. Dossou, Chanakya Ekbote, Jie Fu, Tianyu Zhang, Micheal
  Kilgour, Dinghuai Zhang, Lena Simine, Payel Das, and Yoshua Bengio.
\newblock Biological sequence design with {GFlowNets}.
\newblock \emph{International Conference on Machine Learning (ICML)}, 2022.

\bibitem[Jain et~al.(2023{\natexlab{a}})Jain, Deleu, Hartford, Liu,
  Hernandez-Garcia, and Bengio]{jain2023gflownets}
Moksh Jain, Tristan Deleu, Jason Hartford, Cheng-Hao Liu, Alex
  Hernandez-Garcia, and Yoshua Bengio.
\newblock Gflownets for ai-driven scientific discovery.
\newblock \emph{Digital Discovery}, 2023{\natexlab{a}}.

\bibitem[Jain et~al.(2023{\natexlab{b}})Jain, Raparthy, Hernandez-Garcia,
  Rector-Brooks, Bengio, Miret, and Bengio]{jain2023multi}
Moksh Jain, Sharath~Chandra Raparthy, Alex Hernandez-Garcia, Jarrid
  Rector-Brooks, Yoshua Bengio, Santiago Miret, and Emmanuel Bengio.
\newblock Multi-objective gflownets.
\newblock In \emph{International Conference on Machine Learning}, pages
  14631--14653, 2023{\natexlab{b}}.

\bibitem[Kaelbling(1993)]{kaelbling1993learning}
Leslie~Pack Kaelbling.
\newblock Learning to achieve goals.
\newblock In \emph{IJCAI}, volume~2, pages 1094--8. Citeseer, 1993.

\bibitem[Kingma and Ba(2015)]{kingma2014adam}
Diederik~P Kingma and Jimmy Ba.
\newblock Adam: A method for stochastic optimization.
\newblock \emph{International Conference on Learning Representations (ICLR)},
  2015.

\bibitem[Kumar et~al.(2020)Kumar, Kumar, Levine, and Finn]{kumar2020one}
Saurabh Kumar, Aviral Kumar, Sergey Levine, and Chelsea Finn.
\newblock One solution is not all you need: Few-shot extrapolation via
  structured maxent rl.
\newblock \emph{Advances in Neural Information Processing Systems},
  33:\penalty0 8198--8210, 2020.

\bibitem[Lahlou et~al.(2023)Lahlou, Deleu, Lemos, Zhang, Volokhova,
  Hern{\'a}ndez-Garc{\i}a, Ezzine, Bengio, and Malkin]{lahlou2023theory}
Salem Lahlou, Tristan Deleu, Pablo Lemos, Dinghuai Zhang, Alexandra Volokhova,
  Alex Hern{\'a}ndez-Garc{\i}a, L{\'e}na~N{\'e}hale Ezzine, Yoshua Bengio, and
  Nikolay Malkin.
\newblock A theory of continuous generative flow networks.
\newblock In \emph{International Conference on Machine Learning}, pages
  18269--18300. PMLR, 2023.

\bibitem[Liu and Abbeel(2021)]{liu2021aps}
Hao Liu and Pieter Abbeel.
\newblock Aps: Active pretraining with successor features.
\newblock In \emph{International Conference on Machine Learning}, pages
  6736--6747. PMLR, 2021.

\bibitem[Lorenz et~al.(2011)Lorenz, Bernhart, Zu~Siederdissen, Tafer, Flamm,
  Stadler, and Hofacker]{lorenz2011viennarna}
Ronny Lorenz, Stephan~H Bernhart, Christian~H{\"o}ner Zu~Siederdissen, Hakim
  Tafer, Christoph Flamm, Peter~F Stadler, and Ivo~L Hofacker.
\newblock {ViennaRNA} package 2.0.
\newblock \emph{Algorithms for molecular biology}, 6\penalty0 (1):\penalty0 26,
  2011.

\bibitem[Madan et~al.(2022)Madan, Rector-Brooks, Korablyov, Bengio, Jain, Nica,
  Bosc, Bengio, and Malkin]{madan2022learning}
Kanika Madan, Jarrid Rector-Brooks, Maksym Korablyov, Emmanuel Bengio, Moksh
  Jain, Andrei Nica, Tom Bosc, Yoshua Bengio, and Nikolay Malkin.
\newblock Learning {GFlowNets} from partial episodes for improved convergence
  and stability.
\newblock \emph{arXiv preprint 2209.12782}, 2022.

\bibitem[Malkin et~al.(2022)Malkin, Jain, Bengio, Sun, and
  Bengio]{malkin2022trajectory}
Nikolay Malkin, Moksh Jain, Emmanuel Bengio, Chen Sun, and Yoshua Bengio.
\newblock Trajectory balance: Improved credit assignment in {GFlowNets}.
\newblock \emph{Neural Information Processing Systems (NeurIPS)}, 2022.

\bibitem[Malkin et~al.(2023)Malkin, Lahlou, Deleu, Ji, Hu, Everett, Zhang, and
  Bengio]{malkin2022gfnhvi}
Nikolay Malkin, Salem Lahlou, Tristan Deleu, Xu~Ji, Edward Hu, Katie Everett,
  Dinghuai Zhang, and Yoshua Bengio.
\newblock Gflownets and variational inference.
\newblock \emph{International Conference on Learning Representations (ICLR)},
  2023.

\bibitem[Mnih et~al.(2015)Mnih, Kavukcuoglu, Silver, Rusu, Veness, Bellemare,
  Graves, Riedmiller, Fidjeland, Ostrovski, et~al.]{mnih2015human}
Volodymyr Mnih, Koray Kavukcuoglu, David Silver, Andrei~A Rusu, Joel Veness,
  Marc~G Bellemare, Alex Graves, Martin Riedmiller, Andreas~K Fidjeland, Georg
  Ostrovski, et~al.
\newblock Human-level control through deep reinforcement learning.
\newblock \emph{nature}, 518\penalty0 (7540):\penalty0 529--533, 2015.

\bibitem[Nair et~al.(2018)Nair, Pong, Dalal, Bahl, Lin, and
  Levine]{nair2018visual}
Ashvin~V Nair, Vitchyr Pong, Murtaza Dalal, Shikhar Bahl, Steven Lin, and
  Sergey Levine.
\newblock Visual reinforcement learning with imagined goals.
\newblock \emph{Advances in neural information processing systems}, 31, 2018.

\bibitem[Pan et~al.(2023{\natexlab{a}})Pan, Malkin, Zhang, and
  Bengio]{pan2023better}
Ling Pan, Nikolay Malkin, Dinghuai Zhang, and Yoshua Bengio.
\newblock Better training of gflownets with local credit and incomplete
  trajectories.
\newblock \emph{arXiv preprint arXiv:2302.01687}, 2023{\natexlab{a}}.

\bibitem[Pan et~al.(2023{\natexlab{b}})Pan, Zhang, Courville, Huang, and
  Bengio]{pan2022gafn}
Ling Pan, Dinghuai Zhang, Aaron Courville, Longbo Huang, and Yoshua Bengio.
\newblock Generative augmented flow networks.
\newblock \emph{International Conference on Learning Representations (ICLR)},
  2023{\natexlab{b}}.

\bibitem[Pan et~al.(2023{\natexlab{c}})Pan, Zhang, Jain, Huang, and
  Bengio]{pan2023stochastic}
Ling Pan, Dinghuai Zhang, Moksh Jain, Longbo Huang, and Yoshua Bengio.
\newblock Stochastic generative flow networks.
\newblock \emph{arXiv preprint arXiv:2302.09465}, 2023{\natexlab{c}}.

\bibitem[Radford et~al.(2019)Radford, Wu, Child, Luan, Amodei, Sutskever,
  et~al.]{radford2019language}
Alec Radford, Jeffrey Wu, Rewon Child, David Luan, Dario Amodei, Ilya
  Sutskever, et~al.
\newblock Language models are unsupervised multitask learners.
\newblock \emph{OpenAI blog}, 1\penalty0 (8):\penalty0 9, 2019.

\bibitem[Schaul et~al.(2015)Schaul, Horgan, Gregor, and
  Silver]{schaul2015universal}
Tom Schaul, Daniel Horgan, Karol Gregor, and David Silver.
\newblock Universal value function approximators.
\newblock In \emph{International conference on machine learning}, pages
  1312--1320. PMLR, 2015.

\bibitem[Sekar et~al.(2020)Sekar, Rybkin, Daniilidis, Abbeel, Hafner, and
  Pathak]{sekar2020planning}
Ramanan Sekar, Oleh Rybkin, Kostas Daniilidis, Pieter Abbeel, Danijar Hafner,
  and Deepak Pathak.
\newblock Planning to explore via self-supervised world models.
\newblock In \emph{International Conference on Machine Learning}, pages
  8583--8592. PMLR, 2020.

\bibitem[Sinai et~al.(2020)Sinai, Wang, Whatley, Slocum, Locane, and
  Kelsic]{sinai2020adalead}
Sam Sinai, Richard Wang, Alexander Whatley, Stewart Slocum, Elina Locane, and
  Eric Kelsic.
\newblock Adalead: A simple and robust adaptive greedy search algorithm for
  sequence design.
\newblock \emph{arXiv preprint}, 2020.

\bibitem[van Krieken et~al.(2022)van Krieken, Thanapalasingam, Tomczak, van
  Harmelen, and Teije]{van2022nesi}
Emile van Krieken, Thiviyan Thanapalasingam, Jakub~M Tomczak, Frank van
  Harmelen, and Annette~ten Teije.
\newblock A-nesi: A scalable approximate method for probabilistic neurosymbolic
  inference.
\newblock \emph{arXiv preprint arXiv:2212.12393}, 2022.

\bibitem[Veeriah et~al.(2018)Veeriah, Oh, and Singh]{veeriah2018many}
Vivek Veeriah, Junhyuk Oh, and Satinder Singh.
\newblock Many-goals reinforcement learning.
\newblock \emph{arXiv preprint arXiv:1806.09605}, 2018.

\bibitem[Zhang et~al.(2023{\natexlab{a}})Zhang, Rainone, Peschl, and
  Bondesan]{zhang2023robust}
David~W Zhang, Corrado Rainone, Markus Peschl, and Roberto Bondesan.
\newblock Robust scheduling with gflownets.
\newblock \emph{arXiv preprint arXiv:2302.05446}, 2023{\natexlab{a}}.

\bibitem[Zhang et~al.(2021)Zhang, Fu, Bengio, and Courville]{zhang2021unifying}
Dinghuai Zhang, Jie Fu, Yoshua Bengio, and Aaron Courville.
\newblock Unifying likelihood-free inference with black-box optimization and
  beyond.
\newblock \emph{arXiv preprint arXiv:2110.03372}, 2021.

\bibitem[Zhang et~al.(2023{\natexlab{b}})Zhang, Dai, Malkin, Courville, Bengio,
  and Pan]{zhang2023let}
Dinghuai Zhang, Hanjun Dai, Nikolay Malkin, Aaron Courville, Yoshua Bengio, and
  Ling Pan.
\newblock Let the flows tell: Solving graph combinatorial optimization problems
  with gflownets.
\newblock \emph{arXiv preprint arXiv:2305.17010}, 2023{\natexlab{b}}.

\bibitem[Zhang et~al.(2023{\natexlab{c}})Zhang, Pan, Chen, Courville, and
  Bengio]{zhang2023distributional}
Dinghuai Zhang, Ling Pan, Ricky~TQ Chen, Aaron Courville, and Yoshua Bengio.
\newblock Distributional gflownets with quantile flows.
\newblock \emph{arXiv preprint arXiv:2302.05793}, 2023{\natexlab{c}}.

\bibitem[Zhao et~al.(2021)Zhao, Gao, Abbeel, Tresp, and Xu]{zhao2021mutual}
Rui Zhao, Yang Gao, Pieter Abbeel, Volker Tresp, and Wei Xu.
\newblock Mutual information state intrinsic control.
\newblock \emph{arXiv preprint arXiv:2103.08107}, 2021.

\bibitem[Zimmermann et~al.(2022)Zimmermann, Lindsten, van~de Meent, and
  Naesseth]{zimmermann2022variational}
Heiko Zimmermann, Fredrik Lindsten, Jan-Willem van~de Meent, and Christian~A.
  Naesseth.
\newblock A variational perspective on generative flow networks.
\newblock \emph{arXiv preprint 2210.07992}, 2022.

\end{thebibliography}
